\documentclass[10pt]{article} % For LaTeX2e
\usepackage[accepted]{tmlr}
\usepackage{amsmath,amsfonts,bm}

\def\eqref#1{equation~\ref{#1}}
\def\1{\bm{1}}

\DeclareMathAlphabet{\mathsfit}{\encodingdefault}{\sfdefault}{m}{sl}
\SetMathAlphabet{\mathsfit}{bold}{\encodingdefault}{\sfdefault}{bx}{n}

\newcommand{\R}{\mathbb{R}}

\usepackage{hyperref}
\usepackage{url}

\usepackage{wrapfig}
\usepackage[acronym]{glossaries}
\usepackage{glossaries-extra}
\makeglossaries
\setglossarystyle{super}
\setabbreviationstyle[acronym]{long-short}
\glsdisablehyper

\newacronym[longplural=Artificial Neural Networks, shortplural=ANN]{ann}{ANN}{Artificial Neural Network}
\newacronym{ml}{ML}{Machine Learning}
\newacronym{mcmc}{MCMC}{Markov Chain Monte Carlo}
\newacronym{sgld}{SGLD}{Stochastic Gradient Langevin Dynamics}
\newacronym{lora}{LoRA}{Low-Rank Adaptation}
\newacronym{llm}{LLM}{Large Language Model}
\newacronym{gpu}{GPU}{Graphics Processing Unit}
\newacronym{ast}{AST}{Audio Spectrogram Transformer}
\newacronym{ece}{ECE}{Expected Calibration Error}
\newacronym{vit}{ViT}{Vision Transformer}
\newacronym{sgd}{SGD}{Stochastic Gradient Descent}
\newacronym{deit}{DeiT}{Data-Efficient Image Transformer}
\newacronym{mcdropout}{MC Dropout}{Monte Carlo Dropout}

\usepackage{multirow}
\usepackage{mathtools}
\DeclarePairedDelimiter{\abs}{\lvert}{\rvert}
\usepackage{graphics}
\DeclareGraphicsExtensions{.pdf, .PDF, .png, .PNG}
\usepackage{graphicx}
\usepackage{caption}
\usepackage{subcaption}
\usepackage{changepage}
\usepackage{listings} % For code formatting
\usepackage{booktabs}
\usepackage{etoc}
\title{LoRA-Ensemble: Efficient Uncertainty Modelling\\ for Self-Attention Networks}

\author{\name Dominik J. Mühlematter\thanks{Equal contribution.} \email dmuehlema@ethz.ch \\
\addr Photogrammetry and Remote Sensing, ETH Zurich
\AND
\name Michelle Halbheer \email hamich@ethz.ch \\
\addr Photogrammetry and Remote Sensing, ETH Zurich
\AND
\name Alexander Becker \email alexander.becker@geod.baug.ethz.ch  \\
\addr Photogrammetry and Remote Sensing, ETH Zurich
\AND
\name Dominik Narnhofer \email dnarnhofer@ethz.ch  \\
\addr Photogrammetry and Remote Sensing, ETH Zurich
\AND
\name Helge Aasen \email helge.aasen@agroscope.admin.ch  \\
\addr Earth Observation of Agroecosystems, Agroscope
\AND
\name Konrad Schindler \email schindler@ethz.ch  \\
\addr Photogrammetry and Remote Sensing, ETH Zurich
\AND \name Mehmet Ozgur Turkoglu\footnotemark[1] \email moturkoglu@gmail.com  \\
\addr  Earth Observation of Agroecosystems, Agroscope 
}
\begin{document}

\maketitle

\begin{abstract}
Numerous real-world decisions rely on machine learning algorithms and require calibrated uncertainty estimates. However, modern methods often yield overconfident, uncalibrated predictions. The dominant approach to quantifying the uncertainty inherent in the model is to train an ensemble of separate predictors and measure their empirical variance. In an explicit implementation, the ensemble has a high computational cost and memory footprint, especially if the base model itself is already large, like modern transformers. This motivates efforts to develop implicit ensemble methods that emulate the ensemble without explicitly instantiating all its members. We introduce LoRA-Ensemble, a parameter-efficient ensembling method for self-attention networks. It is based on Low-Rank Adaptation (LoRA), originally developed for efficient LLM fine-tuning, and extends it into an implicit ensembling scheme, where all ensemble members share the same, pre-trained self-attention network, but have individual low-rank matrices for the attention projections. The resulting method not only outperforms state-of-the-art implicit techniques like BatchEnsemble, but even matches or exceeds the accuracy of an Explicit Ensemble, while at the same time achieving superior calibration.
\end{abstract}

\section{Introduction}
Machine learning models are increasingly applied also in fields where incorrect estimates can have severe consequences, e.g., autonomous driving, medical diagnosis, (extreme) weather event prediction, or decision support for agriculture. In such applications, well-calibrated predictive uncertainties are crucial to enable self-diagnosis. Uncertainty can be separated into two components. \emph{Aleatoric uncertainty}, a.k.a.\ irreducible noise, is inherent in the data. In contrast, \emph{epistemic uncertainty} stems from a lack of knowledge about certain regions of the input space, due to a lack of training data \citep{DerKiureghian2009AleatoryMatter}.

Quantification of epistemic uncertainty in large machine learning models is non-trivial. Analytical computation is usually intractable; thus research has focused on efficient approximations \citep{Graves2011PracticalNetworks, Blundell2015WeightNetworks, Welling2011BayesianDynamics}. To date, probabilistic ensembles remain the best-performing approach \citep{Lakshminarayanan2017SimpleEnsembles}. In a naïve implementation, such an ensemble consists of multiple independently trained models. Individual models are interpreted as Monte Carlo samples from the posterior weight space and are used to obtain an unbiased estimator of the posterior distribution. To achieve a low correlation between ensemble members, one can capitalize on the stochastic nature of the training process and start from different initial weights, and/or sample different random batches of data. The basic principle is that the predictions of different ensemble members will agree near observed training samples, whereas they may vary far away from the training data. Their spread therefore serves as a measure of epistemic uncertainty. Empirically, even small ensembles often capture the uncertainty well (in expectation), i.e., they are well calibrated.

An issue with naïve ensembles is that their computational cost and memory footprint grow proportional to the number of ensemble members. For smaller models, the added cost and energy use may be acceptable. But for modern neural networks with up to several billion parameters, hardware restrictions render the naïve approach intractable, in particular, one can no longer hold the entire ensemble in memory.
Consequently, research has focused on ways to create ensembles implicitly, without requiring multiple copies of the full base model \citep{Wen2020BatchEnsemble:Learning, Wenzel2020HyperparameterQuantification, Huang2017SnapshotFree, Turkoglu2022FiLM}.
%
%Unfortunately, most of these parameter-efficient ensembling techniques are not applicable to the newest generation of neural networks. Transformer networks \citep{Vaswani2017AttentionNeed} have become popular due to their superior ability to capture complex structures in data. However, implicit ensembling schemes tend to underperform for transformers, as demonstrated in our experiments, or are incompatible with them, as detailed in Appendix~\ref{app_sec:imlicit_baseline}.
%
%\textcolor{blue}{These methods are developed for MLP or CNN architecture and it is not easily transferable or not even applicable to due its unique complexe computational structire. For instance FilM Ensemble is not directly applicable because its specifically designed to modulate via BAchNOrm layer whoch does not even exist in Transformer achitecture or SNGP is designed thingkathat bilptchit constrait but it does not hold for Transformers as detailed in Appensix~\ref{sec:sngp}.}
%
Unfortunately, most of these parameter-efficient ensembling techniques are not readily applicable to the newest generation of neural networks. Transformer networks \citep{Vaswani2017AttentionNeed} have become popular due to their superior ability to capture complex structures in data. However, implicit ensembling schemes tend to underperform for transformers, as demonstrated in our experiments, or are architecturally incompatible with them. These methods were developed primarily for MLP and CNN architectures and there is no straightforward way to port them to transformers due to fundamental differences in computation structure. For instance, FiLM-Ensemble~\citep{Turkoglu2022FiLM} modulates feature maps through Batch Normalization layers, which are absent in standard transformer architectures that rely on Layer Normalization instead (see Appendix~\ref{app_sec:imlicit_baseline} for details). Similarly, SNGP~\citep{liu2020simple} enforces bi-Lipschitz constraints via spectral normalization to ensure distance-aware uncertainty; however, dot-product self-attention has an unbounded Lipschitz constant~\citep{kim2021lipschitz}, violating SNGP's core theoretical assumptions (see Appendix~\ref{app:sngp} for details).

Several studies have shown that modern neural networks are heavily overparametrized and that their results have low intrinsic dimension \citep{Li2018MeasuringLandscapes, Aghajanyan2020IntrinsicFine-Tuning}. This led \citet{Hu2021LoRA:Models} to propose \glsxtrfullpl{lora} as a way of fine-tuning \glsxtrfullpl{llm} for different tasks while avoiding the prohibitively large memory and compute requirements of retraining them. It turns out that the weight matrices in such models can be factorized to have very low rank, with hardly any loss in prediction performance.

We show that \glsxtrshort{lora} can also serve as a basis for a novel, parameter-efficient ensemble method tailored to the transformer architecture. In line with the trend towards parameter-efficient fine-tuning, our method uses a pre-trained transformer model, which is expanded into an implicit ensemble by varying the \glsxtrshort{lora} factorization, while keeping the backbone weights frozen. In this way, our method requires a small number of additional parameters to turn an existing transformer model into a diverse ensemble whose performance across various tasks is comparable to an Explicit Ensemble. 
%
%\textcolor{green}{Compared to existing implicit approaches such as the widely used BatchEnsemble \citep{Wen2020BatchEnsemble:Learning}, which also relies on low-rank perturbations for parameter-efficient ensembling, our method introduces simple yet effective modifications that significantly improve model performance.}
%
In summary, our contributions are:
\begin{itemize}
\item We introduce \glsxtrshort{lora}-Ensemble, a parameter-efficient probabilistic ensemble method for self-attention networks.
%\item \glsxtrshort{lora}-Ensemble can be readily combined with most pre-trained transformer networks, irrespective of their specific architecture and application domain: it simply replaces the linear projection layers in the attention module with \glsxtrshort{lora} layers.
\item \glsxtrshort{lora}-Ensemble can be readily combined with a wide range of standard pre-trained transformer architectures that employ the standard self-attention mechanism: it simply replaces the linear projection layers in the attention module with \glsxtrshort{lora} layers. 
\item We apply \glsxtrshort{lora}-Ensemble to different classification tasks, including
conventional image labeling, skin lesion classification in dermatoscopic images, fine-grained image classification, sound classification, out-of-distribution (OOD) detection, and language modeling; and demonstrate significant gains in accuracy and uncertainty modeling.

%\textcolor{red}{Delete: conventional image labeling, classification of skin lesions in dermatoscopic images, fine-grained image labeling on a large dataset,  sound classification from spectrograms, and out-of-distribution (OOD) detection.} In these experiments, \glsxtrshort{lora}-Ensemble not only consistently outperforms other implicit ensemble schemes but also, surprisingly, its classification accuracy and uncertainty calibration are often even better than that of an Explicit Ensemble.
%\item We analyze the superior performance of LoRA-Ensemble compared to Explicit Ensemble by exploring the diversity of ensemble members in both function and weight spaces. Further, our ablation studies explore the effects of varying the LoRA rank, initialization schemes, and model sizes, different parameter sharing strategies, as well as applying LoRA-Ensemble to convolutional neural networks (CNNs).

\item We demonstrate that LoRA-Ensemble outperforms traditional Explicit Ensembles by fostering greater diversity among members, both in their learned functions and in weight space. 

\item We conduct extensive empirical analyses of how LoRA rank, initialization scheme, model scale, and parameter-sharing strategies impact performance, and we adapt LoRA-Ensemble for convolutional neural networks (CNNs) to demonstrate its broad applicability.

\end{itemize}

\section{LoRA-Ensemble}
The \glsxtrfull{lora} technique makes it possible to use a pre-trained model and fine-tune it without having to retrain all its parameters. This is particularly beneficial for modern neural networks with large parameter spaces. The underlying principle is to freeze the pre-trained model weights $W_0\in \R^{k\times d}$ and instead constrain the updates to a low-rank decomposition. This can be expressed mathematically as: 
\begin{equation}
    W = W_0 + \Delta W= W_0 + B\!\cdot\!A\;.
\end{equation}
Here $B\in\R^{k\times r}$ and $A\in\R^{r\times d}$ are two trainable low-rank matrices, where $r\ll\min(d,k)$. $W$ and $\Delta W$ are then multiplied by the same input $x$, which yields the following modified forward pass:
\begin{equation}
    h = W_0\cdot x + \Delta W\cdot x = W_0\cdot x + B\!\cdot\!A\cdot x\;.
\end{equation}
%
%\glsxtrshort{lora} applies this low-rank updating scheme only to weights in the self-attention modules of a transformer model while leaving the interleaved MLP modules untouched. I.e., the weight matrices being updated are $W_q$, $W_k$, and $W_v$ for the query, key, and value of the attention mechanism, as well as the $W_o$ for merging the multi-head outputs. The former three are each treated as a single matrix, disregarding the fact that they are typically sliced into multiple attention heads \citep{Hu2021LoRA:Models}. {\color{blue}Additional ablations on layer placement and ensemble design choices are provided in Appendix~\ref{app:placement_lora_ensemble}.}
%
\glsxtrshort{lora} applies its low-rank adaptation scheme exclusively to the weight matrices of the self-attention modules in a transformer, while leaving the interleaved MLP layers untouched. Concretely, the adapted weights are $W_q$, $W_k$, and $W_v$, which project the input into query, key, and value representations, along with $W_o$, which merges the outputs of the attention heads. As in \citet{Hu2021LoRA:Models}, the projection matrices are treated as single units, disregarding their typical partitioning into multiple attention heads. In Appendix~\ref{app:placement_lora_ensemble}, we provide additional ablations on the placement of LoRA layers within the transformer, as well as on ensemble design choices, illustrating how these factors impact predictive performance, calibration, and efficiency.

\begin{figure}
    \centering
    \includegraphics[width=0.8\textwidth]{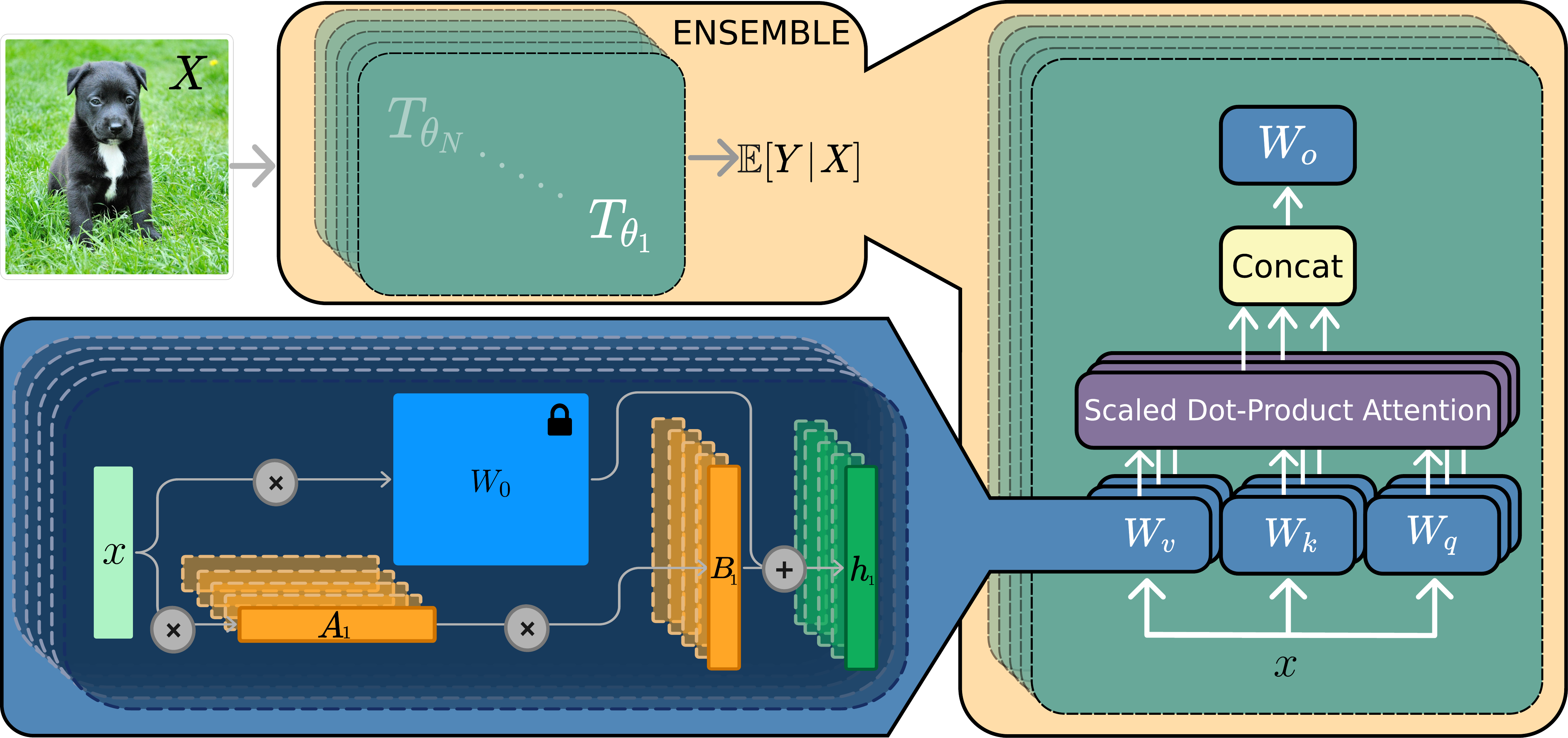}
    \caption{A schema of a \glsxtrshort{lora}-Ensemble. The computation structure of the multi-head self-attention module (right), and \glsxtrshort{lora}-Ensemble module (bottom left). $X$ denotes the actual input, and $x$ represents the intermediate input representation.}
    %The stacked states of the $A$ and $B$ matrices and the output $h$ represent the different ensemble members.}
    \label{fig:schema_lora_ensemble}
\end{figure}

Although not designed with uncertainty calibration in mind, the \glsxtrshort{lora} concept fulfills all the requirements of an implicit deep ensemble: By modifying the weights of the highly nonlinear self-attention mechanism one is able to generate a diverse collection of networks with the same architecture and objective. By learning an additive, low-rank update $\Delta W = B\!\cdot\!A$ rather than directly tuning the weight matrices, the expansion into a model ensemble adds only a small number of parameters and is efficient.
In detail, we start from a single, pre-trained model with frozen parameters $W_0$ and expand it with a set of trainable low-rank matrices $\Delta W_i$, $\forall i = 1 \ldots N$.
At each transformer block, there now is a different forward pass per ensemble member $i$, as illustrated in Fig.~\ref{fig:schema_lora_ensemble}:
\begin{equation} \label{eq:LoRA_member}
    h_i = W_0\cdot x + \Delta W_i\cdot x = W_0\cdot x + B_i\!\cdot\!A_i\cdot x\;,
\end{equation}
leading to $N$ different predictions $T_{\theta_i}(X)$ for a given input $X$. From those individual predictions, we compute the ensemble mean and variance in the standard manner:
\begin{equation}
\mathbb{E}[Y | X] \approx \frac{1}{N} \sum_{i=1}^{N} T_{\theta_i}(X)
\quad,\quad
\text{Var}[Y | X] \approx \frac{1}{N} \sum_{i=1}^N\big(T_{\theta_i}(X)-\mathbb{E}[Y|X]\big)^2\;.
\end{equation}

Refer to Appendices \ref{app:implementation}, \ref{subsec:2trainingsettings}, and \ref{app_A:init_lora}, respectively, for implementation, training, and initialization details. We publicly release the PyTorch implementation of LoRA-Ensemble, along with pre-trained weights, on GitHub\footnote{\url{https://github.com/prs-eth/LoRA-Ensemble}}.

\section{Experiments}

\subsection{Experimental Setup}
We evaluate \glsxtrshort{lora}-Ensemble across multiple datasets spanning three modalities in terms of predictive accuracy, uncertainty calibration, and computational efficiency. Unless stated otherwise, reported results correspond to the mean over five independent runs with different random initializations, with $1\cdot\sigma$ error bars. For iNaturalist, we report results over three runs due to computational cost.

We evaluated each method's predictive performance using classification accuracy and F1 score, and its calibration quality through Expected Calibration Error (ECE), Negative Log-Likelihood (NLL), and Brier score. The ECE measures how far predicted confidences deviate from observed error rates, i.e., perfect calibration occurs when the estimated uncertainties match the actual likelihood of a misclassification. The definitions of all metrics are given in Appendix \ref{app:metrics}. 

%Implementation details of \glsxtrshort{lora}-Ensemble are provided in Appendix~\ref{app:implementation}. 

%A detailed sensitivity analysis of the LoRA rank, along with practical guidelines for hyperparameter selection, can be found in Appendix~\ref{subsec:sensitivity}.

The primary hyperparameter introduced by our method is the LoRA rank, $r$, which controls the dimensionality of the low-rank adaptation matrices and thus directly affects both model capacity and computational overhead. For each dataset, we specify the selected rank in the corresponding experimental subsection. A comprehensive sensitivity analysis of the LoRA rank, together with practical guidelines for its selection, is provided in Appendix~\ref{subsec:sensitivity}.

%We first describe the baselines in Section~\ref{subsec:baselines} and analyze computational cost in Section~\ref{subsec:compute}, then present each dataset along with its corresponding results in Sections~\ref{subsec:cifar100}--\ref{subsec:ood}.

\subsection{Baselines}
\glsxtrshort{lora}-Ensemble is evaluated against a range of baselines. As a sanity check, we include a single \glsxtrfull{vit} model as well as a \glsxtrshort{vit} model augmented with \glsxtrshort{lora} in the attention modules. These models lack dedicated mechanisms for uncertainty calibration and rely solely on class-conditional likelihoods to quantify uncertainty.

As a gold-standard reference for uncertainty estimation, we compare against an Explicit Ensemble, where the full set of weights is independently trained for every member starting from the same pre-trained backbone (see Appendix~\ref{app_B:explicit_init} for details).

In addition, we evaluate several common implicit approaches: (i) \glsxtrfull{mcdropout} as implemented in \citep{li2023dropkey}, (ii) Batch-Ensemble~\citep{Wen2020BatchEnsemble:Learning}, (iii) Snapshot Ensemble~\citep{Huang2017SnapshotFree}, and (iv) Last-layer Ensemble, similar to MIMO~\citep{Havasi2020TrainingPrediction}. Implementation details are provided in Appendix~\ref{app_sec:batchEnsemble}, \ref{app_sec:snapshot_implementation}, and \ref{app_sec:lle}. For challenges related to adapting certain implicit methods to transformers, we refer to Appendix~\ref{app_sec:imlicit_baseline}.

We also include Epistemic Neural Networks (EpiNet)~\citep{osband2022epistemic}, which augment a base network with a small auxiliary network that takes random epistemic indices as input, enabling ensemble-like uncertainty estimation without maintaining multiple full models (see Appendix~\ref{appendix:epinet} for details).

For the CIFAR-100 experiments, we additionally compare with Spectral-normalized Neural Gaussian Process (SNGP)~\citep{liu2020simple}, an efficient single-model uncertainty estimation method. For HAM10000, we include two recent adaptations designed to satisfy SNGP's bi-Lipschitz assumptions in transformers: L2~\citep{kim2021lipschitz} and LRFormer~\citep{ye2023lrformer} (see Appendix~\ref{app:sngp} for details). For the SST-2 experiments, we further compare against Bayesian-LoRA~\citep{yang2024bayesian}, a method for improving uncertainty calibration in fine-tuned large language models.

\subsection{Compute Cost}\label{subsec:compute}

LoRA-Ensemble achieves substantial efficiency gains over Explicit Ensembles: with 16 members on the CIFAR-100 dataset, it requires approximately 14 times fewer parameters, 9 times less inference memory, and delivers over 5 times faster inference (when batch size 1). See Tab.~\ref{tab:model_resources} and refer to Appendix~\ref{subsec:computationl_cost} for more details. Training time is comparable between the two methods, as LoRA-Ensemble processes all members jointly while the Explicit Ensemble trains members sequentially on a single GPU. The primary advantage of LoRA-Ensemble lies in its memory efficiency, enabling deployment of large ensembles on resource-constrained hardware where Explicit Ensembles would be infeasible. We note that Explicit Ensembles can be parallelized across multiple GPUs by distributing members to separate devices; however, this requires proportionally more hardware.

\subsection{CIFAR-100}\label{subsec:CIFAR100_results}
As a first sandbox experiment, we perform image classification for the popular, widely used CIFAR-100 benchmark \citep{Krizhevsky2009LearningImages} (see Appendix~\ref{app_A:more_experiment} for CIFAR-10 experiments). The dataset consists of 100 object classes, each with 600 samples, for a total size of 60$\,$000 images. From that set, 10$\,$000 images are designated as test data, with all classes equally distributed between the training and testing portions.

Quantitative results are summarized in Tab.~\ref{tab:model_performance_cifar100}. Reliability diagrams, along with plots depicting classification accuracy and \glsxtrshort{ece} as a function of ensemble size, are provided in Appendix~\ref{app_sec:CIFAR100_extra_diagrams}. The LoRA rank was set to 8 for this experiment. 

\glsxtrshort{lora}-Ensemble consistently reaches higher accuracy than \glsxtrshort{mcdropout}, and Snapshot Ensemble, with a notable edge of approximately 5 percentage points. 
Last-Layer Ensemble improves calibration but at the cost of a substantial accuracy drop, which we attribute to the limited number of trainable parameters.
SNGP fails to converge on the ViT backbone, reaching only approximately 32\% accuracy; we provide a detailed analysis in Appendix~\ref{app:sngp}.
EpiNet approaches the accuracy of Explicit Ensemble while its calibration falls short in comparison.
Despite its conceptual similarity to the \glsxtrshort{lora}-Ensemble, the Batch-Ensemble is the weakest performer among all methods when applied to transformers. Appendix~\ref{app_sec:batchEnsemble} examines this finding in detail and outlines key distinctions between the two approaches.
Surprisingly, \glsxtrshort{lora}-Ensemble also consistently surpasses the Explicit Ensemble by about 2 percentage points, apparently a consequence of the fact that already a single \glsxtrshort{vit} model, and thus every ensemble member, benefits from the addition of \glsxtrshort{lora}.

The \glsxtrshort{lora}-Ensemble also achieves better-calibrated predictive uncertainties than all implicit ensembling methods and the Explicit Ensemble. Interestingly, although a single LoRA network is already very well calibrated, forming an ensemble slightly degrades its calibration, an effect not observed for the NLL or Brier score (Tab.~\ref{tab:model_performance_cifar100}). The reliability diagram in Fig.~\ref{fig:reliability_diagram_lora_16_cifar100} in the appendix somewhat elucidates this unexpected behavior: \glsxtrshort{lora}-Ensemble is under-confident on CIFAR-100, i.e., its predictions are more accurate than its confidence suggests. As noted by \citep{Rahaman2020UncertaintyEnsembles}, ensembling under-confident models can worsen calibration since accuracy grows faster than confidence. While under-confidence may be preferable in safety-critical settings, where over-estimating uncertainty is safer than being over-confident, we show in Appendix~\ref{app_sec:temp_scaling} that simple post-hoc Temperature Scaling effectively corrects this and yields near-perfect calibration. 

Beyond transformers, we also adapt LoRA-Ensemble to convolutional neural networks, evaluating a ResNet-18 backbone on CIFAR-100; results are reported in Appendix~\ref{app:CNN}.

%For generalization to the CNN architecture, see Appendix~\ref{app:CNN}.

\begin{table}
    \caption{Model performance on the CIFAR-100 dataset for the compared methods. Ensembles have 16 members. Best score for each metric in \textbf{bold},  second-best \underline{underlined}.}
    \centering
  %   \scriptsize % Reduce font size
   % \setlength{\tabcolsep}{10pt} % Reduce horizontal space between columns
   \resizebox{1.0\linewidth}{!}{
    \begin{tabular}{lccccc}
        \toprule
        \textbf{Method} & \textbf{Accuracy ($\uparrow$)} & \textbf{F1 ($\uparrow$)} & \textbf{ECE ($\downarrow$)} & \textbf{NLL ($\downarrow$)} & \textbf{Brier ($\downarrow$)} \\
        \midrule
        Single Network & $76.6\pm0.3$ & $76.6\pm0.3$ & $0.145\pm0.004$ & $1.181\pm0.019$ & $0.370\pm0.004$ \\
        Single Net w/ LoRA & $79.6\pm0.2$ & $79.4\pm0.2$& $\textbf{0.014}\pm0.003$ & $\underline{0.671}\pm0.005$ & $0.286\pm0.003$\\
        MC Dropout & $77.1\pm0.5$ & $77.2\pm0.4$& $0.055\pm0.002$ & $1.138\pm0.014$ & $0.336\pm0.005$\\

        Last-layer Ensemble & $73.4\pm0.0$ & $73.0\pm0.0$ & $0.093\pm0.000$ & $0.978\pm0.000$ & $0.376\pm0.000$\\

        Snapshot Ensemble & $77.0\pm0.1$ & $77.2\pm0.2$& $0.123\pm0.002$ & $4.416\pm0.046$ & $1.614\pm0.007$\\

        SNGP & $32.2\pm0.4$ & $30.1\pm0.4$ & $0.072\pm0.004$ & $2.744\pm0.010$ & $0.817\pm0.002$\\
         
        ENN (EpiNet) & $79.7\pm0.2$ & $79.7\pm0.2$ & $0.128\pm0.003$ & $1.016\pm0.015$ & $0.323\pm0.005$ \\
         Batch-Ensemble & $68.8\pm0.1$ & $68.5\pm0.1$& $0.102\pm0.002$ & $1.093\pm0.002$ & $0.437\pm0.001$\\
        Explicit Ensemble & $\underline{79.8}\pm0.1$ & $\underline{79.8}\pm0.2$& $0.100\pm0.001$ & $0.745\pm0.003$ & $\underline{0.284}\pm0.002$\\
        
        \midrule
        LoRA-Ensemble & $\textbf{82.5}\pm0.1$ & $\textbf{82.5}\pm0.1$& $\underline{0.035}\pm0.001$ & $\textbf{0.587}\pm0.001$ & $\textbf{0.253}\pm0.000$\\
        \bottomrule
    \end{tabular}
    }
    \label{tab:model_performance_cifar100}
\end{table}

\subsection{HAM10000 Lesion Classification}
The HAM10000 dataset was proposed for the \emph{Human Against Machine with 10$\,$000 training images} study \citep{Tschandl2018TheLesions}. It consists of 10$\,$015 dermatoscopic images of pigmented skin lesions, collected from different populations. We use a stratified 80/20 split at the image level, without prior near-duplicate filtering, patient- or lesion-level grouping. The dataset was initially assembled to compare machine learning methods against medical professionals on the task of classifying common pigmented skin lesions. Compared to CIFAR-100, this is arguably a more relevant test bed for our method: in the medical domain, uncertainty calibration is critical, due to the potentially far-reaching consequences of incorrect diagnoses and treatment planning.

We therefore evaluate the same group of models on this task, focusing on both predictive accuracy and calibration. The results are summarized in Tab.~\ref{tab:model_performance_HAM10000}. The LoRA rank was set to 4 for this experiment. 
Similar to the CIFAR-100 evaluation, \glsxtrshort{lora}-Ensemble outperforms all other methods by a clear margin, with respect to both classification accuracy and calibration.

The experiments also further support the above discussion of confidence vs.\ ensemble size (Sec.~\ref{subsec:CIFAR100_results}). For HAM10000, \glsxtrshort{lora}-Ensemble is slightly over-confident (just like the Explicit Ensemble) and, indeed, its calibration error decreases with ensemble size in this case, see Appendix~\ref{app_sec:HAM10000_extra_diagrams}.

We conducted further experiments on HAM10000 using different backbone architectures (DeiT) with varying numbers of parameters. See Tab.~\ref{tab_app:model_performance_HAM10000_smaller_models} in Appendix~\ref{app:model_size_study}. 
LoRA-Ensemble generalizes smoothly across different backbones, and as the number of parameters in the backbone increases, its advantage over the Explicit Ensemble becomes more pronounced, in both accuracy and calibration.

\begin{table}[ht]
    \caption{Model performance on the HAM10000 dataset for the compared methods. Ensembles have 16 members. Best score for each metric in \textbf{bold},  second-best \underline{underlined}}.
    \centering
    %\scriptsize % Reduce font size
    %\setlength{\tabcolsep}{10pt} % Reduce horizontal space between columns
    \resizebox{1.0\linewidth}{!}{
    \begin{tabular}{lccccc}
        \toprule
    \textbf{Method} & \textbf{Accuracy ($\uparrow$)} & \textbf{F1 ($\uparrow$)} & \textbf{ECE ($\downarrow$)} & \textbf{NLL ($\downarrow$)} & \textbf{Brier ($\downarrow$)} \\
    \midrule
    Single Network & $84.1\pm0.3$ & $71.4\pm0.7$ & $0.139\pm0.004$ & $1.138\pm0.040$ & $0.291\pm0.009$ \\
    Single Net w/ LoRA & $83.2\pm0.7$ & $70.7\pm1.3$& $0.085\pm0.004$ & $0.569\pm0.027$ & $0.256\pm0.011$\\
    LRFormer & $74.3\pm 1.9$ & $52.1\pm3.2$ & $\underline{0.053} \pm0.022$ & $0.737\pm0.014$ & $0.354 \pm 0.011$ \\
    L2 & $74.1\pm1.8$ & $50.7\pm3.9$ & $0.065\pm0.024$ & $0.766\pm 0.036$ & $0.360\pm0.021$ \\
    
    MC Dropout & $83.7\pm0.4$ & $71.0\pm0.9$& $0.099\pm0.007$ & $0.631\pm0.023$ & $0.270\pm0.009$\\

    %\textcolor{blue}{Last-layer Ensemble} & $\pm0.$ & $\pm0.$& $\pm0.00$ & $\pm0.0$ & $\pm0.00$\\

    %\textcolor{blue}{ENN (Epinet)} & $\pm0.$ & $\pm0.$& $\pm0.00$ & $\pm0.0$ & $\pm0.00$\\
    
    Snapshot Ensemble & $84.9\pm0.3$ & $73.7\pm0.9$& $0.058\pm0.004$ & $\underline{0.431}\pm0.007$ & $\underline{0.217}\pm0.004$\\
    Batch-Ensemble & $76.8\pm1.6$ & $58.4\pm2.8$& $0.064\pm0.021$ & $0.651\pm0.003$ & $0.332\pm0.002$\\
    
    Explicit Ensemble & $\underline{85.8}\pm0.2$ & $\underline{74.6}\pm0.4$& $0.105\pm0.002$ & $0.536\pm0.007$ & $0.218\pm0.002$\\
    \midrule
    LoRA-Ensemble & $\textbf{88.0}\pm0.2$ & $\textbf{78.3}\pm0.6$& $\textbf{0.037}\pm0.002$ & $\textbf{0.342}\pm0.003$ & $\textbf{0.175}\pm0.002$\\
    \bottomrule
    \end{tabular}
    }
    \label{tab:model_performance_HAM10000}
\end{table}

\subsection{Large-Scale Fine-Grained Image Classification with iNaturalist.}
To demonstrate that LoRA-Ensemble scales to large, fine-grained, real-world datasets, we apply it to iNaturalist 2017~\citep{van2018inaturalist}, comprising 675$\,$170 images across 5$\,$089 species, an order of magnitude larger than CIFAR-100. Severe class imbalance, high intra-class variability, and subtle inter-class differences make uncertainty quantification especially challenging due to the difficulty to avoid overconfident errors among similar species, and to flag uncertain predictions for rare species. The \glsxtrshort{lora} rank was set to 64 for this experiment.

As shown in Tab.~\ref{tab_app:model_performance_INat2017}, our LoRA-Ensemble almost matches the Explicit Ensemble in accuracy, while substantially improving the calibration, using only a fraction of the parameters and compute. This demonstrates that the method scales well and enables reliable uncertainty estimation for large, fine-grained, imbalanced datasets. Refer to Appendix~\ref {app:joint_inat} for additional results.

\begin{table}[h]
    \small
    \caption{Performance on the iNat 2017 dataset for all compared methods using three different random seeds. Ensembles have 4 members. Best score for each metric in \textbf{bold},  second-best \underline{underlined}.}
    \centering
    \begin{tabular}{lccccc}
    \toprule
    \textbf{Method} & \textbf{Accuracy ($\uparrow$)} & \textbf{F1 ($\uparrow$)} & \textbf{ECE ($\downarrow$)} & \textbf{NLL ($\downarrow$)} & \textbf{Brier ($\downarrow$)} \\
    \midrule
    Single Network & $42.6\pm0.2$ & $37.8\pm0.2$& $0.293\pm0.002$ & $1.054\pm0.001$ & $0.207\pm0.001$\\
    Single Net w/ LoRA & $47.7\pm0.1$ & $43.1\pm0.1$& $\underline{0.096}\pm0.001$ & $\underline{0.662}\pm0.001$ & $0.166\pm0.000$\\
    MC Dropout & $47.5\pm0.1$ & $40.3\pm0.1$& $0.206\pm0.002$ & $0.895\pm0.002$ & $0.172\pm0.000$\\
    Explicit Ensemble & $\textbf{49.6}\pm0.2$ & $\textbf{44.6}\pm0.3$& $0.199\pm0.002$ & $0.716\pm0.002$ & $\underline{0.165}\pm0.000$\\
    \midrule
    LoRA-Ensemble & $\underline{49.3}\pm0.1$ & $\underline{44.1}\pm0.2$& $\textbf{0.045}\pm0.001$ & $\textbf{0.610}\pm0.000$ & $\textbf{0.160}\pm0.000$\\
    \bottomrule
    \end{tabular}
    \label{tab_app:model_performance_INat2017}
\end{table}

\subsection{Extension to Additional Modalities: Audio (ESC-50) and Language (SST-2)}
As a further benchmark from a different application domain, we process the ESC-50 environmental sounds dataset \citep{Piczak2015ESC:Classification}. It consists of 2$\,$000 sound samples, each five seconds long, that represent 50 different semantic classes with 40 samples each. To prepare the raw input waveforms for analysis, they are converted into 2-dimensional time/frequency spectrograms, see \citep{Gong2021AST:Transformer}. These spectrograms form the input for \glsxtrlong{ast}, a state-of-the-art transformer model for sound classification. The LoRA rank is set to 16 in this experiment.

%We also extend our evaluation to natural language processing with the SST-2 sentiment classification dataset, using BERT base uncased~\citep{socher2013recursive, Devlin2019BERTPO}. 
%
Moreover, we extend our empirical evaluation to the natural language domain using the SST-2 binary sentiment classification benchmark~\citep{socher2013recursive}. SST-2 consists of short movie review sentences annotated with positive or negative sentiment and is part of the GLUE benchmark~\citep{wang2018glue}. The standard GLUE split contains 67,349 training samples and 872 validation samples.
We use BERT base uncased~\citep{Devlin2019BERTPO} as a backbone, and the LoRA rank is set to 64 in this experiment.

Results for both datasets are provided in Appendix~\ref{app_sec:ESC_50_results} and Appendix~\ref{app:sst2}.
These results suggest that the benefits of LoRA-Ensemble extend beyond the vision domain: while the Explicit Ensemble retains a slight edge in predictive accuracy on both tasks, LoRA-Ensemble remains competitive overall and continues to deliver strong calibration, consistent with the trends observed in the image classification experiments.

\subsection{Out-of-Distribution Detection \& Dataset Shift Robustness}
\label{subsec:ood_detection}

For the out-of-distribution (OOD) experiment, models are trained on CIFAR-100 and evaluated on in-distribution samples from CIFAR-100 as well as OOD samples from CIFAR-10 or SVHN~\citep{netzer2011reading}, following standard practice~\citep{hendrycks2016baseline}. Following \citet{Sim2023AttentionMasking} and \citet{Chen2023SplitEnsemble}, we use the maximum softmax probability as the confidence score. Performance is measured using AUROC, AUPRC, and FPR 95\% TPR.

Table~\ref{tab:model_performance_ood} shows that \glsxtrshort{lora}-Ensemble achieves superior performance across both settings and metrics, surpassing all baselines, including the recently proposed Split-Ensemble approach~\citep{Chen2023SplitEnsemble}, which was specifically designed for OOD detection. Consistent with earlier observations, even a single \glsxtrshort{lora} model outperforms the Explicit Ensemble, highlighting its robustness in OOD scenarios.

To further assess robustness under distribution shifts, we evaluate on the CIFAR-10/100-C benchmarks across varying corruption severities. As detailed in Appendix~\ref{app_sec:distribution_shift}, LoRA-Ensemble consistently maintains superior accuracy and calibration under increasing shift intensity.

\begin{table}[t]
  \caption{Model performance on the OOD task. CIFAR-100 is used as the in-distribution dataset and CIFAR-10 and SVHN as the out-of-distribution dataset. Ensembles for all methods consist of 16 members. Results for Split-Ensemble are taken from \citep{Chen2023SplitEnsemble}. The best score for each metric is highlighted in \textbf{bold}, with the second-best score \underline{underlined}.}
    \centering
    \resizebox{1.0\linewidth}{!}{
    \begin{tabular}{lcccccc}
        \toprule
        \textbf{OOD Dataset} & \multicolumn{3}{c}{\textbf{CIFAR-10}} & \multicolumn{3}{c}{\textbf{SVHN}} \\
        \midrule
        \textbf{Method} & \textbf{AUROC  ($\uparrow$)} & \textbf{AUPRC  ($\uparrow$)}& \textbf{FPR 95\% TPR ($\downarrow$)}  & \textbf{AUROC  ($\uparrow$)} & \textbf{AUPRC  ($\uparrow$)}& \textbf{FPR 95\% TPR ($\downarrow$)}\\
        \midrule
        Split-Ensemble \citep{Chen2023SplitEnsemble} & $79.2$  & 81.7 & 78.5 &81.2& 69.9& 75.0 \\
        \midrule
        Single Network& $75.6\pm0.3$ &  $71.5\pm0.4$& $64.7\pm1.2$ & $76.4\pm1.8$ & $86.8\pm1.0$& $55.9\pm3.0$ \\ 
        Single Network with LoRA & $\underline{80.0}\pm0.1$ & $\underline{78.0}\pm0.3$ & $\underline{57.3}\pm0.8$ &$\underline{85.9}\pm0.9$ & $\underline{93.1}\pm0.3$& $\underline{49.7}\pm3.0$ \\
        MC Dropout & $56.6\pm10.3$ & $55.6\pm9.6$ & $92.2\pm4.3$ & $52.3\pm12.4$ & $74.5\pm8.9$ & $94.8\pm3.4$ \\

        ENN (EpiNet) & $77.4\pm0.2$ & $73.8\pm0.1$ & $63.0\pm0.9$ & $78.6\pm0.6$ & $88.0\pm0.2$ & $50.5\pm0.4$\\

        Explicit Ensemble & $79.0\pm0.1$ & $75.4\pm0.2$ & $59.7\pm0.9$ &$74.8\pm1.3$ & $86.6\pm0.7$ & $61.0\pm1.8$\\
        \midrule
        LoRA-Ensemble & $\textbf{82.1}\pm0.1$ & $\textbf{80.4}\pm0.1$ & $\textbf{54.1}\pm0.3$ &$\textbf{89.9}\pm0.6$ & $\textbf{95.2}\pm0.3$& $\textbf{41.6}\pm1.6$ \\
        \bottomrule
    \end{tabular}
    }
    \label{tab:model_performance_ood}
\end{table}

\section{Enhanced Diversity in LoRA-Ensemble}
\label{section:diversity}
%This section explores the diversity of ensemble members in function and weight space for \glsxtrshort{lora}-Ensemble and Explicit Ensemble, using the HAM10000 dataset with 16 ensemble members. Diversity is crucial for effective ensembles, as highly correlated members offer limited value~\citep{zhang2012ensemble}. With finite training data, capturing diverse parameter configurations that equally explain the observations is key to quantifying epistemic uncertainty~\citep{kendall2017uncertainties,fort2019deep}. %
To better understand the behavior of \glsxtrshort{lora}-Ensemble, we explore the diversity of its members and compare it to the Explicit Ensemble. The experiments are run on HAM10000 with 16 ensemble members. Diversity is crucial for effective ensembles, as highly correlated members offer little added value~\citep{zhang2012ensemble}. If an ensemble contains diverse parameter configurations that equally explain observations, then it will more comprehensively capture the epistemic uncertainty~\citep{kendall2017uncertainties}. For empirical evidence, refer to Appendix~\ref{app:diversity_performance_correlation}.

\begin{figure}[th]
    \setlength{\tabcolsep}{1pt}
    \centering
    \scriptsize
    \begin{tabular}{ccccc}
        Explicit & LoRA & Explicit & LoRA & \\
    \includegraphics[width=.19\textwidth]{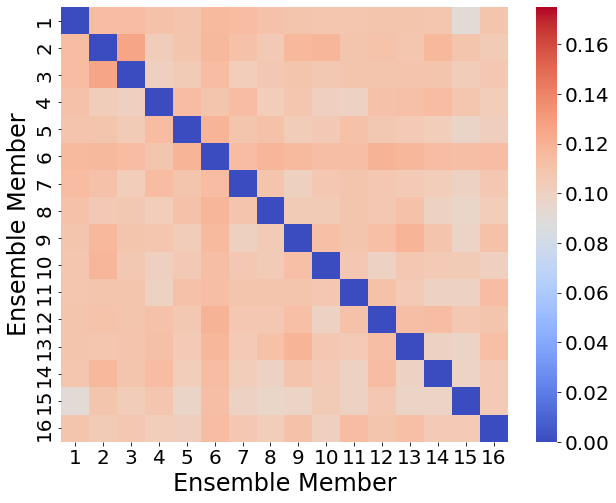} &   \includegraphics[width=.19\textwidth]{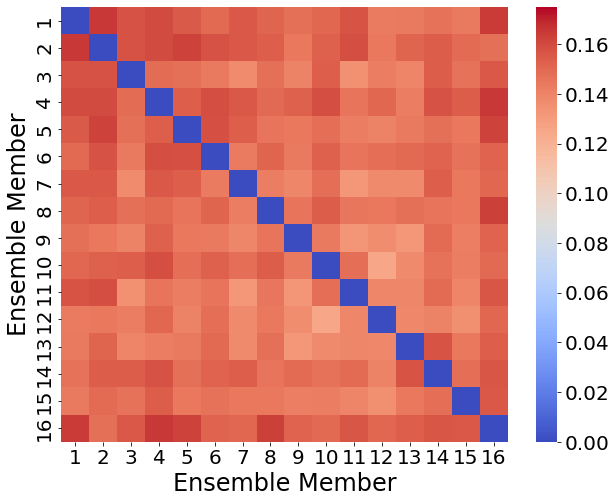} &
    \includegraphics[width=.19\textwidth]{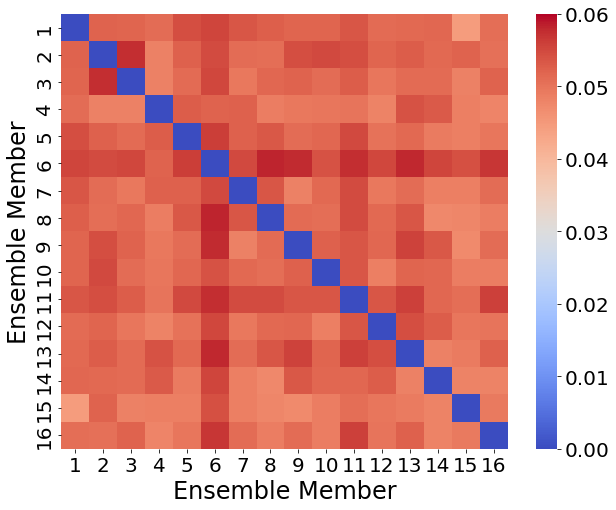} &
    \includegraphics[width=.19\textwidth]{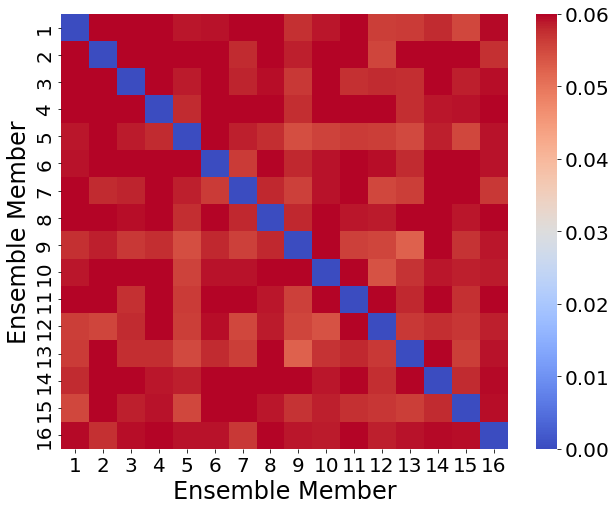} &
    \includegraphics[width=.19\textwidth]{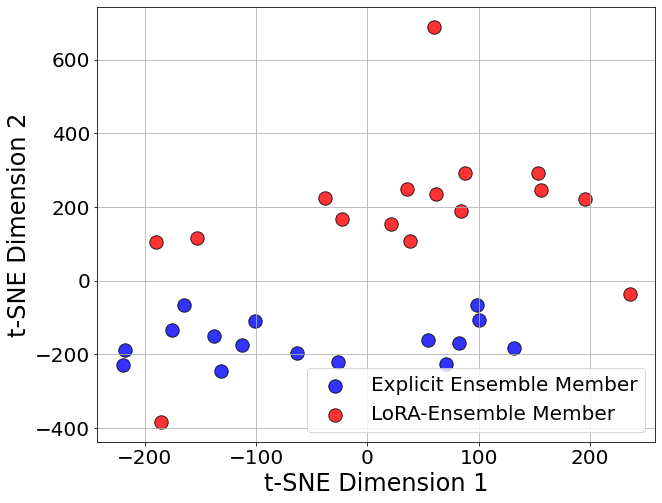}\\
        \multicolumn{2}{c}{(a) pairwise disagreement rate} &
    \multicolumn{2}{c}{(b) Jensen-Shannon divergence} &
    (c) t-SNE\\
    \end{tabular}
 \caption{Function space analysis of \glsxtrshort{lora}-Ensemble vs.\ Explicit Ensemble.}
    \label{fig:diversity_prediction}
\end{figure}

Following \citep{fort2019deep}, we first assess function space diversity through the predictions of individual ensemble members. 
In Fig.~\ref{fig:diversity_prediction}, we first compute the disagreement rate on the test set, defined as \( \frac{1}{N} \sum_{n=1}^N \mathbb{I}[T_{\theta_i}(X_n) \neq T_{\theta_j}(X_n)] \), where \( T_{\theta_i}(X_n) \) represents the class label predicted by ensemble member \( i \) for input \( X_n \), and \( \mathbb{I} \) is the indicator function.
Next, we construct a probability distribution for each ensemble member by aggregating their softmax outputs across all test samples, then compute pairwise Jensen-Shannon divergences (JSD). Finally, we use t-SNE~\citep{van2008visualizing} to visualize their spread in function space (aggregated softmaxes).
The analysis reveals that LoRA-Ensemble exhibits significantly higher diversity among ensemble members compared to an Explicit Ensemble. I.e., LoRA-Ensemble appears to capture a wider range of modes in function space.

We further inspect the weight spaces of LoRA-Ensemble and Explicit Ensemble with spectral analysis, focusing on the projection matrices within the attention blocks of the ViT (Base-32) model pre-trained on ImageNet. We show the analysis for value projection matrices, given their strong association with learned representations; details for query and key projection matrices can be found in Appendix~\ref{app:weight_space}.
%
%Using Singular Value Decomposition, a matrix \( W \in \mathbb{R}^{m \times n} \) can be expressed as \( W = U \Sigma V^\top \), where \( U \in \mathbb{R}^{m \times m} \) and \( V \in \mathbb{R}^{n \times n} \) contain orthonormal columns representing the singular vectors of \( W \), and \( \Sigma \in \mathbb{R}^{m \times n} \) is a diagonal matrix of singular values. Here, \( U \) and \( V \) describe rotational components, while \( \Sigma \) captures the scaling effect. Singular vectors associated with larger singular values reflect the most significant transformations encoded in \( W \).
%
We employ Singular Value Decomposition (SVD) to identify the most significant transformations encoded in the weights, following the logic that larger singular values correspond to the most impactful components.
As proposed by \citep{shuttleworth2024lora}, we analyze the similarity between the initial (pre-trained) weights and the final trained weights of ensemble members. LoRA-Ensemble and Explicit Ensemble lead to very different parameter updates. LoRA-Ensemble introduces new high-ranking singular vectors that are near-orthogonal to those in the initial weights, referred to as "intruder dimensions" \citep{shuttleworth2024lora}. In contrast, Explicit Ensemble members tend to adhere closely to the spectral structure of the initial weights (see Fig.~\ref{fig:weight_diversity_intruder} in Appendix).

The random initialization of matrices \(A\) and \(B\) in the LoRA module leads to an intriguing phenomenon: the intruder dimensions of different LoRA-Ensemble members are near-orthogonal, as shown by the cosine similarities between the highest-ranking singular vectors of different members in Fig.~\ref{fig:weight_diversity} (for details see Appendix~\ref{app:weight_space}). The figure shows similarity values averaged over layers and pairs of members, for rank 4. Notably, the highest-ranked singular vectors of distinct members exhibit almost no similarity; in contrast to the Explicit Ensemble, where they are highly correlated.
The weight-space cosine similarity provides further evidence of enhanced diversity. LoRA-Ensemble members exhibit greatly increased diversity in weight space. 
To visualize training trajectories, we apply t-SNE to plot the evolution of the model weights during training. LoRA-Ensemble members span a larger part of the loss landscape, indicating diverse learning dynamics. In contrast, Explicit Ensemble members remain closer to the initial weights, reflecting reduced diversity. 
Overall, these results suggest LoRA-Ensemble better explores the weight space, and thus the epistemic uncertainty.

\begin{figure}[th]
    \setlength{\tabcolsep}{1pt}
    \centering
    \scriptsize
    \begin{tabular}{ccccc}
        Explicit & LoRA & Explicit & LoRA & \\
        \includegraphics[width=.19\textwidth]{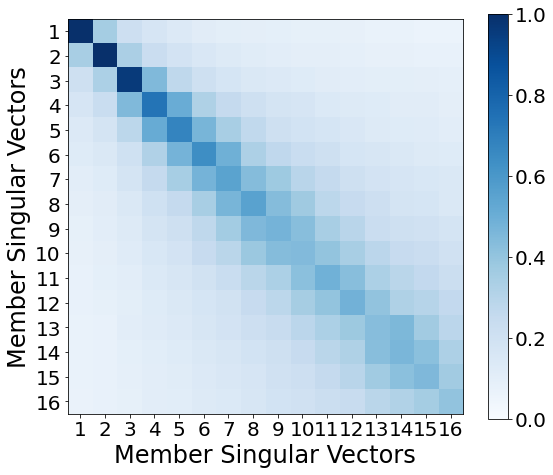} &
        \includegraphics[width=.19\textwidth]{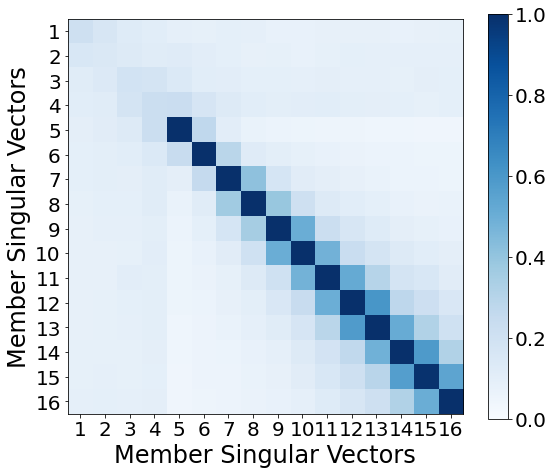} &
        \includegraphics[width=.19\textwidth]{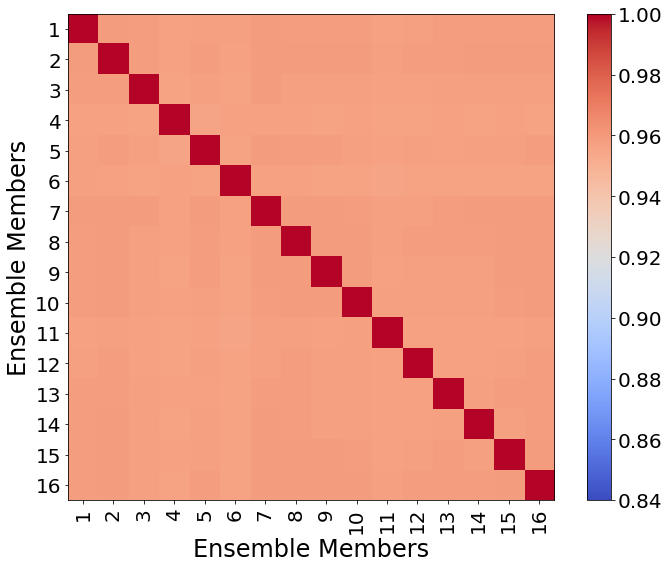} &
        \includegraphics[width=.19\textwidth]{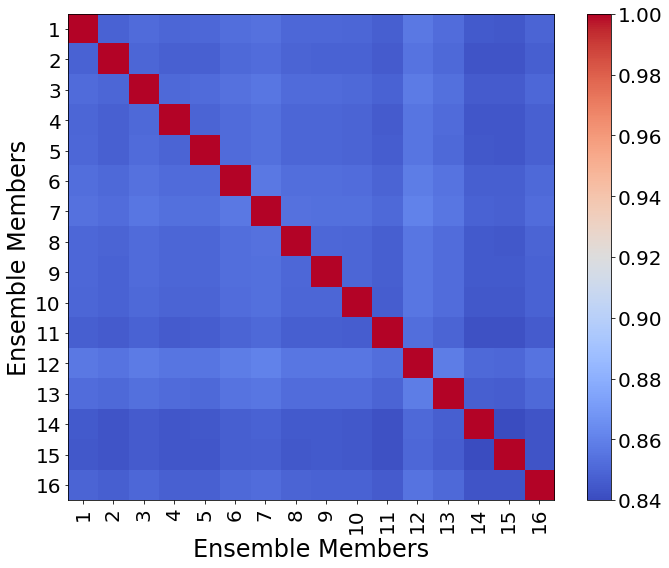} &
        \includegraphics[width=.19\textwidth]{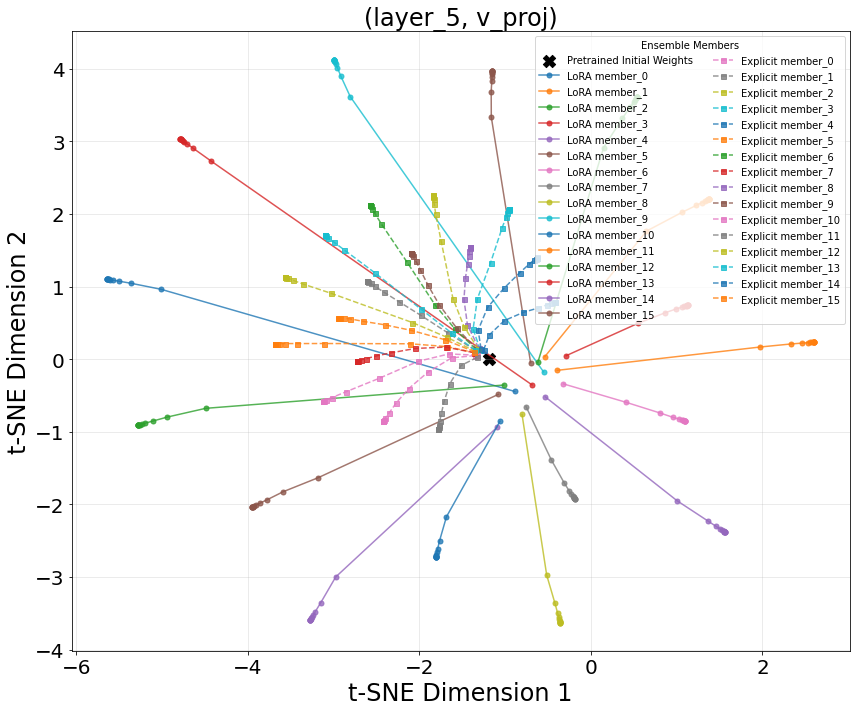} \\
        \multicolumn{2}{c}{(a) cos-similarity of high-ranking singular vectors} &
    \multicolumn{2}{c}{(b) weight-space cosine similarity} &
    (c) training trajectories\\
    \end{tabular} 
 \caption{Weight space analysis of \glsxtrshort{lora}-Ensemble vs.\ Explicit Ensemble.}
 \vspace{-0.5em}
\label{fig:weight_diversity}
\end{figure}

\section{Discussion} %\& Limitations \& Future Work}
\textbf{Effectiveness of \glsxtrshort{lora}-Ensemble.}
Across diverse tasks, our experiments show that LoRA-Ensemble matches or surpasses the predictive performance of the state-of-the-art Explicit Ensemble while offering superior calibration. Note, though, that there are exceptions to this trend: e.g., on ESC-50, the performance of the Explicit Ensemble is comparable, and on SST-2 it retains a small advantage.
Adding \glsxtrshort{lora} to a single model \emph{without any ensembling} improves calibration in most experiments beyond that of a 16-member Explicit Ensemble. This effect may be linked to the well-documented over-parameterization of modern neural networks, which often achieve higher predictive accuracy at the cost of poorer calibration \cite[e.g.,][]{Guo2017OnNetworks}. By incorporating LoRA while treating all pre-trained weights as constants, we significantly reduce the trainable parameter space, potentially favoring better calibration.
However, limiting trainable parameters alone does not ensure better accuracy or calibration, e.g., many forms of regularization or selective training may fall short.
We believe that the effectiveness of LoRA-Ensemble stems from its unique learning dynamics, which we explore in Sec.~\ref{section:diversity} and Appendix~\ref{app:weight_space}. Its members converge to diverse solutions that span a broader area of the loss landscape, enabling better exploration of the weight space and more effective estimates of epistemic uncertainty.
Increasing the number of members in the \glsxtrshort{lora}-Ensemble enhances predictive power, potentially improving accuracy while maintaining good calibration due to the limited number of trainable weights. However, if the trainable weights are not limited, e.g. when increasing the LoRA rank too far, calibration can worsen, as shown in Fig.\ref{fig:ham10000_rank_study}, Tab.\ref{tab:model_performance_large_rank}.
This effect aligns with the findings of \citet{shuttleworth2024lora}, which indicate that excessively increasing the rank of LoRA may cause it to lose its unique learning dynamics.
Furthermore, jointly training the backbone and optimizing all parameters simultaneously degrades performance, see Appendix \ref{app:joint_inat} for details.
Conversely, enhancing predictive power by increasing the pre-trained weights (while keeping trainable weights constant) further improves the effectiveness of the \glsxtrshort{lora}-Ensemble, see Appendix~\ref{app:model_size_study}. 
Lastly, the LoRA-Ensemble remains effective even when pre-training is not available; see Appendix~\ref{app:finetuned_same_dataset} for the experiment where it was pre-trained and fine-tuned on the same target dataset.

\textbf{Comparison to Bayesian LoRA.}
On the language modeling task (SST-2 sentiment classification), LoRA-Ensemble consistently outperforms Bayes-LoRA~\citep{yang2024bayesian} across accuracy, F1, NLL, and Brier score, with Bayes-LoRA only achieving a marginally lower ECE.
Refer to Tab.~\ref{tab_app:model_performance_Bert} and Appendix~\ref{app:sst2}.
%
%This behavior is consistent with Laplace methods, which often improve calibration but sacrifice predictive accuracy or fall short compared to ensembles~\citep{daxberger2021laplace}. 
%
This result is consistent with previous works, showing that Laplace-based methods improve calibration at the expense of predictive accuracy~\citep{deng2022accelerated} or often fall short compared to ensembles~\citep{daxberger2021laplace,eschenhagen2021mixtures}. 
Notably, in our experiments, LoRA-Ensemble is more than 10x faster at inference, demonstrating that it combines strong accuracy, reliable calibration, and efficiency. In contrast, the Laplace-based Bayesian method trades accuracy for improved uncertainty estimates while remaining less efficient. 
Refer to Appendix~\ref{app:bayes_details} for more details and discussion.

\textbf{Practical Guidance.} Our results suggest that the LoRA rank should be kept low unless accuracy saturates, as unnecessarily high ranks may degrade calibration. Calibration behavior also varies with ensemble size and dataset difficulty: on simpler datasets, increasing ensemble size can induce under-confidence, while on harder datasets it improves calibration. In cases of under-confidence, simple post-hoc temperature scaling is effective (Appendix~\ref{app_sec:temp_scaling}). Practical hyperparameter guidelines are summarized in Appendix~\ref{subsec:sensitivity}.

\textbf{Limitations \& Future Work.}
Despite its memory-efficient design and reduced per-member training and inference overhead, our ensembling approach maintains computational complexity similar to that of conventional ensembles, since each batch still requires separate forward passes for every member. While LoRA-Ensemble is directly applicable to transformers with standard linear attention projections, its behavior in the context of more complex variants -- such as mixture-of-experts (MoE) models \citep{Jiang2024MixtralExperts} or architectures with non-standard attention parameterizations -- remains an open question and a promising direction for future work.
%
%Our approach relies on a fixed set of LoRA-adapted models a future direction is to integrate Bayesian LoRA~\citep{wang2024blob, yang2024bayesian} with ensembling, enabling posterior sampling of unbounded members that combine efficiency with principled uncertainty estimation.
%
%Our work is primarily empirical, and a promising direction for future research is to develop theoretical justification for why LoRA-Ensemble improves calibration and diversity, for instance by analyzing their behavior through the lens of the bias–variance trade-off or PAC-Bayesian generalization bounds.
%
%In addition, a
As discussed by \citet{Rahaman2020UncertaintyEnsembles}, our work also suggests that in a high-parameter regime, deep ensembles may not exhibit the same behavior as they do in a low-parameter regime, where they typically improve calibration properties. This type of phase shift in the bias-variance trade-off, the so-called Double Descent Phenomenon, has previously been observed for large neural networks \citep{Nakkiran2021DeepHurt}. It would be valuable to conduct an in-depth analysis of how deep ensembles behave in high-parameter regimes.

\section{Related Work}
\paragraph{Estimation of Epistemic Uncertainty.}
A lot of work has gone into estimating the epistemic uncertainty in Artificial Neural Networks (ANN). As the analytical computation of the posterior in such models is generally intractable, methods for approximate Bayesian inference have been proposed. Such methods rely on imposing an appropriate prior on the weights and using the likelihood of the training data to get an approximate posterior of the weight space.

The main techniques are, on the one hand, Variational Inference  \citep{Graves2011PracticalNetworks, Ranganath2014BlackInference}, which \citep{Blundell2015WeightNetworks} have specialized to neural networks as \emph{Bayes by Backprop}. And on the other hand variants of \glsxtrfull{mcmc} \citep{Neal1996BayesianNetworks, Chen2014StochasticCarlo}, including \glsxtrfull{sgld} \citep{Welling2011BayesianDynamics}. These, however, are often not able to accurately capture high-dimensional and highly non-convex loss landscapes, like the ones usually encountered in deep learning \citep{Gustafsson2019EvaluatingVision}.
More recently, Bayesian LoRA methods have been explored, with \citet{yang2024bayesian} using a Laplace approximation for improved calibration and \citet{wang2024blob} jointly learning mean and covariance during fine-tuning.
An alternative line of work focuses on single-model uncertainty estimation. Spectral-normalized Neural Gaussian Process (SNGP)~\citep{liu2020simple} combines spectral normalization with a Gaussian Process output layer to achieve distance-aware uncertainty. However, SNGP's bi-Lipschitz assumptions do not hold for transformers, as dot-product self-attention has an unbounded Lipschitz constant. To address this, \citet{kim2021lipschitz} propose L2 Self-Attention and \citet{ye2023lrformer} introduce LRFormer, both designed to satisfy Lipschitz constraints in transformer architectures.
Epistemic Neural Networks (EpiNet)~\citep{osband2022epistemic} take a different approach by augmenting a base network with a small auxiliary network that takes learnable epistemic indices as input, producing diverse predictions without maintaining multiple full models.

\paragraph{Ensembles and Implicit Ensembling.}
\citet{Lakshminarayanan2017SimpleEnsembles} have proposed a method known as deep ensembles. It uses a set of neural networks with identical architecture that are independently and randomly initialized, and (as usual) trained with variants of \glsxtrfull{sgd}. While the latter introduces further stochasticity, \citet{Fort2019DeepPerspective} have shown that the initialization of the weights is more important to explore the admissible weight space. Ensemble members will generally converge to different modes of the loss function, such that they can be considered Monte Carlo samples of the posterior distribution \citep{Wilson2020BayesianGeneralization, Izmailov2021WhatLike}. While ensembles, in general, yield the best results in terms of accuracy and uncertainty calibration, a straightforward implementation suffers from high memory and compute requirements, since multiple instances of the full neural network must be trained and stored. This can become prohibitive for modern neural networks with many millions, or even billions, of parameters.

Consequently, researchers have attempted to find ways of mimicking the principle of deep ensembles without creating several full copies of the base model. \citep{Gal2015DropoutLearning} have proposed \glsxtrlong{mcdropout}, where the posterior is approximated by sampling different dropout patterns at inference time. While this is less expensive in terms of memory, performance is often worse. Masksembles \citep{Durasov2020MasksemblesEstimation} are a variant that attempts to select suitable dropout masks in order to obtain better uncertainty estimates.
Snapshot Ensembles \citep{Huang2017SnapshotFree} use cyclic learning rates to steer the learning process such that it passes through multiple local minima, which are then stored as ensemble members. This reduces the training effort but does not address memory requirements or inference time.

Particularly relevant for our work are attempts that employ a shared backbone and modify only selected layers. \citep{Havasi2020TrainingPrediction} follow that strategy, in their case only the first and last layer of a neural network are replicated and trained independently to emulate an ensemble. %\textcolor{blue}{EpiNets \citep{osband2022epistemic} similarly attach an auxiliary epistemic network to a shared backbone to induce function-space diversity without replicating the full model.}
Packed-Ensemble~\citep{packed} leverage grouped convolutions to train lightweight ensembles within a single shared backbone.
Batch-Ensemble \citep{Wen2020BatchEnsemble:Learning} is similar to \glsxtrshort{lora}-Ensemble in that it also uses low-rank matrices to change the model parameters. More specifically, shared weight matrices are modulated by element-wise multiplication with different rank-1 matrices to achieve the behavior of a deep ensemble while adding only a small number of parameters. \citet{Wenzel2020HyperparameterQuantification} take this concept further by also ensembling over different hyper-parameter settings. \citet{Turkoglu2022FiLM} freeze all weights of the base model and instead vary the feature-wise linear modulation \cite[FiLM,][]{Li2018AdaptiveAdaptation, Takeda2021TrainingModulation}.
A related concept was recently introduced for \glsxtrshortpl{llm}: the Mixtral of Experts model \citep{Jiang2024MixtralExperts} averages over a sparse mixture of experts to efficiently generate text.

\paragraph{Low-Rank Adaptation in Transformer Networks.}
\glsxtrlong{lora} was originally conceived as a parameter-efficient way of fine-tuning \glsxtrfullpl{llm} \citep{Hu2021LoRA:Models}. It is based on the observation that, while modern neural networks have huge parameter spaces, the solutions they converge to have much lower intrinsic dimension \citep{Li2018AdaptiveAdaptation, Aghajanyan2020IntrinsicFine-Tuning}. \glsxtrshort{lora} exploits this and \citet{Hu2021LoRA:Models} show that even when fine-tuning only low-rank update matrix $B\!\cdot\!A$ (sometimes with rank as low as one or two), the resulting models are competitive with much more expensive fine-tuning schemes. The method quickly became popular and has since also been extended with weight-decomposition \citep{Liu2024DoRA:Adaptation}. The \glsxtrfull{lora} idea has been applied in various fields, notably for denoising diffusion models \citep{Luo2023LCM-LoRA:Module, Golnari2023LoRA-EnhancedModels}.
As we have shown, the \glsxtrshort{lora} adaptation mechanism naturally lends itself to parameter-efficient ensembling, which we investigate in the context of uncertainty calibration, with a primary focus on vision transformers but not limited to them. A similar idea has concurrently been explored for fine-tuning LLMs~\citep{Wang2023LoRAFine-tuning}, yielding promising results in both predictive performance and uncertainty estimation.

\section{Conclusion}
We have presented \glsxtrshort{lora}-Ensemble, a novel, parameter-efficient method for probabilistic learning that is tailored to the transformer architecture (and potentially other architectures that make use of the attention mechanism). \glsxtrshort{lora}-Ensemble uses a simple, but efficient trick to turn a single base model into an implicit ensemble: the weights of the base model are kept frozen, but are modulated with the \glsxtrlong{lora} mechanism. 
%
%By training multiple, stochastically varying instances of the low-rank matrices that define the modulation one obtains a diverse set of ensemble members that share most of the weights (namely those of the base model) and add only a small overhead (the coefficients of their individual low-rank matrices).
%
By training multiple, stochastically varying instances of the low-rank matrices that define the modulation, one obtains a diverse set of ensemble members that share the majority of their weights and introduce only minimal overhead through the coefficients of their individual low-rank matrices.
%
%\textcolor{green}{Our experiments, conducted across four computer vision classification tasks, a sound classification task, a natural language processing task, and an out-of-distribution (OOD) detection task, demonstrate that the proposed approach excels in both classification performance and uncertainty calibration. Not only does it surpass other implicit ensembling state-of-the art methods, but it also outperforms explicit ensembles on many tasks. This challenges the prevailing notion in the literature that explicit ensembles represent the upper bound of efficient ensembling methods, as suggested in \citep{Wen2020BatchEnsemble:Learning}.}
%
Our extensive experiments demonstrate that the proposed approach excels in both predictive performance and uncertainty calibration. Not only does it surpass other state-of-the-art implicit ensembling methods, but it also outperforms Explicit Ensembles on many tasks. This challenges the prevailing notion in the literature that Explicit Ensembles represent the upper bound for efficient ensembling methods~\citep{Wen2020BatchEnsemble:Learning}.

%Our experiments on two different computer vision tasks, a sound classification task, and an OOD detection task show that the proposed approach can outperform other, implicit as well as explicit, ensembling strategies in terms of both classification performance and uncertainty calibration.
%

%\paragraph{Broader Impact}
%
%In recent years, the size of machine learning models has expanded rapidly. GPT-3 \citep{Brown2020LanguageLearners} has 175 billion parameters, while its successor, GPT-4, is rumored to contain over 1.7 trillion parameters, with training costs exceeding \$100 million. As the trend toward larger models continues, growing computational resources are required.
%
%With this work, \glsxtrshort{lora}-Ensemble aims to contribute to more efficient ensemble methods, considering the resource usage and environmental impact of AI models. This effort strives for more sustainable practices, advancing the concept of "Green AI."

\section{Broader Impact}
Better-calibrated uncertainty estimates allow users to recognize unreliable model predictions and treat them with the necessary caution, thus improving decision-making in high-stakes settings. We point out that good calibration on benchmark datasets does not, by itself, guarantee safe or reliable behavior in real-world deployment: uncertainty may degrade under distribution shift, so external validation for the intended use case is advised before employing any uncertainty estimation method in practice.

More broadly, as models grow in size, running ensembles of multiple model instances further increases computational cost and energy consumption. Efficient ensembling approaches like ours aim to preserve ensemble performance while reducing resource usage and the associated environmental impact, contributing to a more sustainable “Green AI”.

%The rapid growth in model sizes has driven substantial performance improvements in AI applications but also increased computational and energy costs. Ensemble methods, which combine multiple models to boost performance, can further multiply these demands. In this work, we introduce an efficient ensembling approach that maintains performance gains while reducing resource usage and the environmental impact of AI models. This effort strives for more sustainable practices, advancing the concept of “Green AI.”

%\clearpage

\bibliography{main}
\bibliographystyle{tmlr}

\clearpage

\appendix

\clearpage
\appendix

\section*{\Large Appendix}

%Make appendices start on new page
%\let\stdsection\section
%\renewcommand\section{\clearpage\stdsection}

\addcontentsline{toc}{part}{Appendix}
\etocsettocdepth{subsection}
\localtableofcontents
\renewcommand{\thesection}{\Alph{section}}  % A, B, C…
\setcounter{section}{0}
\clearpage

\section{Additional Experiments and Results}\label{app_A:more_experiment}

This section presents comprehensive experimental results for the newly introduced CIFAR-10 dataset, Environmental Sound Classification on ESC-50. It also includes additional results for the CIFAR-100 and HAM10000 datasets and expanded baseline comparisons for CIFAR-100.

\subsection{CIFAR-10}
The results for the CIFAR-10 dataset, as shown in Tab.~\ref{tab_app:model_performance_CIFAR10}, indicate that \glsxtrshort{lora}-Ensemble outperforms all other methods. Close behind is a single network enhanced with \glsxtrshort{lora}. This mirrors the results found in the main paper for CIFAR-100, with the exception of the calibration for a single model. It is important to note that although all methods achieve high accuracy and the differences between them are minimal, calibration is nearly perfect for most approaches. This suggests that the CIFAR-10 dataset is relatively easy for modern transformer models, and the results should not be over-interpreted. Nevertheless, the consistent performance across different random seeds suggests that the ranking is likely significant. Given the balanced nature of the CIFAR-10 dataset, the accuracy and F1-score are almost identical.

\begin{table}[h]
    \small
    \caption{Performance on the CIFAR-10 dataset for all compared methods. Ensembles have 16 members. Best score for each metric in \textbf{bold},  second-best \underline{underlined}.}
    \centering
    \begin{tabular}{lccccc}
    \toprule
    \textbf{Method} & \textbf{Accuracy ($\uparrow$)} & \textbf{F1 ($\uparrow$)} & \textbf{ECE ($\downarrow$)} & \textbf{NLL ($\downarrow$)} & \textbf{Brier ($\downarrow$)} \\
    \midrule
    Single Network & $92.8\pm0.1$ & $92.8\pm0.1$ & $0.051\pm0.001$ & $0.333\pm0.003$ & $0.120\pm0.002$ \\
    Single Net w/ LoRA & $\underline{94.5}\pm0.0$ & $\underline{94.5}\pm0.0$& $\underline{0.009}\pm0.001$ & $\underline{0.163}\pm0.002$ & $\underline{0.082}\pm0.001$\\
    MC Dropout & $92.9\pm0.2$ & $92.9\pm0.2$& $0.023\pm0.002$ & $0.260\pm0.005$ & $0.110\pm0.003$\\
    
    Snapshot Ensemble & $93.1\pm0.1$ & $93.1\pm0.1$& $0.037\pm0.002$ & $1.062\pm0.021$ & $0.510\pm0.008$\\
     Batch-Ensemble & $88.5\pm0.1$ & $88.5\pm0.1$& $0.048\pm0.001$ & $0.347\pm0.001$ & $0.172\pm0.000$\\
    Explicit Ensemble & $94.1\pm0.1$ & $94.1\pm0.1$& $0.031\pm0.001$ & $0.181\pm0.002$ & $0.087\pm0.001$\\
    
    \midrule
    LoRA-Ensemble & $\textbf{95.9}\pm0.1$ & $\textbf{95.9}\pm0.1$& $\textbf{0.003}\pm0.001$ & $\textbf{0.128}\pm0.001$ & $\textbf{0.064}\pm0.000$\\
    \bottomrule
    \end{tabular}
    \label{tab_app:model_performance_CIFAR10}
\end{table}

%CIFAR-10 results are given in Tab.~\ref{tab_app:model_performance_CIFAR10}. \glsxtrshort{lora}-Ensemble performs best in all metrics on this dataset, followed by a single network augmented with \glsxtrshort{lora}. This aligns with the findings presented in the main paper for CIFAR-100, except for the calibration of a single model. Note, that all evaluated methods reach high accuracy and the differences are small, while calibration is near perfect for most approaches. I.e., the dataset is too easy for contemporary transformer models and one should not over-interpret the results. Nonetheless, the high stability across different random seeds indicates that the ranking in all likelihood is significant. Since the CIFAR-10 dataset is balanced, the accuracy and F1-score are nearly identical.

\subsection{ESC-50 Environmental Sound 
Classification}\label{app_sec:ESC_50_results}
Like for the \glsxtrshort{vit} model, we train an \glsxtrlong{ast} version of \glsxtrshort{lora}-Ensemble by modifying the attention layers with different sets of \glsxtrshort{lora} weights. That ensemble is then compared to a single instance of \glsxtrshort{ast} with and without \glsxtrshort{lora}, to an Explicit Ensemble of \glsxtrshort{ast}-models, and to an \glsxtrshort{mcdropout} variant of \glsxtrshort{ast}, similar to \citep{li2023dropkey}. For ESC-50 a \glsxtrshort{lora} rank of 16 worked best, presumably due to the larger domain gap between (image-based) pre-training and the actual audio classification task.
The experimental evaluation in \citep{Gong2021AST:Transformer} employs the same performance metrics as before, but a slightly different evaluation protocol. %, which we follow. 
Model training (and evaluation) is done in a 5-fold cross-validation setting, where the epoch with the best average accuracy across all five folds is chosen as the final model. The performance metrics given below are calculated by taking the predictions of all five folds at the chosen epoch and evaluating 
%accuracy and \glsxtrshort{ece} jointly. 
accuracy and calibration metrics jointly.
While the accuracy calculated this way is equivalent to the average of all five folds, 
%\glsxtrshort{ece} 
%
others are not, so this method results in a more realistic calculation of the calibration metrics.

%Consequently, the performance metrics given below \citep[and in][]{Gong2021AST:Transformer} are the average classification accuracy and \glsxtrshort{ece} across all five folds, taken at the best epoch.

%
%To go beyond computer vision tasks, \glsxtrshort{lora}-Ensemble is also applied to an audio dataset, using the \glsxtrlong{ast} as the backbone model. 
%
%Performance is again evaluated in terms of accuracy and \glsxtrlong{ece}. 
%
The results are summarized in Tab.~\ref{tab:model_performance_ESC50}.
On this dataset \glsxtrshort{lora}-Ensemble does not significantly outperform the Explicit Ensemble, but still matches its performance with much lower computational demands, see Appendix~\ref{app_sec:resource_ast}. Accuracy is insignificantly lower, whereas calibration is slightly better in terms of ECE. We note that, remarkably, the weights used in the transformer modules and for creating patch embeddings were pre-trained on images rather than audio streams.

%\begin{table}
%    \caption{Model performance on the ESC-50 dataset for the compared methods. Ensembles have 8 members due to memory limitations. Best score for each metric in \textbf{bold},  second-best \underline{underlined}.}
%    \centering
%    \begin{tabular}{lccccc}
%        \toprule
%        \textbf{Method} & \textbf{Accuracy ($\uparrow$)} & \textbf{F1 ($\uparrow$)} & \textbf{ECE ($\downarrow$)} & \textbf{NLL ($\downarrow$)} & \textbf{Brier ($\downarrow$)} \\
%        \midrule
%        Single Network & $89.3\pm0.7$ & $\pm$ &$0.040\pm0.005$& $\pm$ & $\pm$ \\
%        Single Net w/ LoRA & $88.1\pm0.1$ & $\pm$& $0.039\pm0.002$ & $\pm$  & $\pm$\\
%        MC Dropout & $89.4\pm0.3$ & $\pm$ & $0.087\pm0.003$&$\pm$ &$\pm$\\
%        Explicit Ensemble & $\textbf{91.3}\pm0.3$ & $\pm$& $\underline{0.026}\pm0.003$&$\pm$ & $\pm$ \\
%        \midrule
%        LoRA-Ensemble & $\underline{91.0}\pm0.2$ & $\pm$& $\textbf{0.022}\pm0.003$&$\pm$ &$\pm$  \\
%        \bottomrule
%    \end{tabular}
%    \label{tab:model_performance_ESC50}
%\end{table}

%\bottomrule

\begin{table}
    \caption{Model performance on the ESC-50 dataset for the compared methods. Ensembles have 8 members due to memory limitations. Best score for each metric in \textbf{bold},  second-best \underline{underlined}.}
    \centering
    \resizebox{1.0\linewidth}{!}{
    \begin{tabular}{lccccc}
        \toprule
        \textbf{Method} & \textbf{Accuracy ($\uparrow$)} & \textbf{F1 ($\uparrow$)} & \textbf{ECE ($\downarrow$)} & \textbf{NLL ($\downarrow$)} & \textbf{Brier ($\downarrow$)} \\
        \midrule 
        Single Network  & $89.6\pm0.7$ & $89.5\pm0.7$& $0.039\pm0.004$ & $0.410\pm0.020$ & $0.164\pm0.009$\\
        Single Net w/ LoRA & $88.0\pm0.3$ & $87.8\pm0.3$& $0.043\pm0.004$ & $0.461\pm0.019$ & $0.186\pm0.005$\\
        MC Dropout & $89.4\pm0.3$ & $89.3\pm0.4$& $0.087\pm0.005$ & $0.553\pm0.012$ & $0.176\pm0.005$\\
        Explicit Ensemble & $\textbf{91.3}\pm0.2$ & $\textbf{91.2}\pm0.3$& $\underline{0.027}\pm0.004$ & $\textbf{0.322}\pm0.004$ & $\textbf{0.133}\pm0.001$\\
        \midrule
        LoRA-Ensemble & $\underline{91.1}\pm0.2$ & $\underline{90.8}\pm0.2$& $\textbf{0.021}\pm0.003$ & $\underline{0.328}\pm0.004$ & $\underline{0.138}\pm0.001$\\
        \bottomrule
    \end{tabular}
    }
    \label{tab:model_performance_ESC50}
\end{table}

\subsection{CIFAR-100}\label{app_sec:CIFAR100_extra_diagrams}
\begin{figure}[th]
    \centering
    \begin{subfigure}[b]{.4\textwidth}
        \includegraphics[width=\textwidth]{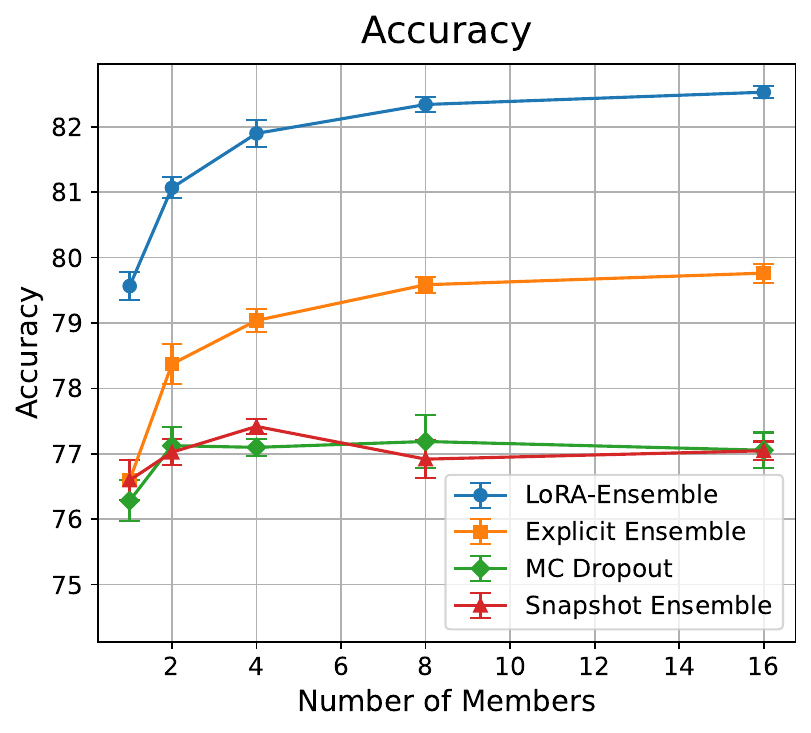}
        %\caption{Accuracy on the CIFAR-100 dataset.}
        %\label{subfig:CIFAR100_accuracy}
    \end{subfigure}
    %\hfill
    \hspace{.5cm}
    \begin{subfigure}[b]{.4\textwidth}
        \includegraphics[width=\textwidth]{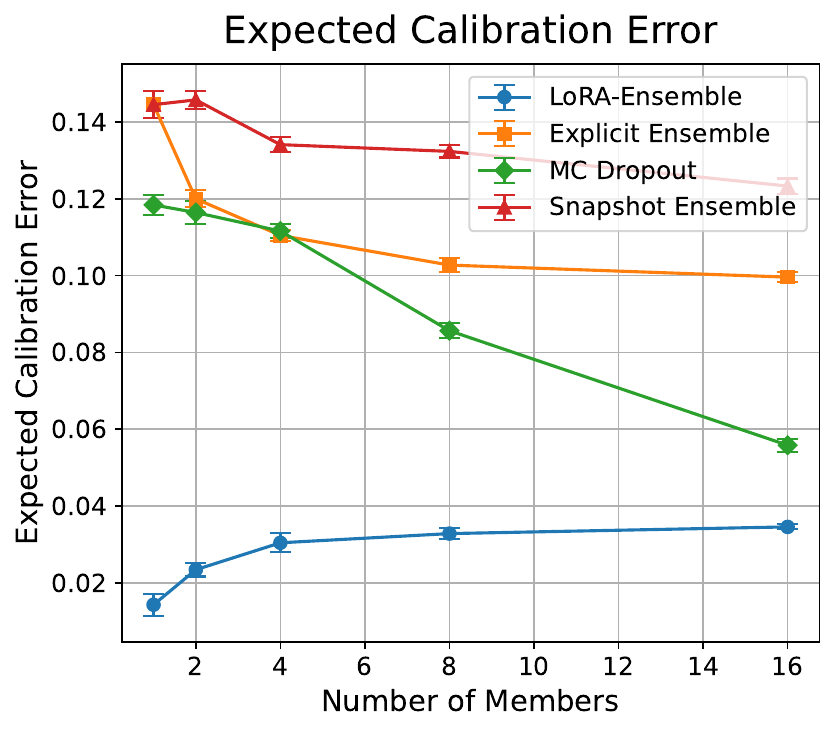}
        %\caption{ECE on the CIFAR-100 dataset.}
        %\label{subfig:CIFAR100_ece}
    \end{subfigure}
    \caption{Accuracy and \glsxtrlong{ece} on  CIFAR-100, with different ensemble sizes.}
    \label{fig:CIFAR100_results}
\end{figure}

\begin{figure}[th]
    \centering
    \begin{subfigure}[b]{.4\textwidth}
        \includegraphics[width=\textwidth]{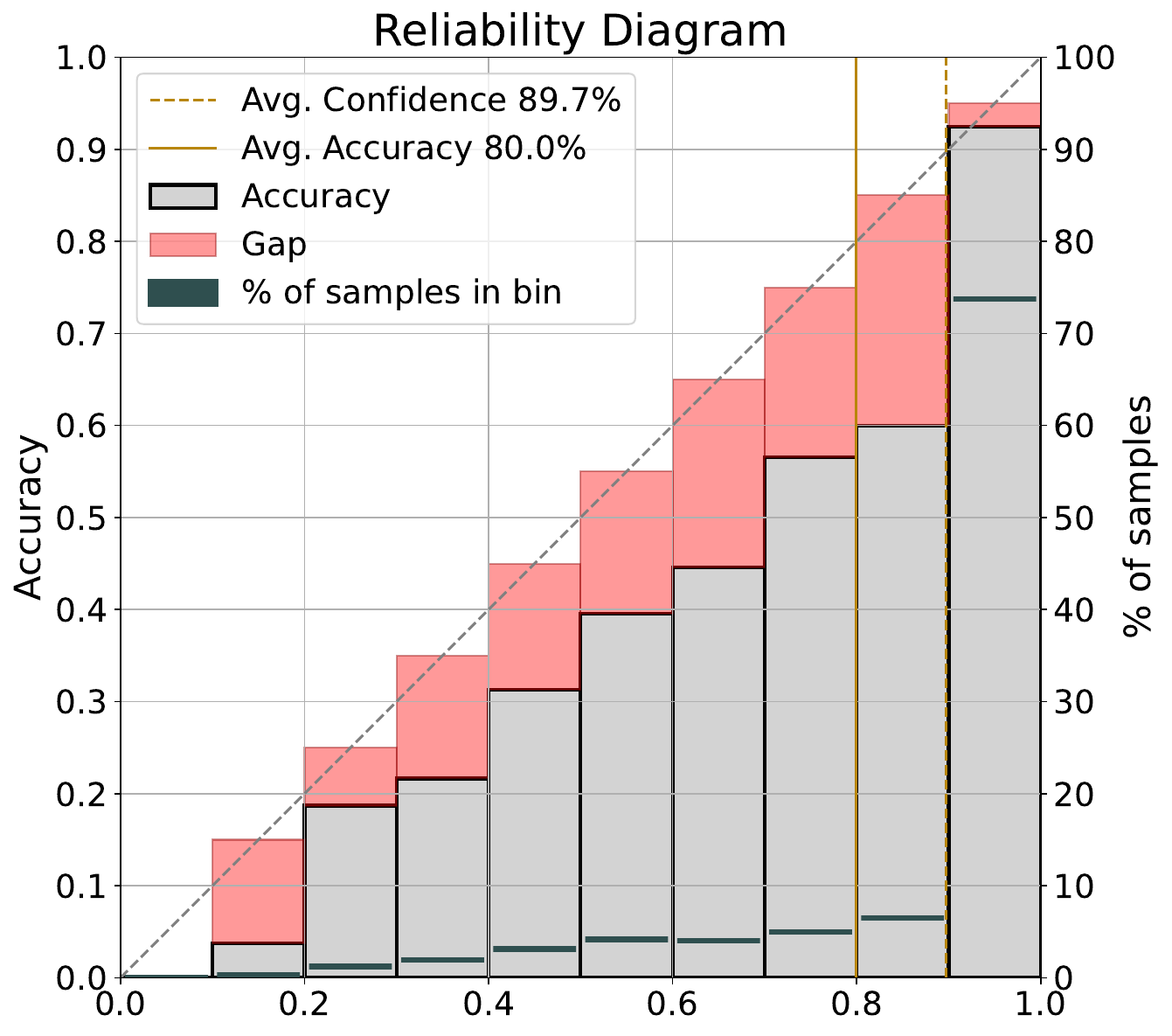}
    \end{subfigure}
    %\hfill
    \hspace{.7cm}
    \begin{subfigure}[b]{.4\textwidth}
        \includegraphics[width=\textwidth]{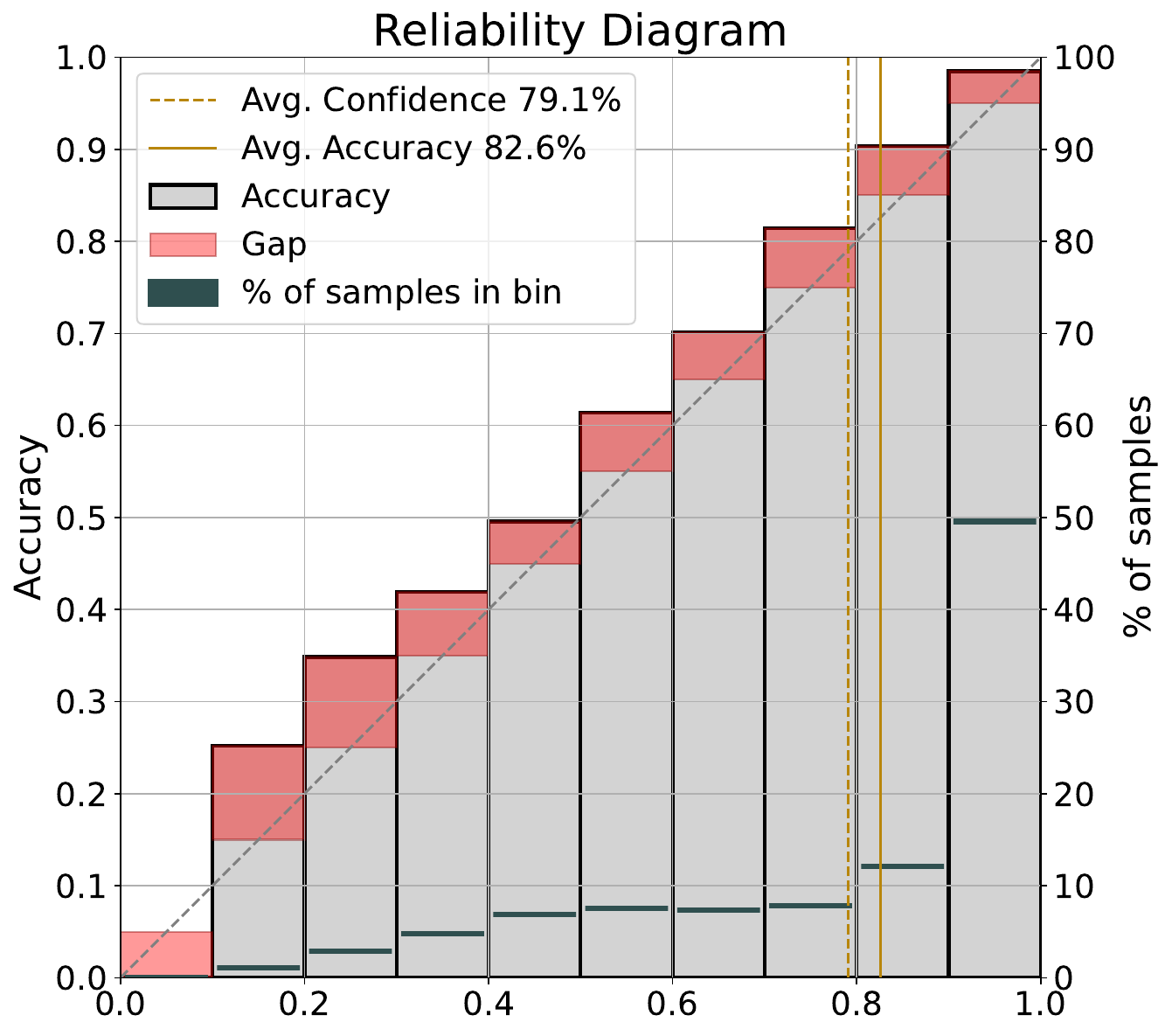}
    \end{subfigure}
 \caption{Reliability diagrams for Explicit Ensemble (left) and \glsxtrshort{lora}-Ensemble (right) with 16 members, on CIFAR-100.}
    \label{fig:reliability_diagram_lora_16_cifar100}
\end{figure}

Increasing the ensemble size of \Glsxtrshort{lora}-Ensemble on CIFAR-100 improves classification accuracy but reduces calibration, as illustrated in Fig.~\ref{fig:CIFAR100_results}. The reliability diagram in Fig.~\ref{fig:reliability_diagram_lora_16_cifar100} highlights this behavior: networks with \glsxtrshort{lora} on CIFAR-100 are generally under-confident, with accuracy exceeding predicted confidence. As observed by \citep{Rahaman2020UncertaintyEnsembles}, ensembling under-confident models can exacerbate this discrepancy, leading to poorer calibration metrics.

\subsection{HAM10000 Lesion Classification}\label{app_sec:HAM10000_extra_diagrams}
\begin{figure}[th]
    \centering
    \begin{subfigure}[b]{.4\textwidth}
        \includegraphics[width=\textwidth]{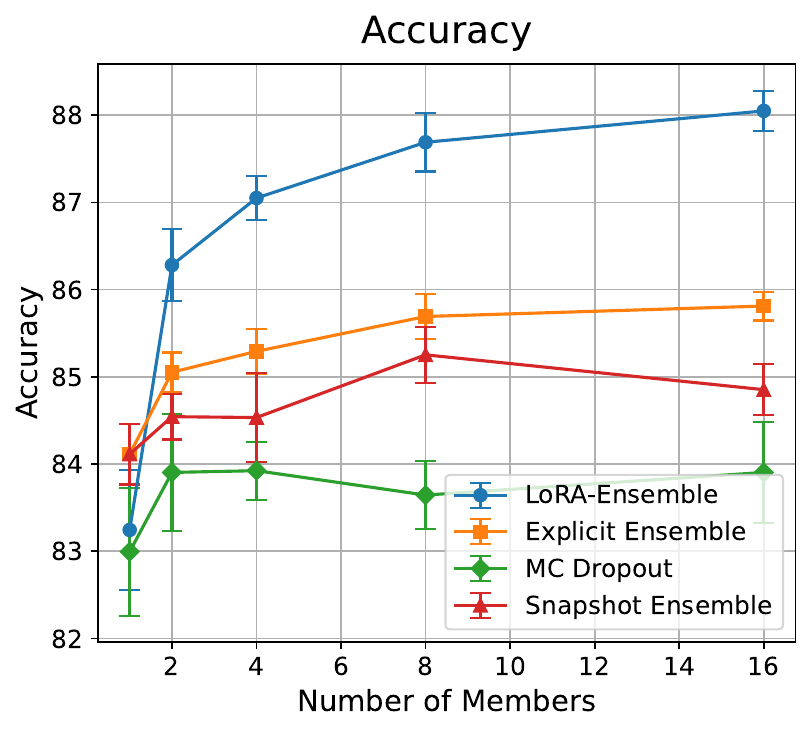}
        %\caption{Accuracy on the CIFAR-100 dataset.}
        %\label{subfig:ham_accuracy}
    \end{subfigure}
    %\hfill
    \hspace{.5cm}
    \begin{subfigure}[b]{.4\textwidth}
        \includegraphics[width=\textwidth]{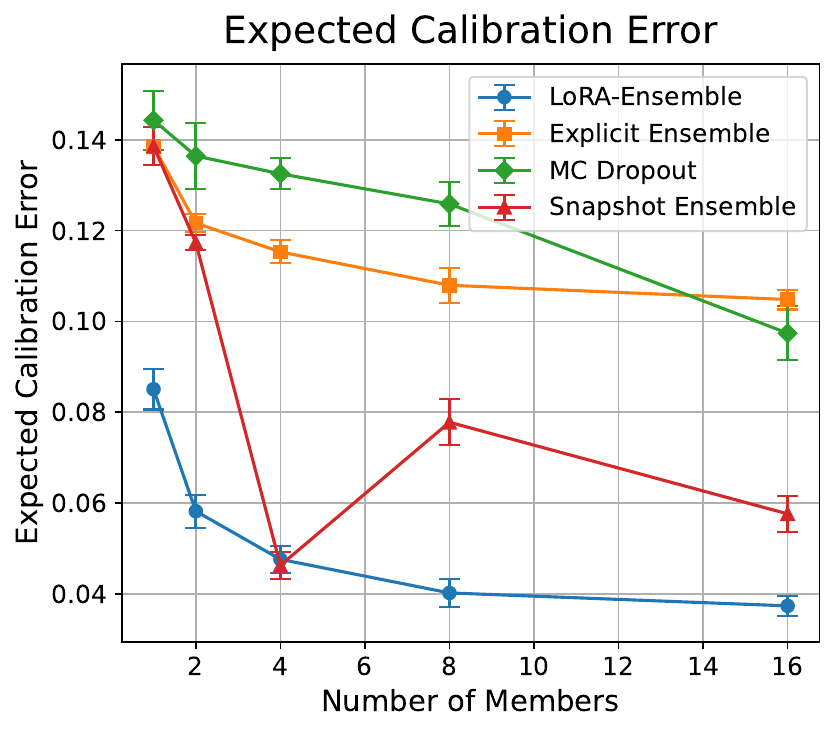}
        %\caption{ECE on the CIFAR-100 dataset.}
        %\label{subfig:ham_ece}
    \end{subfigure}
    \caption{Accuracy and \glsxtrlong{ece} on  HAM10000, with different ensemble sizes.}
    \label{fig:ham_results}
\end{figure}

\begin{figure}[th]
    \centering
    \begin{subfigure}[b]{.4\textwidth}
        \includegraphics[width=\textwidth]{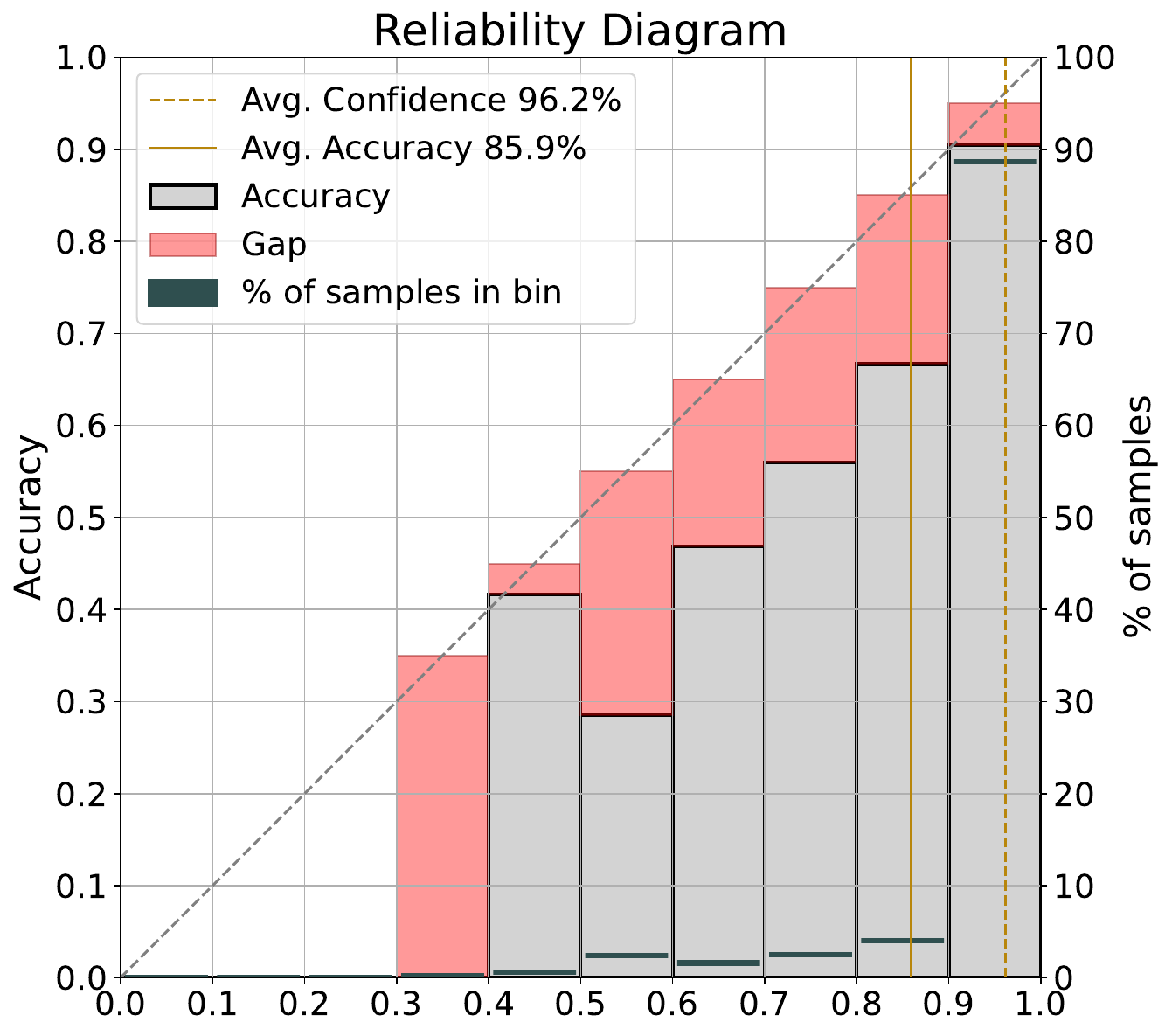}
        %\caption{Accuracy on the CIFAR-100 dataset.}
        %\label{subfig:ham_accuracy}
    \end{subfigure}
    %\hfill
    \hspace{.5cm}
    \begin{subfigure}[b]{.4\textwidth}
        \includegraphics[width=\textwidth]{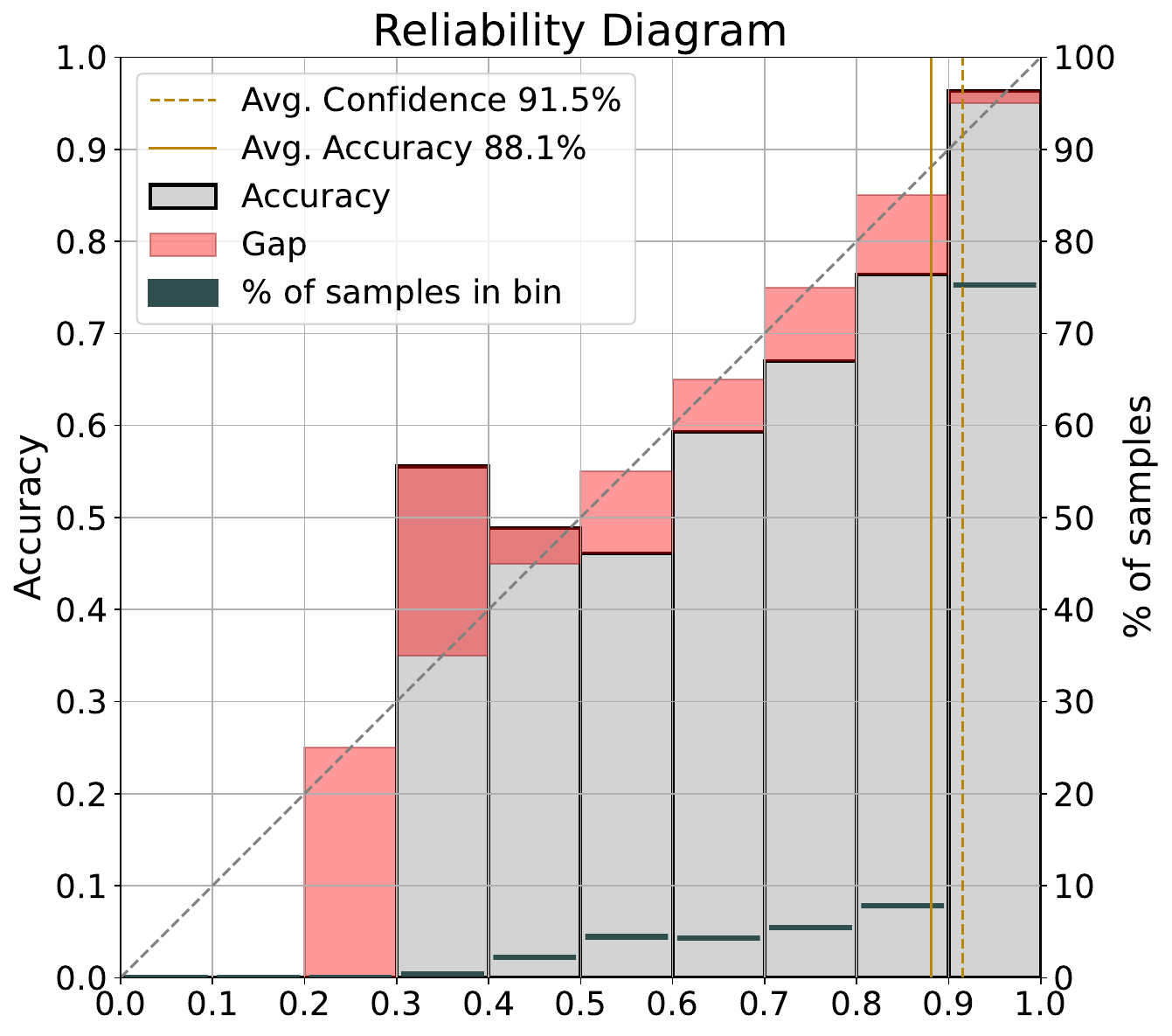}
        %\caption{ECE on the CIFAR-100 dataset.}
        %\label{subfig:ham_ece}
    \end{subfigure}
    \caption{Reliability diagrams for Explicit Ensemble (left) and \glsxtrshort{lora}-Ensemble (right) with 16 members, on HAM10000.}
    \label{fig:ham_results_reliability_diagramm}
\end{figure}

Classification accuracy and \glsxtrshort{ece} for HAM10000 dataset are both graphed against ensemble size in Fig.~\ref{fig:ham_results}. Again, \glsxtrshort{lora}-Ensemble outperforms all baselines for larger ensembles. In Fig.~\ref{fig:ham_results_reliability_diagramm} the reliability diagrams for \glsxtrshort{lora}-Ensemble and an Explicit Ensemble with 16 members each on the HAM10000 dataset are shown. Here, the models are overconfident, further supporting our reasoning regarding the surprising behaviour of calibration with growing ensemble size in the case of CIFAR-100.

\subsection{Comparison with a Single, High-rank LoRA Network}\label{app_sec:more_baselines}
We compare the proposed \glsxtrshort{lora}-Ensemble method with an additional baseline: a single high-rank \glsxtrshort{lora} model configured to have the same total number of trainable \glsxtrshort{lora} parameters as the \glsxtrshort{lora}-Ensemble. This evaluation is conducted on the CIFAR-100 classification task to examine the relative effectiveness of ensembling versus increasing parameter capacity within a single model.

Notably, as shown in Tab.~\ref{tab:model_performance_large_rank}, the high-rank \glsxtrshort{lora} model underperforms compared to the low-rank \glsxtrshort{lora} model. This result indicates that the performance gains of the \glsxtrshort{lora}-Ensemble are not solely due to an increased number of trainable parameters but are instead attributable to the ensembling approach.

\begin{table}
    \caption{Model performance on the CIFAR-100 dataset for the compared methods. Ensembles have 16 members. Best score for each metric in \textbf{bold},  second-best \underline{underlined}.}
    \centering
    \scriptsize % Reduce font size
    \resizebox{1.0\linewidth}{!}{
    \begin{tabular}{lccccccc}
        \toprule
        \textbf{Method} & Rank & Trainable params. &\textbf{Accuracy ($\uparrow$)} & \textbf{F1 ($\uparrow$)} & \textbf{ECE ($\downarrow$)} & \textbf{NLL ($\downarrow$)} & \textbf{Brier ($\downarrow$)} \\
        \midrule
        Single Net w/ LoRA & 8 & 666'724& $\underline{79.6}\pm0.2$ & $\underline{79.4}\pm0.2$& $\textbf{0.014}\pm0.003$ & $\underline{0.671}\pm0.005$ & $\underline{0.286}\pm0.003$\\
        Single Net w/ LoRA & 128 & 9'514'084 &$77.0\pm0.1$ & $77.0\pm0.1$& $0.080\pm0.001$ & $0.867\pm0.007$ & $0.332\pm0.002$\\
        \midrule
        LoRA-Ensemble & 8 & 10'667'584 &$\textbf{82.5}\pm0.1$ & $\textbf{82.5}\pm0.1$& $\underline{0.035}\pm0.001$ & $\textbf{0.587}\pm0.001$ & $\textbf{0.253}\pm0.000$\\
        \bottomrule
    \end{tabular}
    }
    \label{tab:model_performance_large_rank}
\end{table}

\subsection{INaturalist 2017 Large-Scale Fine-Grained Image Classification}
\begin{figure}[th]
    \centering
    \begin{subfigure}[b]{.4\textwidth}
        \includegraphics[width=\textwidth]{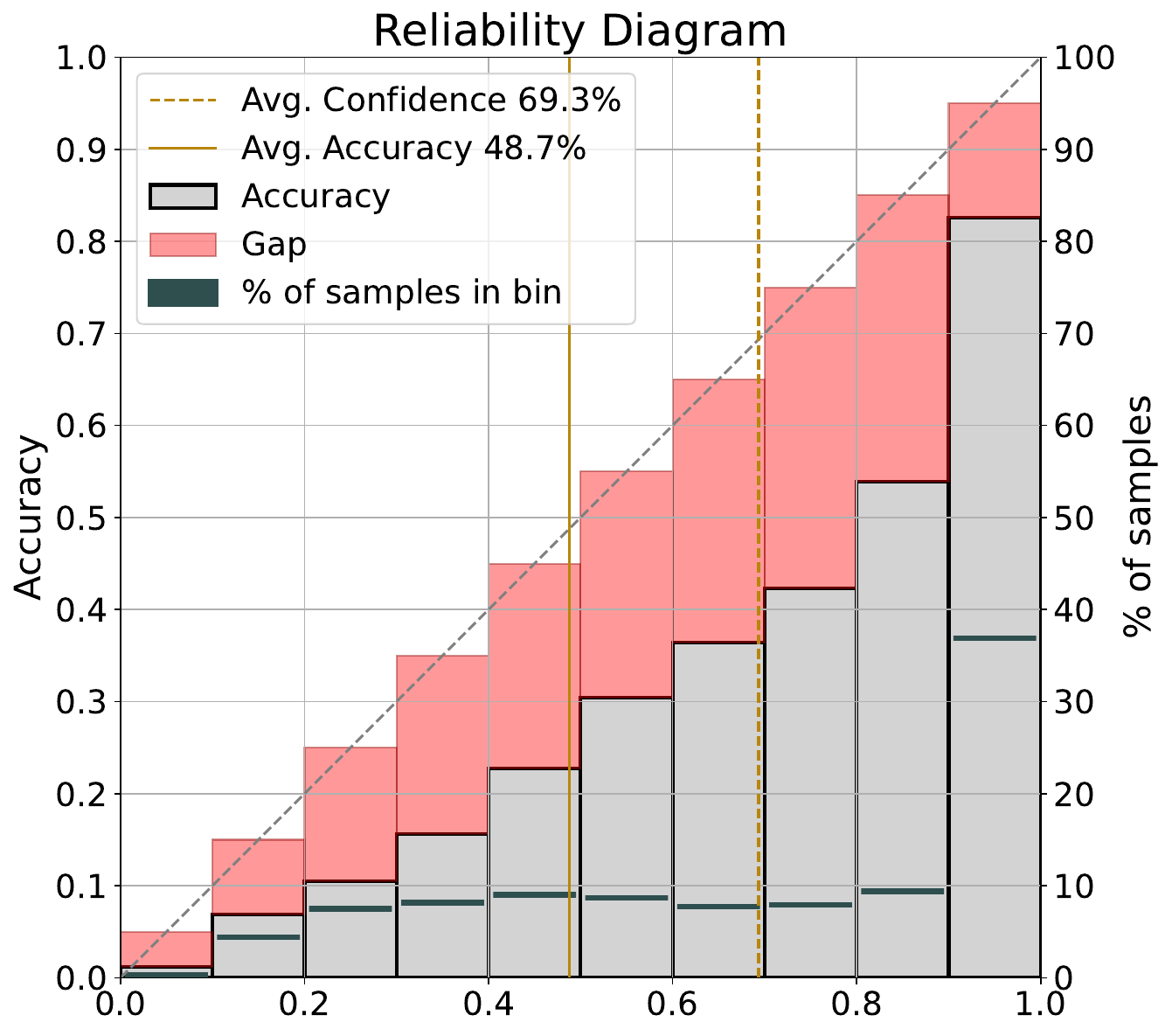}
    \end{subfigure}
    %\hfill
    \hspace{.7cm}
    \begin{subfigure}[b]{.4\textwidth}
        \includegraphics[width=\textwidth]{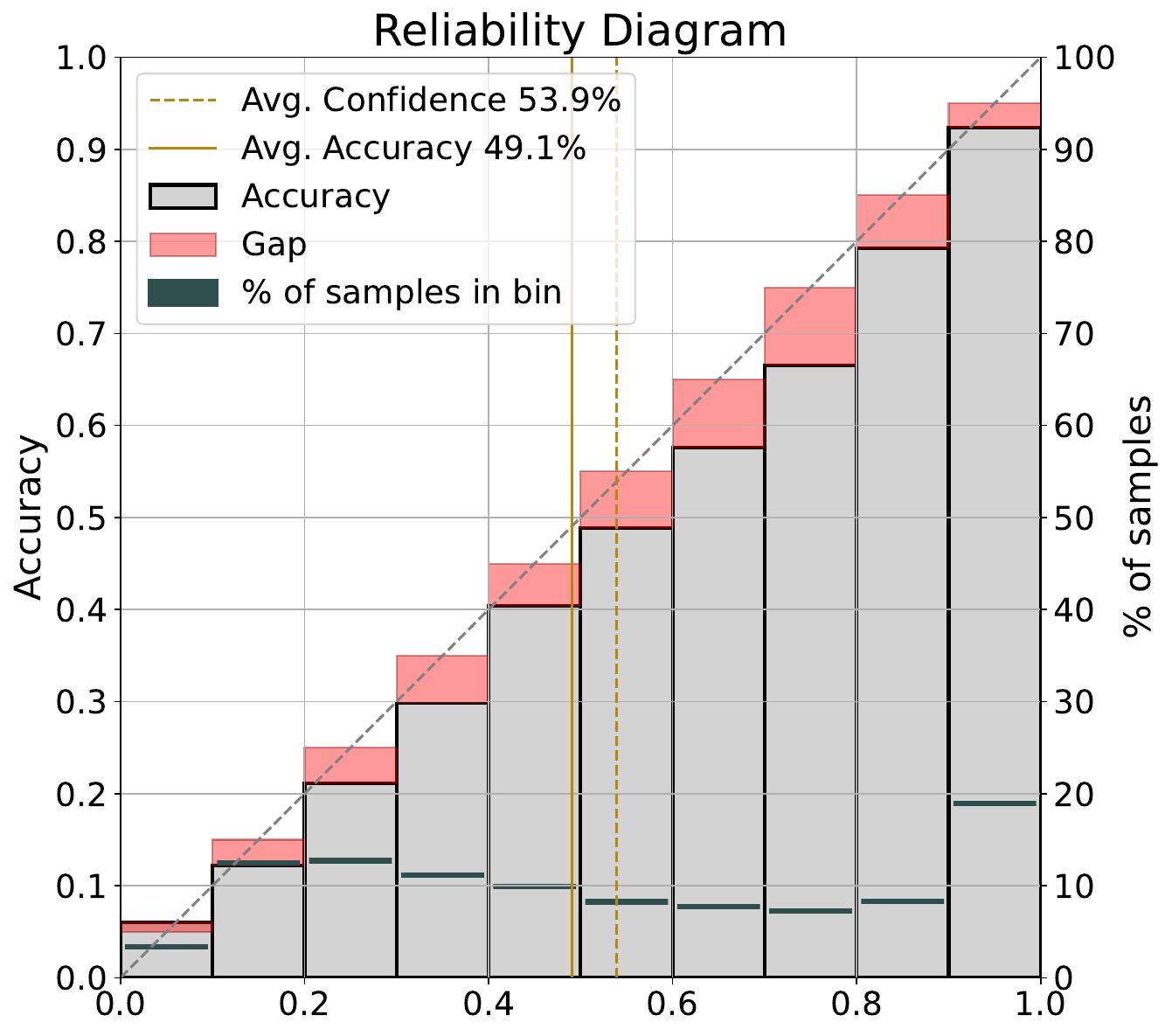}
    \end{subfigure}
 \caption{Reliability diagrams for Explicit Ensemble (left) and \glsxtrshort{lora}-Ensemble (right) with 4 members, on INat2017.}
    \label{fig:reliability_diagram_lora_4_inat2017}
\end{figure}

In Fig.~\ref{fig:reliability_diagram_lora_4_inat2017}, reliability diagrams for the iNat 2017 dataset are shown, once for \glsxtrshort{lora}-Ensemble and once for an Explicit Ensemble, both with 4 members. One can clearly see the over-confidence of the Explicit model, and the much improved uncertainty calibration of \glsxtrshort{lora}-Ensemble at almost the same accuracy (49.6\% vs.\ 49.3\%, c.f.\ Tab.~\ref{tab_app:model_performance_INat2017}).

\subsection{SST-2 Language Modeling for Sentiment Classification} \label{app:sst2}
To further illustrate the generality of our method, we extend the evaluation to also include language processing (NLP). Indeed, we find that \glsxtrshort{lora}-Ensemble also handles this very different modality and estimates well-calibrated uncertainties for language data.

We chose the Stanford Sentiment Treebank 2 (SST-2) dataset \citep{socher2013recursive} for the experiment, a widely used binary sentiment classification benchmark, and part of the GLUE benchmark suite \citep{Wang2018GLUEAM}. The model we use is the uncased BERT base model, which we fine-tune for SST-2.

%Table~\ref{tab_app:model_performance_Bert} presents the performance on the SST-2 validation set. The \glsxtrshort{lora}-Ensemble achieves superior calibration in terms of NLL and second best ECE after Bayes-LoRA (the difference is insignificant) and outperforms all baselines, including the Explicit Ensemble, while the latter has only a tiny edge of 0.5 percentage points in accuracy. In contrast, MC Dropout improves calibration over single models, but greatly compromises accuracy. This behavior is consistent with our other experiments, and in line with findings in the literature \citep{li2023dropkey}.
%, suggesting that further investigation and analysis could provide additional insights.
%A single LoRA-augmented model matches the calibration of its explicit counterpart but lags behind it in predictive performance.
%

Tab.~\ref{tab_app:model_performance_Bert} presents the performance on the SST-2 validation set. Among the methods, the LoRA-Ensemble shows strong overall performance: it achieves superior calibration in terms of negative log-likelihood (NLL), second-best ECE after Bayes-LoRA (with only a negligible difference), and outperforms all baselines including the Explicit Ensemble. The Explicit Ensemble holds only a marginal advantage of 0.5 percentage points in accuracy. In contrast, Monte Carlo Dropout improves calibration compared to single models but suffers from a substantial loss in accuracy, a pattern consistent with our other experiments and aligned with findings reported in the literature~\citep{li2023dropkey}.
A single LoRA-augmented model shows better calibration than a single model, but lags in accuracy.
Bayes-LoRA achieves competitive uncertainty calibration, obtaining the best ECE, but its NLL, Brier score, and accuracy are worse than those of both the Explicit Ensemble and the LoRA-Ensemble, and its accuracy is even lower than a single model, reflecting a trade-off where improved calibration comes at the expense of predictive performance. Refer to Appendix~\ref{app:bayes_details} for more details and discussion about Bayes-LoRA.

\begin{table}[th]
    \small
    \caption{Performance on the SST-2 validation dataset, evaluated using five different random seeds per model. Ensembles have 8 members. Best score for each metric in \textbf{bold},  second-best \underline{underlined}.}
    \centering
    \resizebox{1.0\linewidth}{!}{
    \begin{tabular}{lccccc}
    \toprule
    \textbf{Method} & \textbf{Accuracy ($\uparrow$)} & \textbf{F1 ($\uparrow$)} & \textbf{ECE ($\downarrow$)} & \textbf{NLL ($\downarrow$)} & \textbf{Brier ($\downarrow$)} \\
    \midrule
    Single Network & $92.5\pm0.2$ & $92.5\pm0.2$ & $0.064\pm0.003$ & $0.345\pm0.012$ & $0.136\pm0.003$ \\
    Single Net w/ LoRA & $91.6\pm0.5$ & $91.6\pm0.5$ & $0.059\pm0.005$ & $0.292\pm0.016$ & $0.148\pm0.008$ \\
    MC Dropout & $84.9\pm1.2$ & $84.9\pm1.3$ & $0.061\pm0.004$ & $0.364\pm0.020$ & $0.223\pm0.015$ \\
    
    Bayes-LoRA & $90.7\pm0.3$ & $90.7\pm0.3$ & $\textbf{0.036}\pm0.003$ & $0.247\pm0.005$ & $0.139\pm0.003$ \\

    Explicit Ensemble & $\textbf{93.2}\pm0.2$ & $\textbf{93.2}\pm0.2$ & $0.047\pm0.002$ & $\underline{0.234}\pm0.004$ & $\textbf{0.112}\pm0.003$ \\

    \midrule
    LoRA-Ensemble & $\underline{92.7} \pm0.2$ & $\underline{92.7}\pm0.2$ & $\underline{0.038}\pm0.003$ & $\textbf{0.208}\pm0.007$ & $\underline{0.114}\pm0.002$ \\
    %LoRA-Ensemble (new) & $92.9\pm0.2$ & $92.7\pm0.1$ & $0.041\pm0.002$ & $0.213\pm0.002$ & $0.116\pm0.002$ \\
    \bottomrule
    \end{tabular}
    }
    \label{tab_app:model_performance_Bert}
\end{table}

\subsection{Robustness to Distribution Shifts: CIFAR-10-C and CIFAR-100-C}\label{app_sec:distribution_shift}

Despite primarily evaluating the \glsxtrshort{lora}-Ensemble on in-distribution tasks, we also assess its robustness to out-of-distribution (OOD) inputs. A critical challenge arises when a model encounters data at test time that differs from the training distribution. If the model then produces poorly calibrated uncertainty estimates, this can lead to unsafe or unreliable predictions \citep{hendrycks2019robustness}.

To examine this, we evaluate our method on the CIFAR-10-C and CIFAR-100-C benchmark datasets. These datasets apply 19 distinct corruption types at five severity levels to the original CIFAR-10 and CIFAR-100 test sets \citep{hendrycks2019robustness}, introducing controlled distribution shifts, similar to prior work \citep{ovadia2019trust}. For this evaluation, we use pretrained models trained on the clean datasets with minimal data augmentation (only rotations), and assess both predictive performance and calibration.

\begin{figure}[ht]
    \centering
    \subfloat[]{\label{fig:}%
        \includegraphics[width=0.45\textwidth]{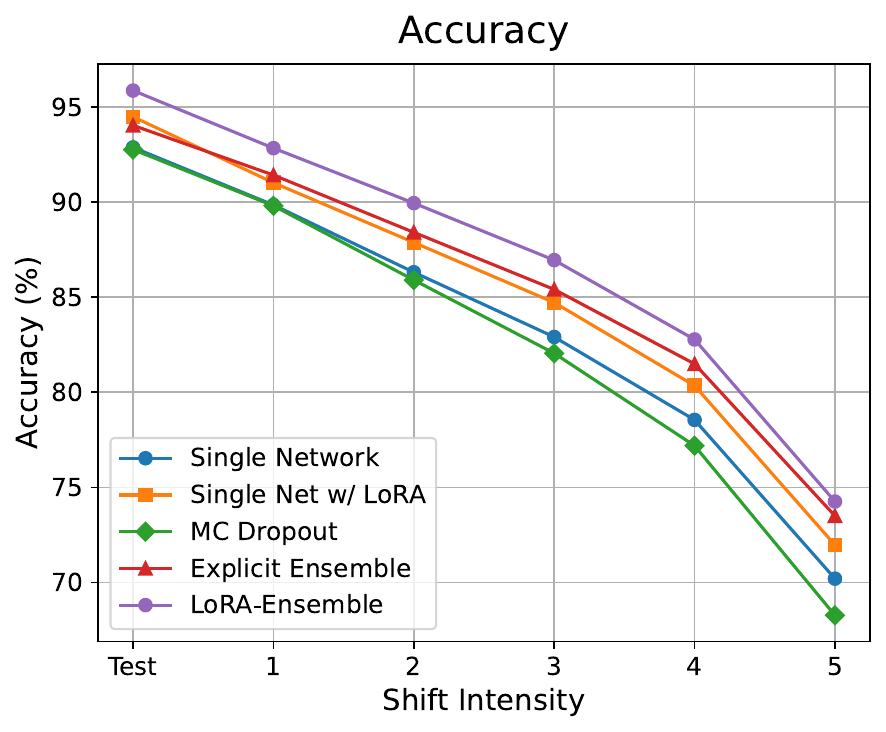}}%
    \hfill
    \subfloat[]{\label{fig:}%
        \includegraphics[width=0.45\textwidth]{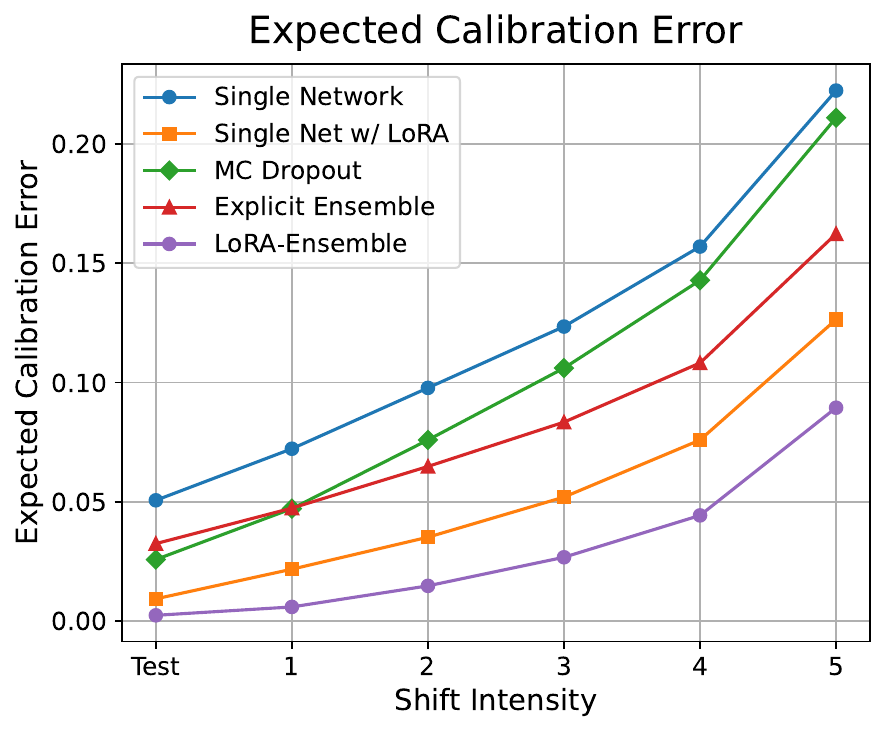}}\\
    
    \subfloat[]{\label{fig:}%
        \includegraphics[width=0.45\textwidth]{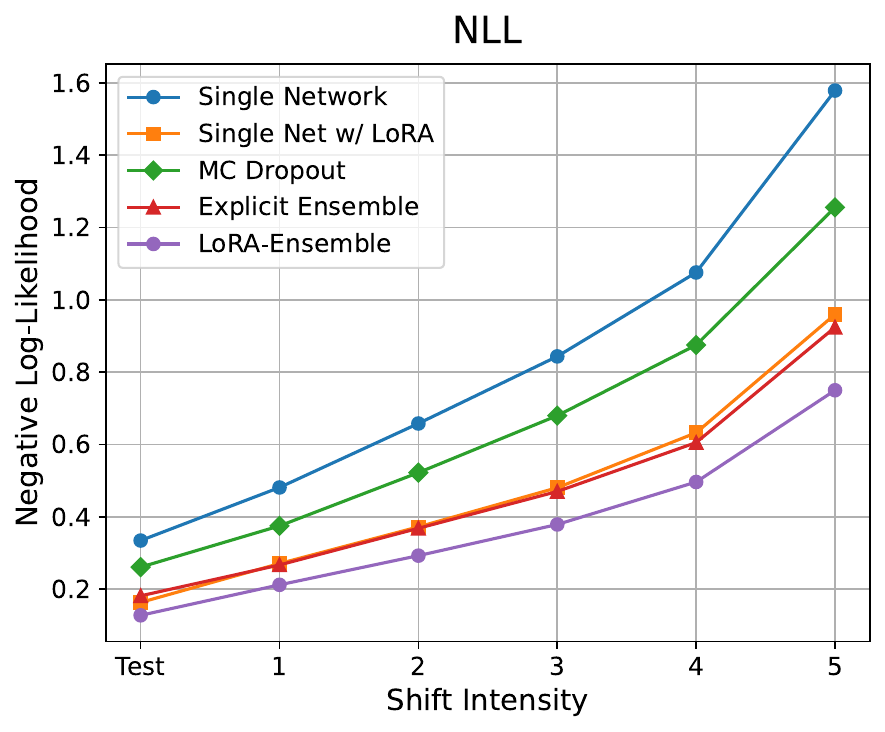}}%
    \hfill
    \subfloat[]{\label{fig:}%
        \includegraphics[width=0.45\textwidth]{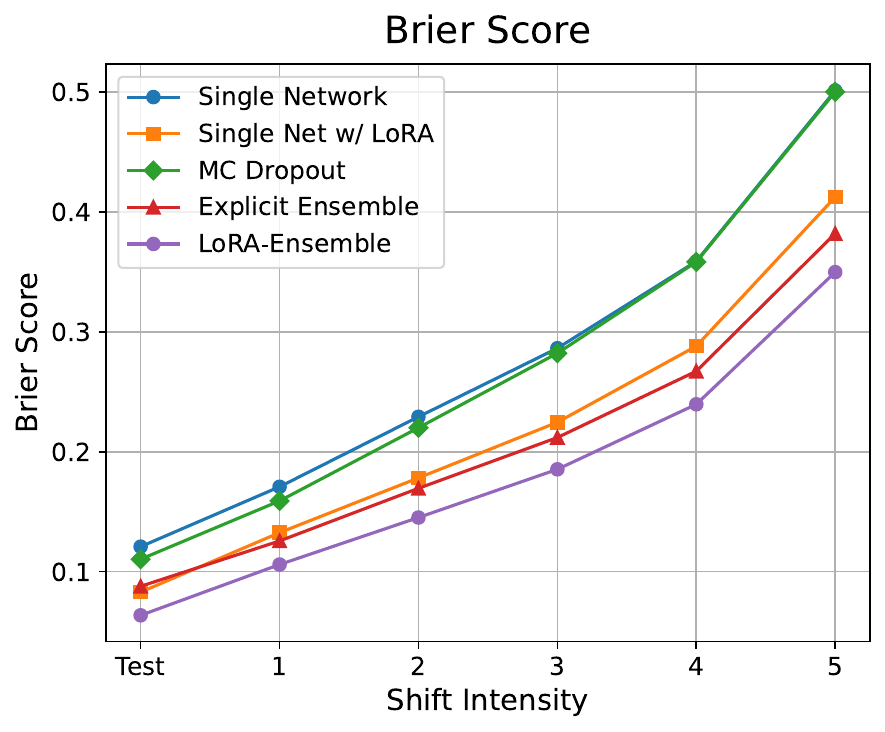}}%

    \caption{\glsxtrshort{lora}-Ensemble evaluated under varying levels of distribution shift on the CIFAR-10-C dataset. Each ensemble consists of 16 members. "Test" refers to the original CIFAR-10 test set, while the corrupted sets include test images subjected to 19 different augmentations at multiple severity levels, introducing distribution shifts.}

    \label{fig:results_CIFAR10C}
\end{figure}

\begin{figure}[ht]
    \centering
    \subfloat[]{\label{fig:}%
        \includegraphics[width=0.45\textwidth]{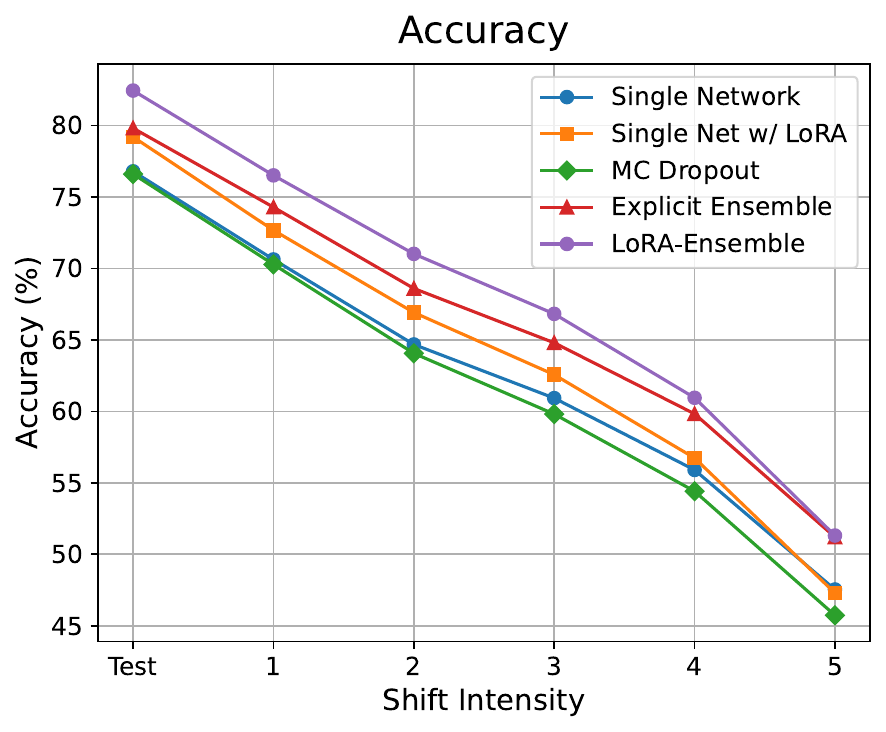}}%
    \hfill
    \subfloat[]{\label{fig:}%
        \includegraphics[width=0.45\textwidth]{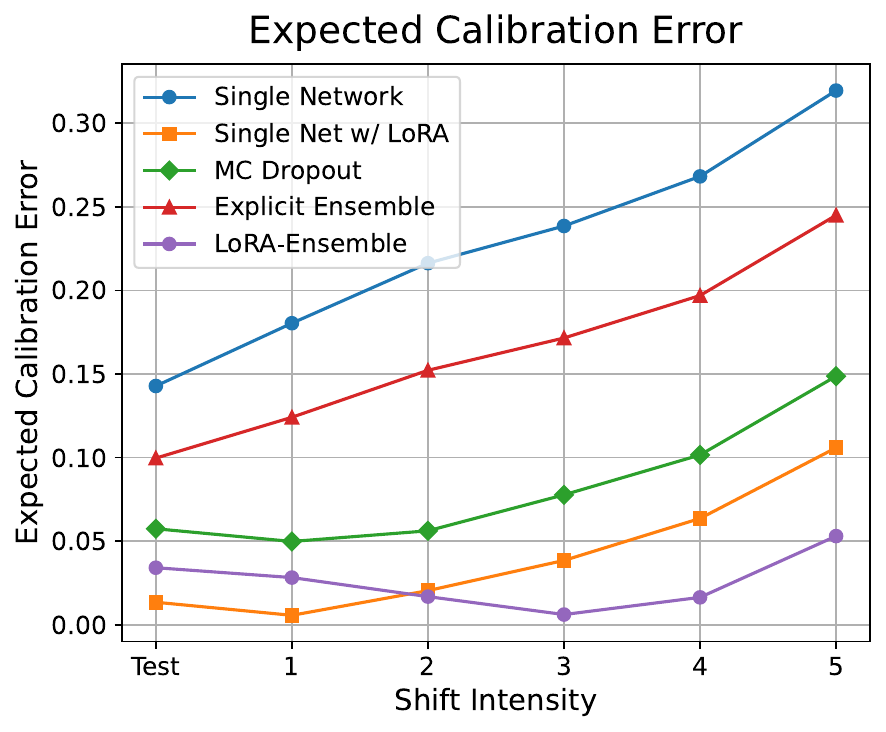}}\\
    
    \subfloat[]{\label{fig:}%
        \includegraphics[width=0.45\textwidth]{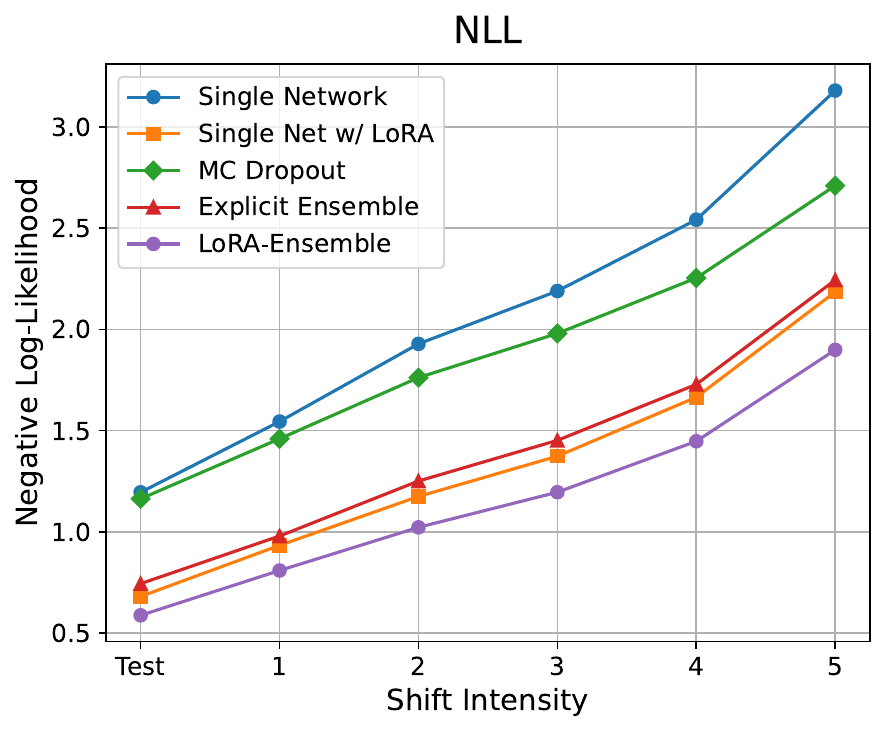}}%
    \hfill
    \subfloat[]{\label{fig:}%
        \includegraphics[width=0.45\textwidth]{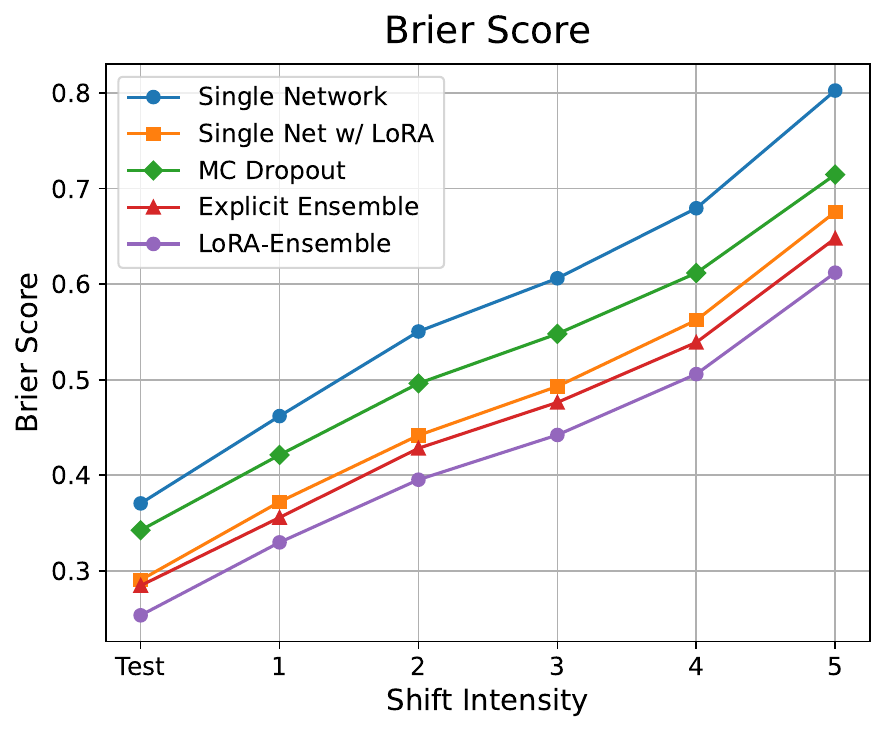}}%

    \caption{\glsxtrshort{lora}-Ensemble evaluated under varying levels of distribution shift on the CIFAR-100-C dataset. Each ensemble consists of 16 members. "Test" refers to the original CIFAR-100 test set, while the corrupted sets include test images subjected to 19 different augmentations at multiple severity levels, introducing distribution shifts.}

    \label{fig:results_CIFAR100C}
\end{figure}

Fig.~\ref{fig:results_CIFAR10C} and~\ref{fig:results_CIFAR100C} present the results. It is evident that the \glsxtrshort{lora}-Ensemble outperforms the other methods, maintaining relatively low \glsxtrshort{ece} scores even under high levels of distribution shift.

\subsection{Computational Cost}\label{subsec:computationl_cost}
%In addition to evaluating classification performance and calibration, we assess the computational cost in terms of parameters, training time and inference time. The required resources are presented in Tab.~\ref{tab:model_resources}.
%
In addition to evaluating classification performance and calibration, we assess the computational cost in terms of parameters, memory usage, training time, and inference time. The required resources are presented in Tab.~\ref{tab:model_resources}.

\begin{table}
    \caption{Parameter counts and computation times for an Explicit Ensemble of 16 ViT models and the corresponding LoRA-Ensemble. Training time is the average duration for one epoch on CIFAR-100, with batch size 32. Inference time is the average duration of a forward pass, with batch size 1. GPU memory denotes the peak allocated CUDA memory.}
    \footnotesize
    \centering
    \begin{tabular}{lccccc}
        \toprule
        \textbf{Method} & \textbf{Params} &
        \textbf{Mem$_\text{train}$ [MiB]} &
        \textbf{Time$_\text{train}$ [s]} &
        \textbf{Mem$_\text{infer}$ [MiB]} &
        \textbf{Time$_\text{infer}$ [ms]} \\
        \midrule
        %New measurements on runpod  & \\
        %Single  & \ \ \ $1\times 87\mathrm{M}$ & $1 \times 2045$ & $1 \times 114$ &  $1 \times 1381$ &$1 \times 5.2$ \\
        Explicit Ensemble  & \ \ \ $16\times 87\mathrm{M}$ & $16 \times 2045$ & $16 \times 114$ &  $16 \times 1381$ &$16 \times 5.2$ \\
        %Explicit Ensemble (all together)  & \ \ \ $16\times 87\mathrm{M}$ & $31606$ & $1595$ &  $21507$ &$89$ \\
        %\glsxtrshort{lora}-Ensemble (vmap) & $1.12\times 87\mathrm{M}$ & 74011 &\ \ \ \ \ \ \ 1868 & 74424 \ \ \ \ \ \ \ &25.0\\ %($16\times20.0$)\\
        \glsxtrshort{lora}-Ensemble& $1.12\times 87\mathrm{M}$ & 15821 &\ \ \ \ \ \ \ 1832 & 2335 \ \ \ \ \ \ \ &14.8\\ %($16\times20.0$)\\
        %\midrule
        %Explicit Ensemble  & \ \ \ $16\times 87\mathrm{M}$ &  &$16 \times 139$ &  &$16 \times 4.6$ \\
        %Explicit Ensemble (non-optimized) & \ \ \ $16\times 87\mathrm{M}$ & $§16\times 127$ & $16 \times 4.9$ \\
        %\glsxtrshort{lora}-Ensemble & $1.12\times 87\mathrm{M}$ & &\ \ \ \ \ \ \ 1108 & \ \ \ \ \ \ \ &22.7\\ %($16\times20.0$)\\
        %\midrule
        %Explicit Ensemble (\glsxtrshort{ast}) & $16 \times 87\mathrm{M}$ & $517$ & $16 \times 7.3$ \\
        %\glsxtrshort{lora}-Ensemble (\glsxtrshort{ast} & $1.08 \times 87\mathrm{M}$ & $348$ & $73.9 (16 \times 20.1)$ \\
    \bottomrule
    \end{tabular}
    \label{tab:model_resources}
\end{table}

The total \emph{number of parameters} is reported for an ensemble of 16 members, with LoRA matrices $A$ and $B$ of rank 8. Choosing a different rank will slightly alter the parameter count; in many cases a lower rank may suffice~\citep{Hu2021LoRA:Models}. All measurements were conducted on a single NVIDIA Tesla A100-80GB GPU. \emph{Training time} is reported as the total wall-clock time per epoch on CIFAR-100 with batch size 32. For the Explicit Ensemble, this corresponds to training 16 members sequentially; LoRA-Ensemble trains all members jointly in a single pass. \emph{Inference time} is the average duration of a forward pass on a single CIFAR-100 example with batch size 1. For the Explicit Ensemble, members are processed sequentially on a single GPU\footnote{Explicit Ensembles can be parallelized by distributing members across multiple GPUs; however, this requires proportionally more hardware and does not reflect a fair comparison under fixed compute constraints.}, so we report the cumulative time for all 16 members.

The results demonstrate that LoRA-Ensemble achieves substantial efficiency gains: approximately 14 times fewer parameters, 9 times less inference memory, and over 5 times faster inference compared to Explicit Ensembles. Training time is comparable between the two methods, as LoRA-Ensemble processes all members jointly while the Explicit Ensemble trains members sequentially. The primary advantage of LoRA-Ensemble lies in its memory efficiency, which enables deployment of large ensembles on resource-constrained hardware where Explicit Ensembles would be infeasible.

%\section{Effect of Model Size on Prediction and Calibration Performance}

\section{LoRA-Ensemble's Generalization to Varying Model Sizes}\label{app:model_size_study}
Building upon our existing experiments with the HAM10000 dataset, we extended our analysis to include different backbone architectures with varying numbers of parameters. Specifically, we utilized various DeiT models pre-trained with distillation, as described by \citet{Touvron2020TrainingAttention}. The results are presented in Table~\ref{tab_app:model_performance_HAM10000_smaller_models}. Notably, the DeiT Base-32 model is the same as the ViT Base-32 model.

In the small parameter regime (Tiny-16, Small-16), the addition of a single LoRA module did not consistently enhance calibration compared to using a single model. This observation contrasts with our findings in most other experiments. However, in the larger parameter regime (ViT Base-32), incorporating even a single LoRA module significantly improved calibration.
Increasing the number of ensembles in the LoRA-Ensemble not only boosted accuracy but also enhanced calibration, enabling it to match the performance of an Explicit Ensemble in both parameter regimes. 

%Finally, as the number of parameters in the backbone architecture increased, the superiority of the LoRA-Ensemble over the Explicit Ensemble in terms of both accuracy and calibration became more pronounced. 

Last but not least, as the number of parameters in the backbone architecture increased, the superiority of the LoRA-Ensemble over the Explicit Ensemble in terms of both accuracy and calibration became more pronounced. %With smaller models, the LoRA-Ensemble received only 2 or 3 gold tokens, whereas with the largest model, it received all 5.
This trend indicates that as backbone size grows, the advantages of LoRA-Ensemble become increasingly dominant.

Overall, the results demonstrate that the LoRA-Ensemble not only transfers successfully to a different backbone architecture (DeiT versus ViT) but also remains effective across varying parameter regimes.

\definecolor{gold}{HTML}{D0B43E}
\definecolor{silver}{HTML}{BFBFBF}
\definecolor{bronze}{HTML}{B28A58}

\begin{table}[ht]
    \caption{
    Performance metrics on the HAM10000 dataset for different Vision Transformer architectures. Ensembles have 16 members. The top two results for each metric are highlighted: \textbf{bold} for the best, \underline{underlined} for the second best.
    }
    \centering
    \setlength{\tabcolsep}{4.5pt}
    \begin{adjustwidth}{-0pt}{}
    \resizebox{1.0\linewidth}{!}{
    \begin{tabular}{clcccccc}
    \toprule
    \textbf{Arch.} & \textbf{Method} & \textbf{\# Params.} & \textbf{Accuracy ($\uparrow$)} & \textbf{F1 ($\uparrow$)} & \textbf{ECE ($\downarrow$)} & \textbf{NLL($\downarrow$)} & \textbf{Brier ($\downarrow$)} \\
    \midrule
     \multirow{4}{*}{\rotatebox{90}{\shortstack[c]{DeiT\\Tiny-16}}} & Single Net & \multirow{4}{*}{\rotatebox{90}{$5\,\mathrm{M}$}}
        & $\underline{89}.0\pm0.3$ 
        & $79.0\pm0.4$ 
        & $0.096\pm0.003$ 
        & $0.909\pm0.037$ 
        & $0.202\pm0.005$ \\
     & Single Net w/ LoRA 
        & 
        & $84.5\pm0.8$ 
        & $71.6\pm1.5$
        & $0.074\pm0.003$ 
        & $0.542\pm0.017$ 
        & $0.237\pm0.009$\\
     & Explicit Ensemble 
        & 
        & $\textbf{90.4}\pm0.3$ 
        & $\textbf{81.4}\pm0.4$ 
        & $\underline{0.069}\pm0.004$ 
        & $\underline{0.340}\pm0.006$ 
        & $\textbf{0.142}\pm0.002$\\
     & LoRA-Ensemble 
        & 
        & $88.9\pm0.4$ 
        & $\underline{80.6}\pm0.2$ 
        & $\textbf{0.025}\pm0.003$ 
        & $\textbf{0.325}\pm0.004$ 
        & $\underline{0.164}\pm0.002$\\
    \midrule
    \multirow{4}{*}{\rotatebox{90}{\shortstack[c]{DeiT\\Small-16}}} & Single Net & \multirow{4}{*}{\rotatebox{90}{$22\,\mathrm{M}$}}
        & $89.6\pm0.4$ 
        & $79.0\pm0.5$ 
        & $0.093\pm0.003$ 
        & $0.876\pm0.032$ 
        & $0.191\pm0.007$ \\
     & Single Net w/ LoRA 
        & 
        & $86.3\pm0.5$ 
        & $76.8\pm1.0$
        & $0.100\pm0.007$ 
        & $0.731\pm0.053$ 
        & $0.234\pm0.010$\\
     & Explicit Ensemble 
        & 
        & $\textbf{91.5}\pm0.1$ 
        & $\underline{82.4}\pm0.2$ 
        & $\underline{0.061}\pm0.002$ 
        & $\underline{0.318}\pm0.003$ 
        & $\textbf{0.130}\pm0.001$\\
     & LoRA-Ensemble 
        & 
        & $\underline{90.4}\pm0.1$ 
        & $\textbf{82.8}\pm0.4$ 
        & $\textbf{0.047}\pm0.002$ 
        & $\textbf{0.292}\pm0.002$ 
        & $\underline{0.144}\pm0.001$\\
    \midrule
    \multirow{4}{*}{\rotatebox{90}{\shortstack[c]{DeiT\\Base-32}}} & Single Net & \multirow{4}{*}{\rotatebox{90}{$86\,\mathrm{M}$}}
        & $84.1\pm0.3$ 
        & $71.4\pm0.7$ 
        & $0.139\pm0.004$ 
        & $1.138\pm0.040$ 
        & $0.291\pm0.009$ \\
     & Single Net w/ LoRA
        & 
        & $83.2\pm0.7$ 
        & $70.7\pm1.3$
        & $\underline{0.085}\pm0.004$ 
        & $0.569\pm0.027$ 
        & $0.256\pm0.011$\\
     & Explicit Ensemble 
        & 
        & $\underline{85.8}\pm0.2$ 
        & $\underline{74.6}\pm0.4$ 
        & $0.105\pm0.002$ 
        & $\underline{0.536}\pm0.007$ 
        & $\underline{0.218}\pm0.002$\\
     & LoRA-Ensemble 
        & 
        & $\textbf{88.0}\pm0.2$ 
        & $\textbf{78.3}\pm0.6$ 
        & $\textbf{0.037}\pm0.002$ 
        & $\textbf{0.342}\pm0.003$ 
        & $\textbf{0.175}\pm0.002$\\
    \midrule
    \end{tabular}
    } \label{tab_app:model_performance_HAM10000_smaller_models}
    \end{adjustwidth}
\end{table}

\section{Hyperparameter Selection and Sensitivity Analysis: LoRA Rank}\label{subsec:sensitivity}
The main hyper-parameter introduced by adding \glsxtrshort{lora} is the rank of the low-rank decomposition (i.e., the common dimension of the matrices $A$ and $B$).
Varying that rank modulates the complexity of the model for the learning task. We have empirically studied the relationship between rank, accuracy, and \glsxtrlong{ece}. Here we show results for HAM10000 and CIFAR-100 dataset.

\begin{figure}[ht]
    \centering
    \subfloat[HAM10000.]{\label{fig:ham10000_rank_study}    \includegraphics[width=.9\textwidth]{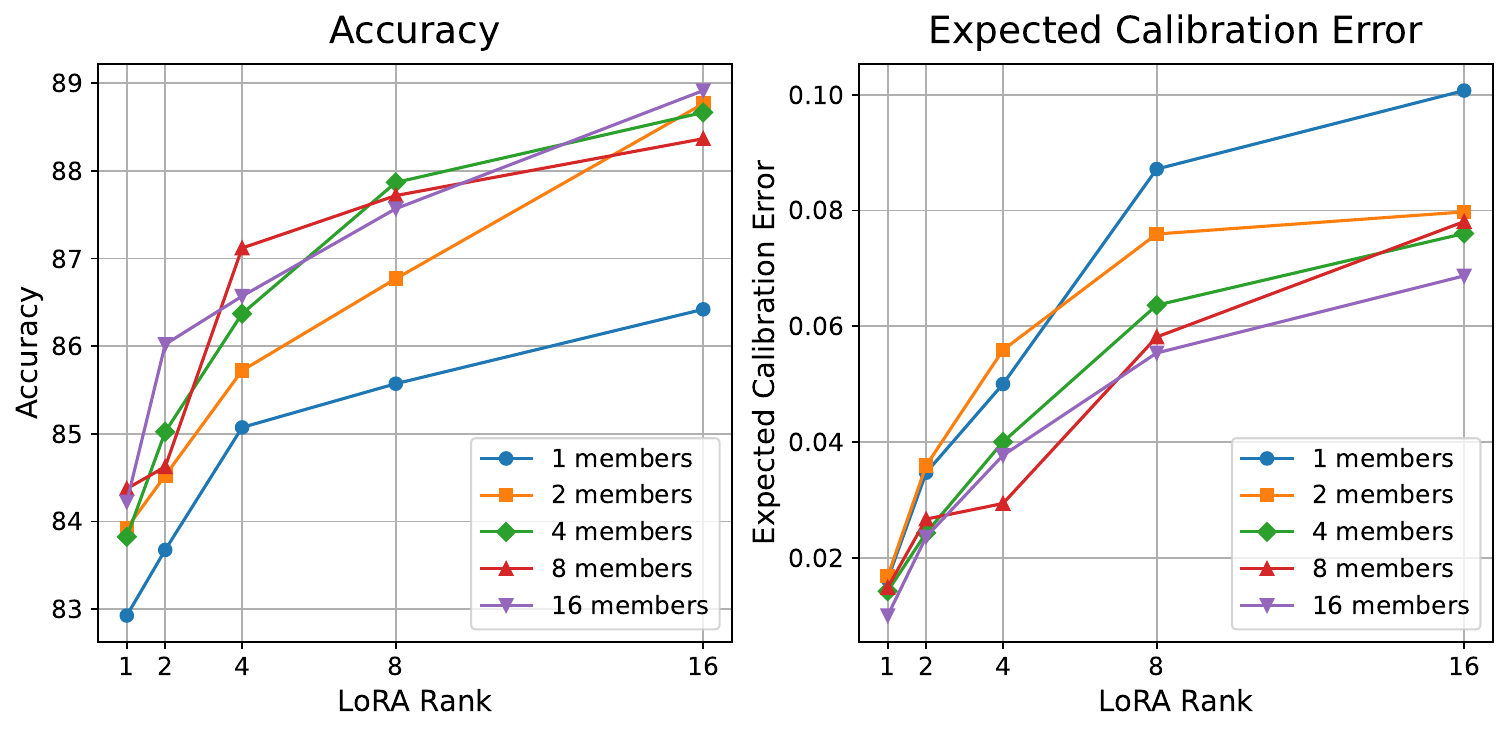}} \\
    \subfloat[CIFAR-100]{\label{fig_app:cifar_rank_ablation}        \includegraphics[width=.9\textwidth]{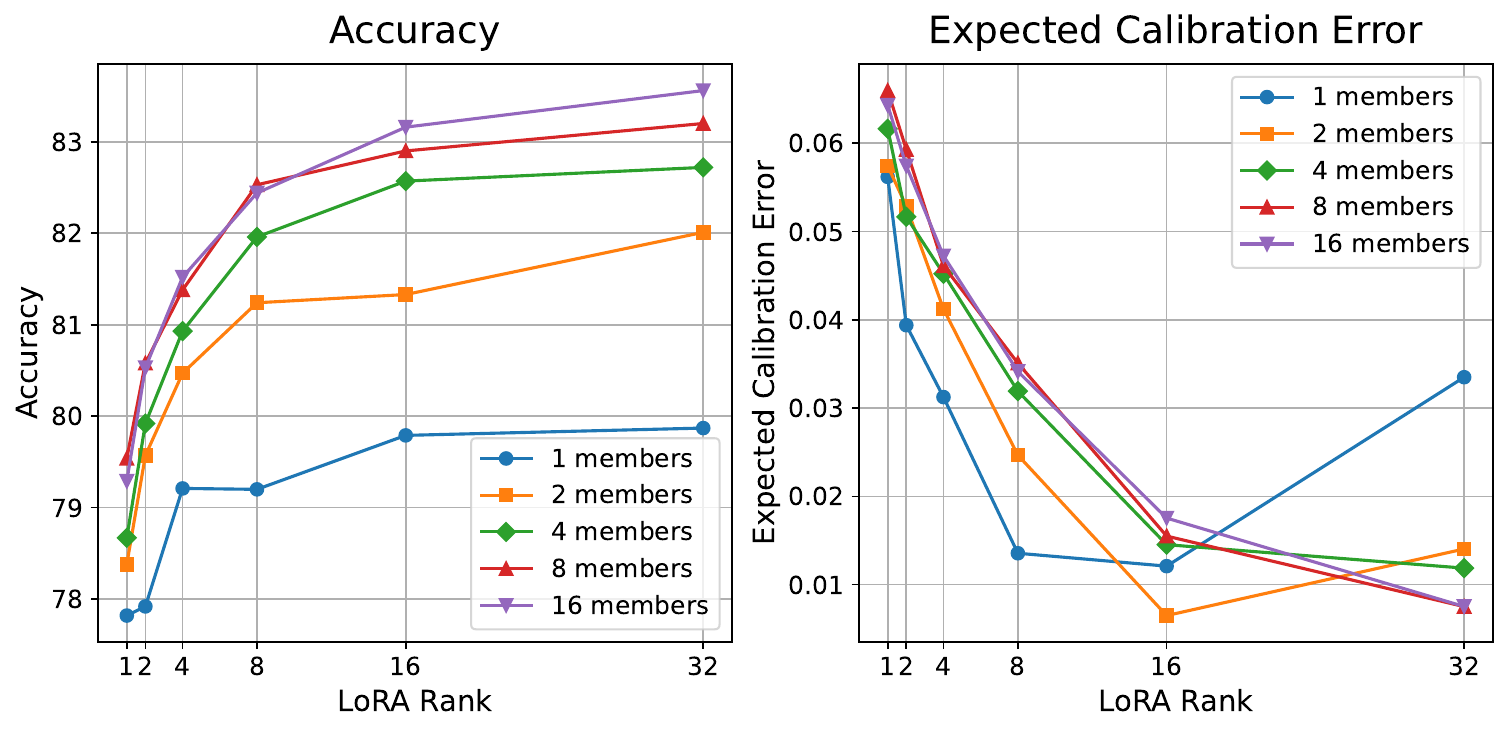}} 
    %\caption{Accuracy and \glsxtrlong{ece} for different initialization methods and varying distribution parameters for changing ensemble size for LoRA-Ensemble.}
    \caption{Impact of \glsxtrshort{lora} rank on accuracy and \glsxtrshort{ece}.}
    \label{fig_app:lora_init}
\end{figure}

On HAM10000 we observe a clear trade-off between accuracy and calibration, Fig.~\ref{fig:ham10000_rank_study}. With increasing rank the classification accuracy increases while the calibration deteriorates, in other words, one can to some degree balance predictive accuracy against uncertainty calibration by choosing the rank.
Our focus in this work is on model calibration. We therefore generally choose the rank to favor calibration, even at the cost of slightly lower classification accuracy.

For the CIFAR-100 dataset, our evaluation of \glsxtrshort{lora}-Ensemble shows both increased accuracy and improved calibration with increasing rank within the studied range. These findings are illustrated in Fig.~\ref{fig_app:cifar_rank_ablation}.

This observation aligns with the findings of \citet{Rahaman2020UncertaintyEnsembles}, as \glsxtrshort{lora}-Ensemble continues to exhibit under-confidence even at higher ranks. Increasing model complexity enhances confidence, thereby improving calibration. However, at rank 32, the calibration of a single network augmented with \glsxtrshort{lora} begins to deteriorate, suggesting that a critical boundary has been reached. Beyond this point, the parameter space becomes insufficiently constrained, leading to effects similar to those observed by \citet{Guo2017OnNetworks}.

At higher ranks, accuracy plateaus while memory demand increases linearly with $\mathcal{O}(d)$ and $\mathcal{O}(k)$ for $A\in\R^{r\times d}$ and $B\in\R^{k \times r}$ respectively, where $d$ and $k$ are the dimensions of the pre-trained weight matrix $W_0\in\R^{k\times d}$. Consequently, we selected rank 8 for our CIFAR-100 experiments.

Overall, the rank serves as the primary control of \glsxtrshort{lora}’s expressive capacity. While larger values tend to improve performance on more complex datasets (e.g., rank~64 for iNaturalist), excessively large choices (e.g., $\geq 256$) suppress the distinctive dimension-learning behavior of \glsxtrshort{lora}-Ensemble, resulting not only in diminishing returns but in some cases an actual decline in accuracy. In practice, we find that a small sweep over $\{4, 8, 16, 32, 64\}$ on a held-out set is typically sufficient to identify a near-optimal rank.

\paragraph{Practical guidelines for choosing the LoRA rank.}
Based on the sensitivity analyses, we propose the following practical recommendations for selecting the LoRA rank in LoRA-Ensemble:
\begin{itemize}
    \item \textbf{Default choice.} For small- to medium-sized datasets and modest domain shifts, low ranks (e.g., $r\in[4\hdots8]$) provide a favorable trade-off between accuracy, ensemble diversity, and calibration.
    \item \textbf{When to increase the rank.} Higher ranks may be needed for large or fine-grained datasets (e.g., iNaturalist), particularly when accuracy saturates at low ranks and calibration remains stable.
    \item \textbf{Avoid excessive ranks.} Increasing the rank too far risk degrading the calibration and tends to reduce the diversity of ensemble members, as the characteristic LoRA learning dynamics are gradually lost.
    \item \textbf{Monitoring calibration.} When increasing the rank, calibration metrics such as ECE should be monitored alongside accuracy, as improvements in predictive performance do not necessarily translate into better-calibrated uncertainty estimates.
\end{itemize}

%\begin{figure}[th]
%    \centering
%    \includegraphics[width=.9\textwidth]{Figures/03_Experiments/HAM10000_lora_rank_results}
%    \caption{Impact of \glsxtrshort{lora} rank on accuracy and \glsxtrshort{ece}, for HAM10000 dataset.}
%    \label{fig:ham10000_rank_study}
%\end{figure}
%%

%\begin{figure}[h]
%    \centering
%    \includegraphics[width=.9\textwidth]{Appendices/Figures/CIFAR100_lora_rank_results.pdf}
%    \caption{Impact of \glsxtrshort{lora} rank on accuracy and ECE, for the CIFAR-100 dataset.}
%    \label{fig_app:cifar_rank_ablation}
%\end{figure}
%

\section{Weight Space Analysis: LoRA-Ensemble versus Explicit Ensemble}
\label{app:weight_space}

\begin{figure}[th]
    \setlength{\tabcolsep}{1pt}
    \centering
    \scriptsize
    \begin{tabular}{ccccc}
         Explicit & LoRA & Explicit & LoRA & \\
        \includegraphics[width=.19\textwidth]{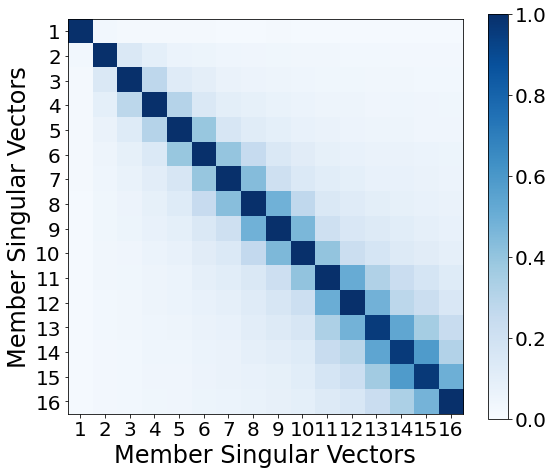} &
        \includegraphics[width=.19\textwidth]{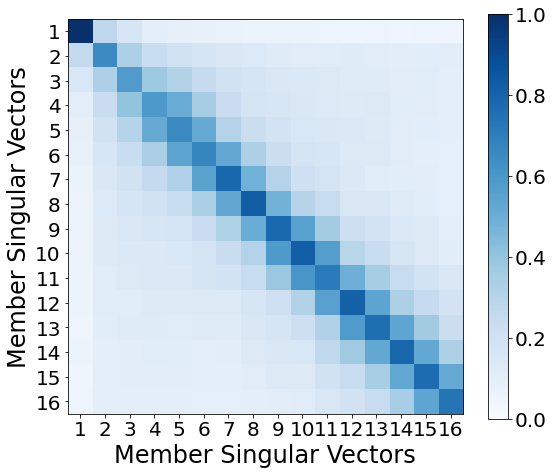} &
        \includegraphics[width=.19\textwidth]{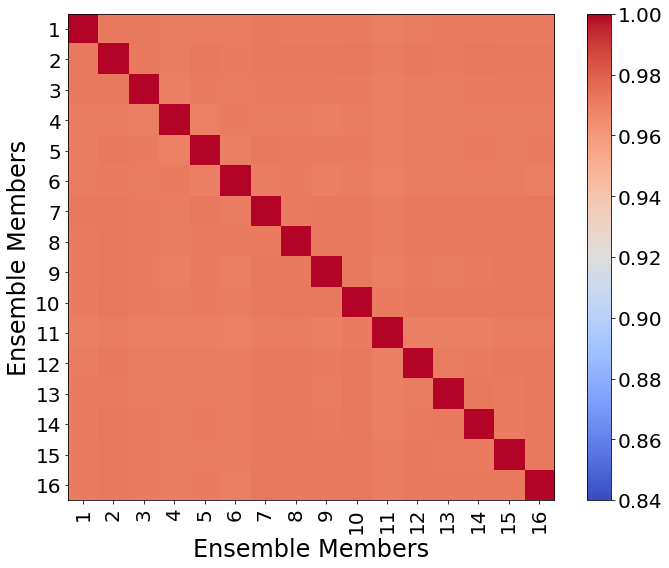} &
        \includegraphics[width=.19\textwidth]{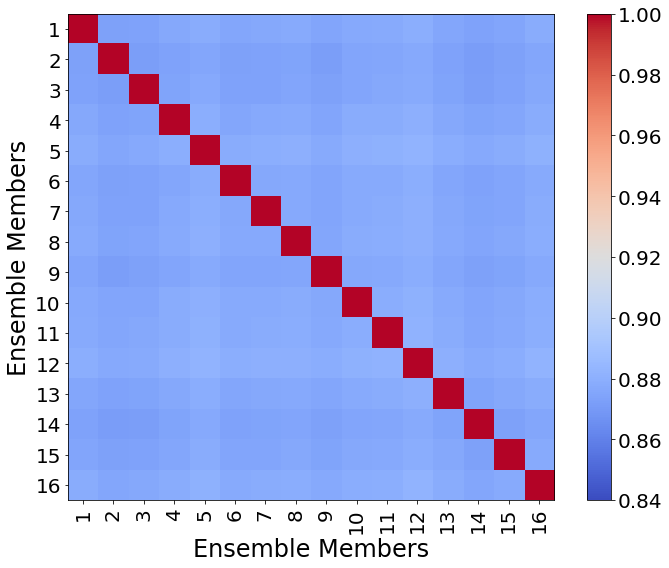} &
        \includegraphics[width=.19\textwidth]{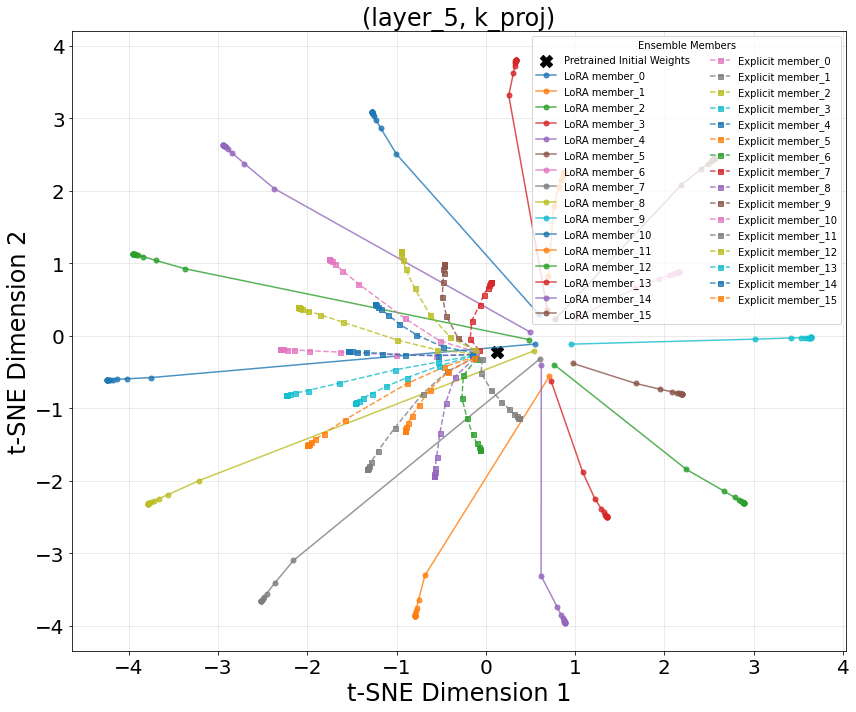} \\
      
        \includegraphics[width=.19\textwidth]{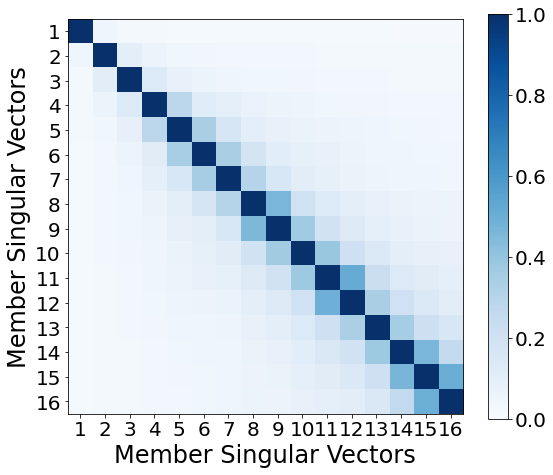} &
        \includegraphics[width=.19\textwidth]{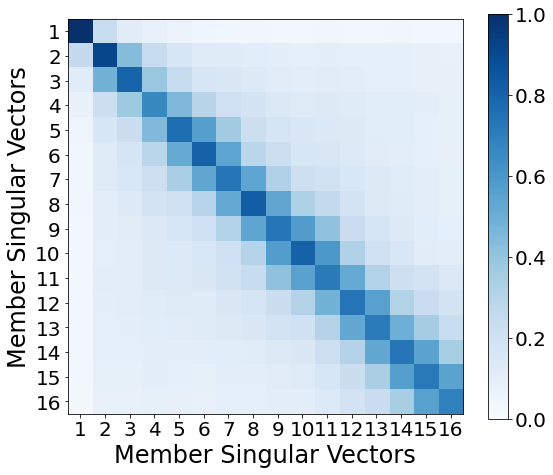} &
        \includegraphics[width=.19\textwidth]{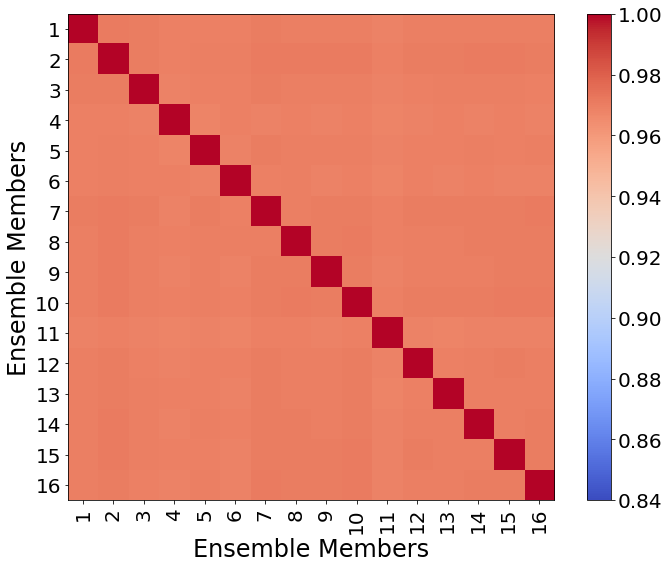} &
        \includegraphics[width=.19\textwidth]{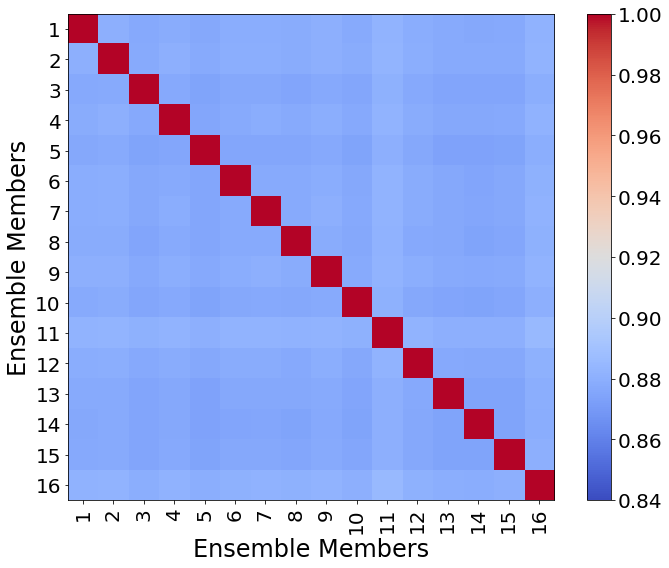} &
        \includegraphics[width=.19\textwidth]{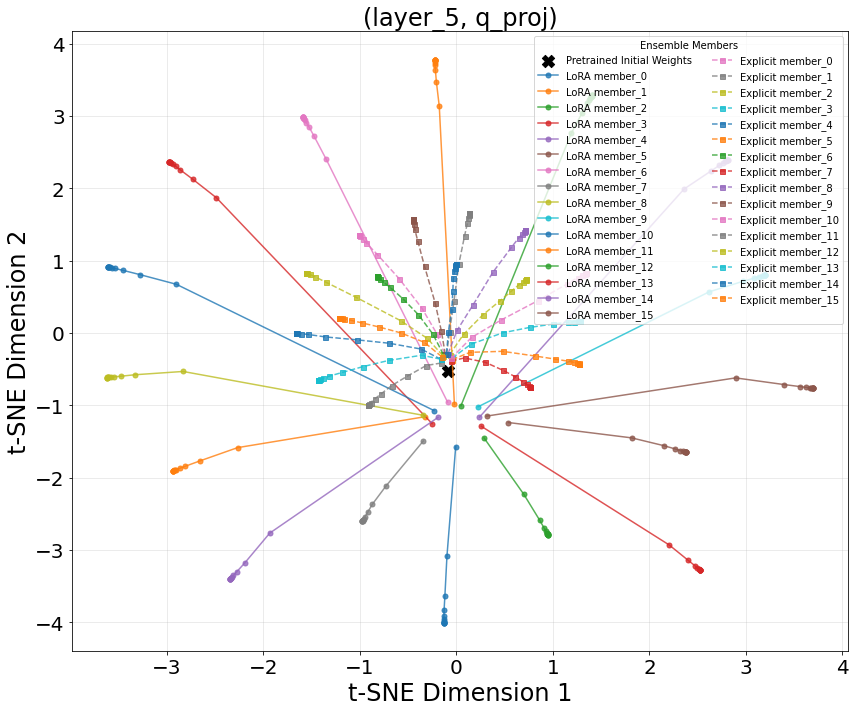} \\
        \multicolumn{2}{c}{(a) cos-similarity of high-ranking singular vectors} &
    \multicolumn{2}{c}{(b) weight-space cosine similarity} &
    (c) training trajectories\\
    \end{tabular} 
 \caption{Weight space analysis of \glsxtrshort{lora}-Ensemble vs. Explicit Ensemble: The first row represents key matrices, while the second row represents query matrices.}
\label{fig:weight_diversity_k_q}
\end{figure}

This section expands on Sec.~\ref{section:diversity}, which examines the diversity of ensemble members in function and weight space for LoRA-Ensemble and Explicit Ensemble, showing that LoRA-Ensemble exhibits greater diversity in both spaces.
While Sec.~\ref{section:diversity} focuses on value projection matrices due to their role in learned representations, this section examines query and key projection matrices, too.
In Fig.~\ref{fig:weight_diversity_k_q}, we observe that LoRA-Ensemble achieves greater diversity in query and key projection matrices, similar to the diversity observed in value projection matrices (Fig.~\ref{fig:weight_diversity}).

Using Singular Value Decomposition (SVD), a weight matrix \( W \in \mathbb{R}^{m \times n} \) is decomposed as:
\[
W = U \Sigma V^\top,
\]
where \( U \in \mathbb{R}^{m \times m}\) and \( V \in \mathbb{R}^{n \times n}\) are orthonormal matrices representing rotational components, and \( \Sigma \in \mathbb{R}^{m \times n}\) is a diagonal matrix of singular values capturing the scaling effect. Singular vectors linked to larger singular values highlight key transformations encoded by \( W \).

%This broader analysis provides a deeper view of the differences in weight space diversity across all attention projection matrices for LoRA-Ensemble and Explicit Ensemble.

\begin{figure}[th]
    \setlength{\tabcolsep}{1pt}
    \centering
    \scriptsize
    \begin{tabular}{cccc}
        & Explicit & LoRA & Singular Values  \\
        \raisebox{16ex}{\rotatebox{90}{VALUE}} &
        \includegraphics[width=.32\textwidth]{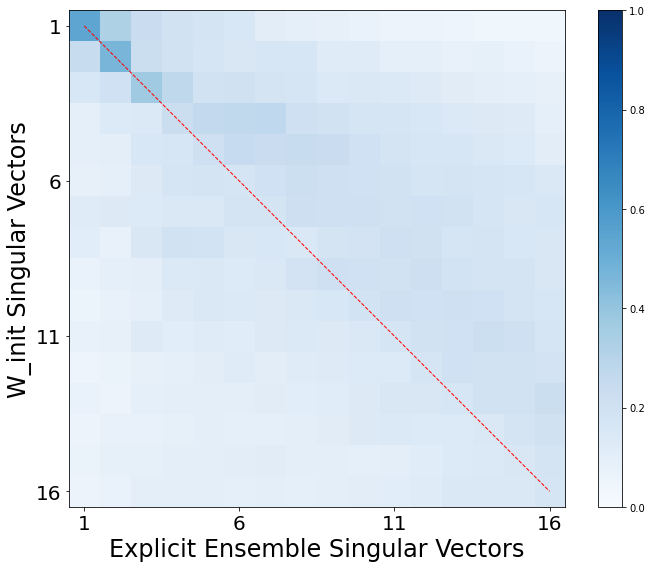} &
        \includegraphics[width=.32\textwidth]{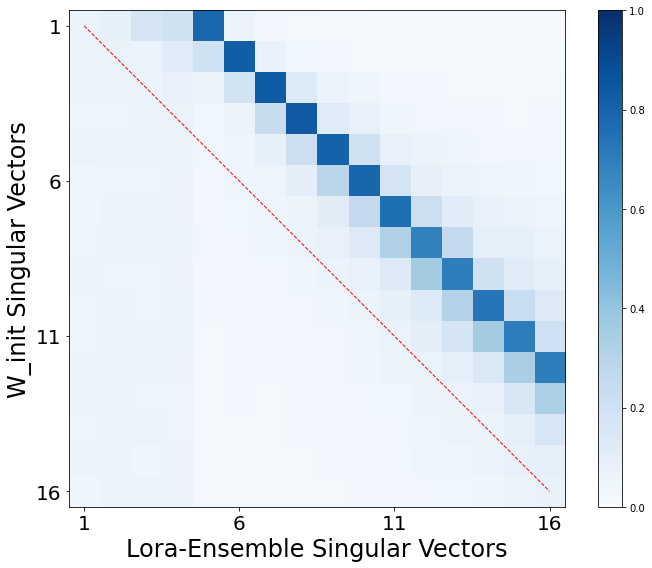} &
        \includegraphics[width=.32\textwidth]{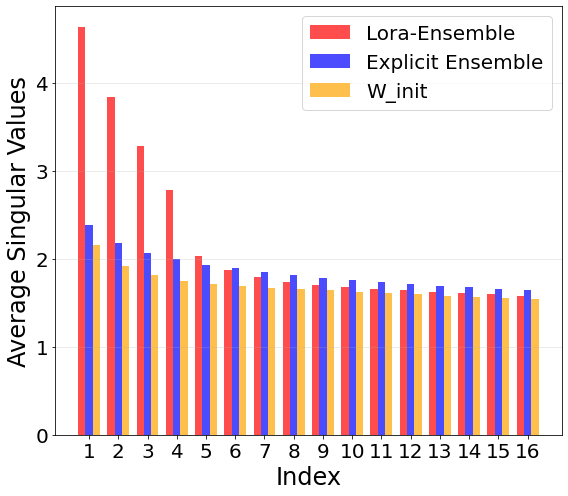} \\
        \raisebox{16ex}{\rotatebox{90}{KEY}} &
        \includegraphics[width=.32\textwidth]{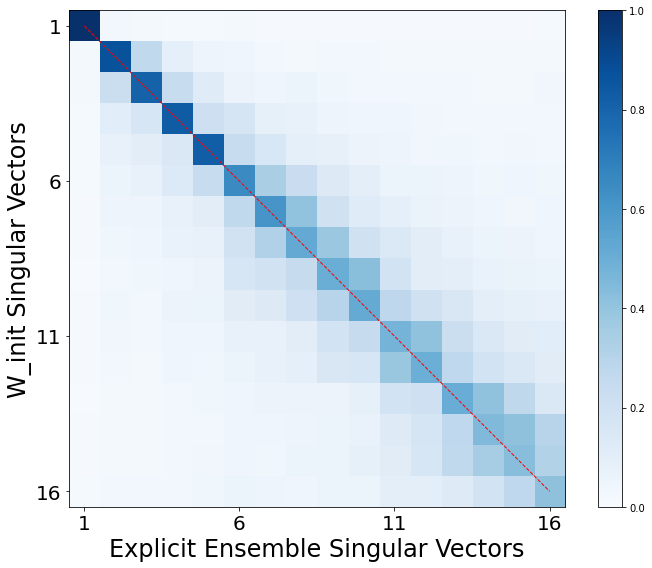} &
        \includegraphics[width=.32\textwidth]{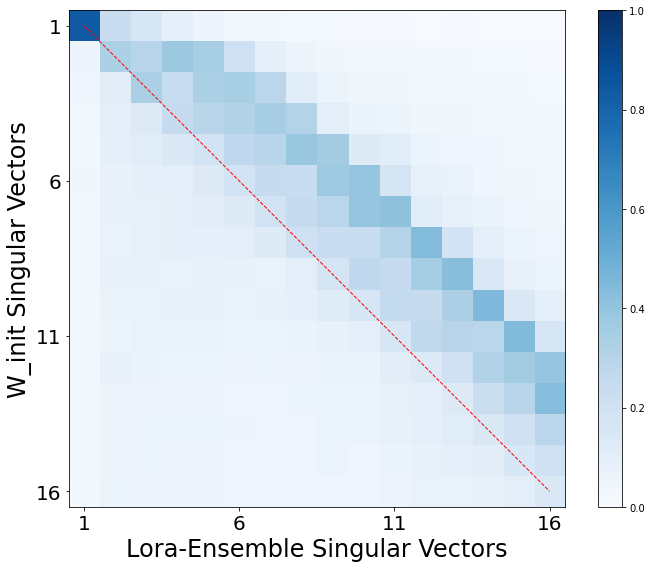} &
        \includegraphics[width=.32\textwidth]{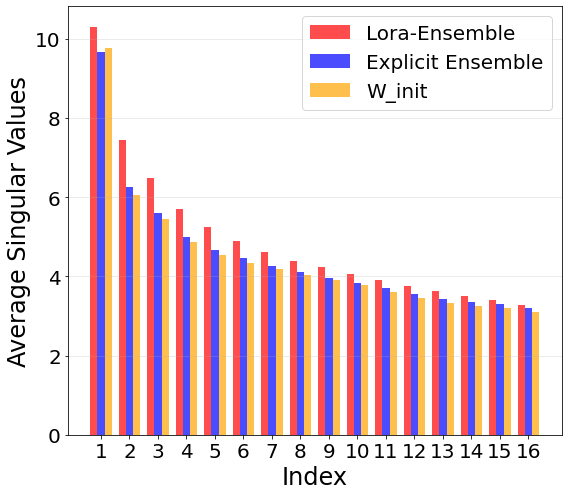} \\
        \raisebox{16ex}{\rotatebox{90}{QUERY}} &
        \includegraphics[width=.32\textwidth]{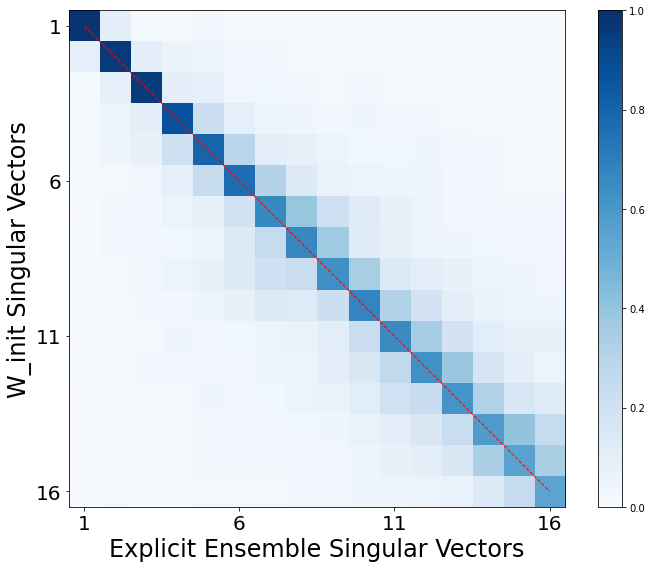} &
        \includegraphics[width=.32\textwidth]{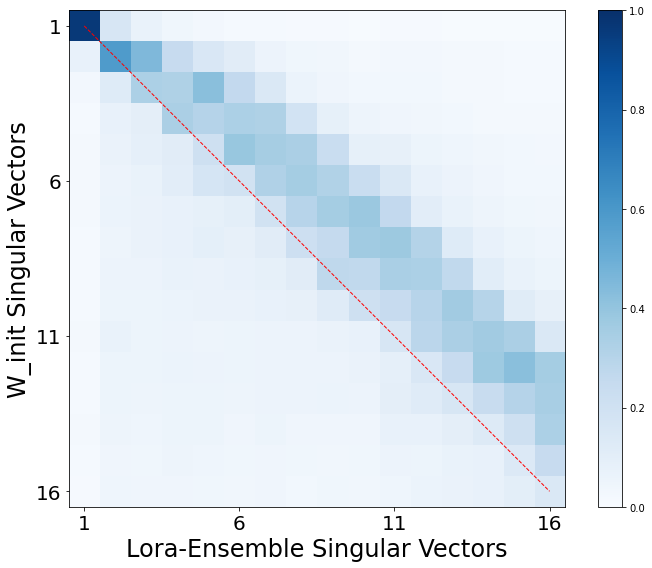} &
        \includegraphics[width=.32\textwidth]{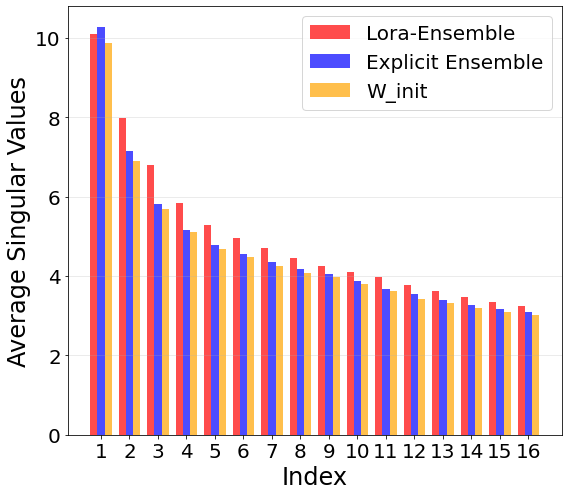} \\
    \end{tabular} 
 \caption{Cosine similarity of top singular vectors (and associated singular values) between initial pre-trained and final trained weights, averaged over layers and ensemble members.}
\label{fig:weight_diversity_intruder}
\end{figure}

In Fig.~\ref{fig:weight_diversity_intruder}, we analyze the differences in weight updates between ensemble methods by computing the Singular Value Decomposition (SVD) of pre-trained and trained weights for ensemble members. Singular vectors corresponding to the top singular values (16 are shown) are extracted and compared using cosine similarity to evaluate changes in the weight structure. These similarities are averaged across layers and ensemble members.
The results highlight distinct parameter update patterns between LoRA-Ensemble and Explicit Ensemble. LoRA-Ensemble introduces new high-ranking singular vectors, referred to as "intruder dimensions" \citep{shuttleworth2024lora}, which are nearly orthogonal to the singular vectors of the pre-trained weights. The number of intruder dimensions depends on the LoRA rank. This effect is particularly pronounced in the value projection matrices, which aligns with their strong association with learned representations.
In contrast, Explicit Ensemble members tend to preserve a structure closely aligned with the spectral properties of the pre-trained weights. This alignment is especially evident in the key and query projection matrices, which exhibit a strong resemblance to the original spectral structure.

\begin{figure}[th]
    \setlength{\tabcolsep}{1pt}
    \centering
    \scriptsize
    \begin{tabular}{ccc}
        VALUE & KEY & QUERY  \\
        \includegraphics[width=.32\textwidth]{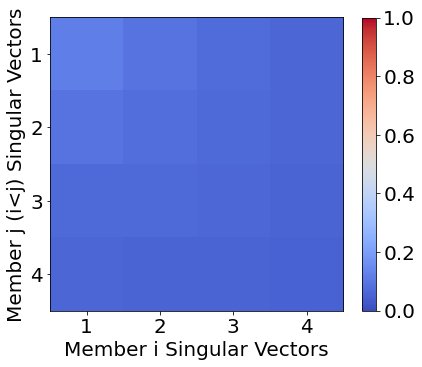} &
        \includegraphics[width=.32\textwidth]{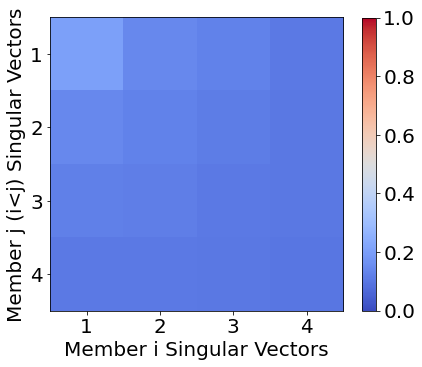} &
        \includegraphics[width=.32\textwidth]{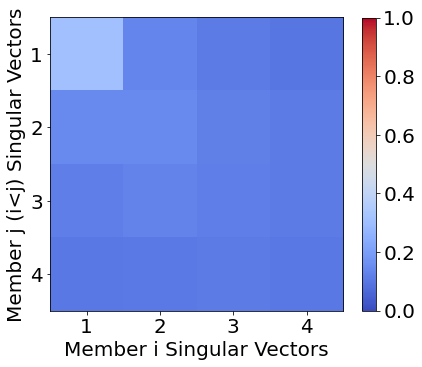} \\
       
        \includegraphics[width=.32\textwidth]{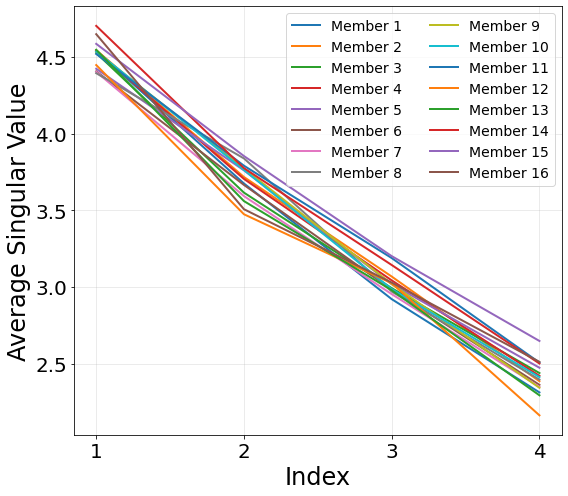} &
        \includegraphics[width=.32\textwidth]{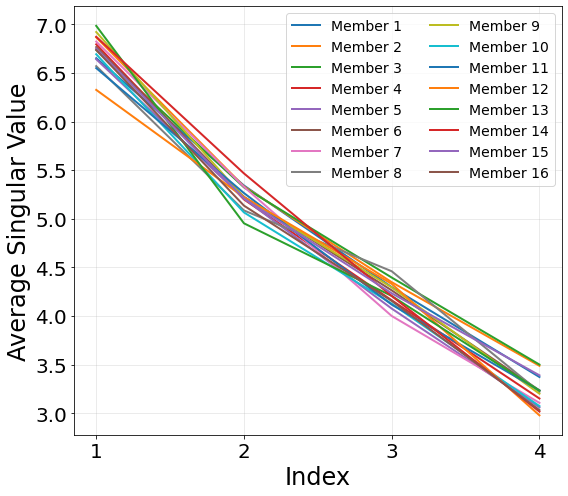} &
        \includegraphics[width=.32\textwidth]{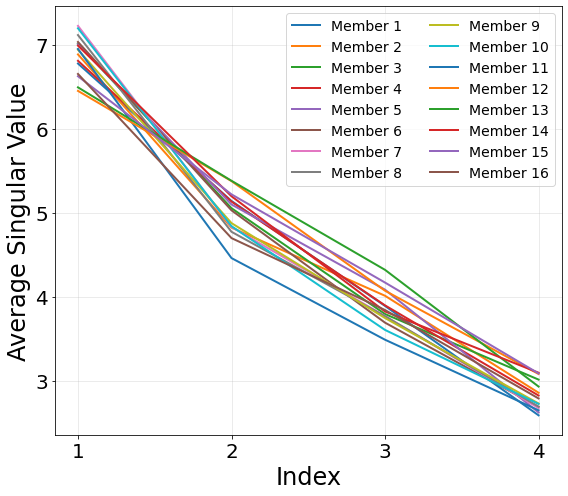} \\
    \end{tabular} 
 \caption{Cosine similarity of top singular vectors from \(B \cdot A\) low-rank matrices (rank set to 4) between LoRA-Ensemble members, averaged over layers and all member pairs (first row), along with corresponding average singular values for different members (second row).}
\label{fig:weight_diversity_AB}
\end{figure}

We further analyze the \( B \cdot A \) matrices learned by different ensemble members. Due to their random initialization, these matrices explore diverse directions in weight space. In Fig.~\ref{fig:weight_diversity_AB}, we plot the largest eigenvalues of these matrices (with only four non-zero eigenvalues as the LoRA rank is set to 4) and the similarity between the corresponding eigenvectors across ensemble members. The similarities are averaged over layers and member pairs.
The results show that while the eigenvalues across members follow a similar trend, the eigenvectors are largely uncorrelated. This indicates that ensemble members explore different regions of weight space while maintaining similar overall transformations. The shared eigenvalue trends suggest consistent semantic contributions across members, while the dissimilar eigenvectors highlight the diversity in their learned representations.

\begin{figure}[th]
    \setlength{\tabcolsep}{1pt}
    \centering
    \scriptsize
    \begin{tabular}{ccccc}
        \includegraphics[width=.49\textwidth]{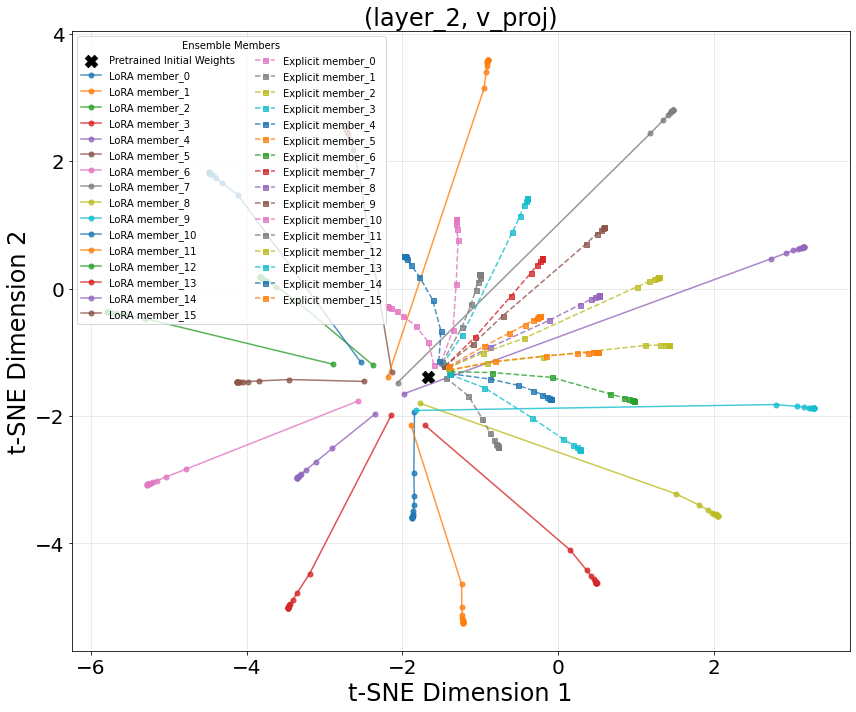} &
        \includegraphics[width=.49\textwidth]{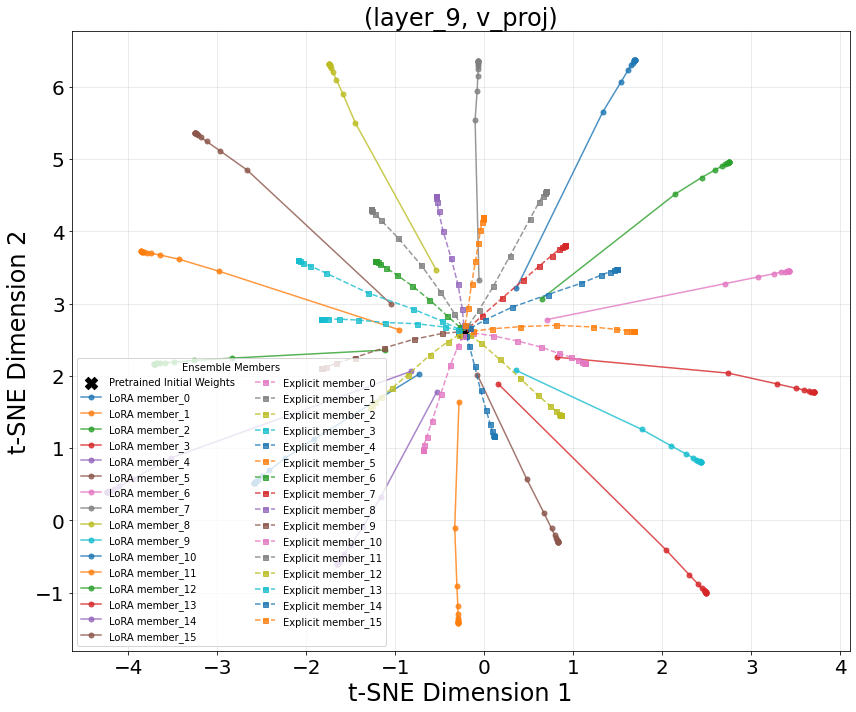} \\
    \end{tabular} 
 \caption{Training trajectories of ensemble members of \glsxtrshort{lora}-Ensemble and \ Explicit Ensemble.}
\label{fig:trajectory}
\end{figure}

We plot the t-SNE visualizations for different layers in Fig.~\ref{fig:trajectory}, capturing the evolution of weights during training. The visualizations include the initial pretrained weights, and for each ensemble member, we plot weights from epoch 5 to epoch 65 at 5-epoch intervals. The plots reveal that LoRA-Ensemble members exhibit broader convergence across the loss landscape in various layers, signifying diverse learning dynamics. Conversely, Explicit Ensemble members tend to remain closer to their initial weights, indicating reduced diversity throughout the training process.

\section{Correlation Analysis Between Ensemble Diversity and Predictive Performance}
\label{app:diversity_performance_correlation}

Prior work has shown that diversity across modes in weight space correlates with improved uncertainty estimates~\citep{fort2019deep, Izmailov2021WhatLike}. To investigate this phenomenon in the context of \glsxtrshort{lora}, we trained eight-member ensembles on the HAM10000 dataset using varying LoRA initialization gains.

We treat the LoRA initialization gain as a simple \emph{diversity knob}: larger gains induce greater spread among the low-rank adapters. To quantify ensemble diversity, we compute the average pairwise correlation between the LoRA $V$ projection matrices of different ensemble members, averaged across all layers. The \emph{diversity score} is then defined as
\[
\text{Diversity score} = 1 - \text{average correlation}.
\]

This metric captures how dissimilar the ensemble members are in terms of their learned parameters, with higher scores indicating less correlated (i.e., more diverse) adapters. 

Tab.~\ref{tab:lora_diversity} summarizes the results. Both performance and calibration improve as diversity increases, up to a point. Beyond a certain threshold, additional diversity provides diminishing returns and eventually plateaus. Moreover, as shown in Fig.~\ref{fig:weight_diversity_AB}, excessively large gains degrade performance. From a Bayesian perspective, moderate gains behave like a well-chosen prior variance, encouraging the ensemble to explore distinct posterior modes. Excessively large gains, however, make the initial weights too diffuse, which can cause activation saturation, unstable gradients, suboptimal convergence, and ultimately higher variance and miscalibration. A sweep of gain values reveals a U-shaped calibration curve: uncertainty estimates improve with increasing diversity until diversity becomes excessive, at which point performance and calibration degrade.

\begin{table}[h]
\centering
\begin{tabular}{cccc}
\toprule
\textbf{Gain} & \textbf{Diversity Score} & \textbf{Accuracy} & \textbf{NLL} \\
\midrule
1   & 0.006 & 0.854 & 0.405 \\
2   & 0.017 & 0.856 & 0.393 \\
4   & 0.048 & 0.860 & 0.374 \\
8   & 0.117 & 0.877 & 0.354 \\
12  & 0.184 & 0.881 & 0.345 \\
20  & 0.304 & 0.882 & 0.345 \\
40  & 0.511 & 0.882 & 0.344 \\
\bottomrule
\end{tabular}
\caption{Effect of LoRA initialization gain on ensemble diversity, accuracy, and calibration (measured by NLL) on HAM10000.}
\label{tab:lora_diversity}
\end{table}

\section{Joint Training of Backbone and LoRA-Ensemble Layers on iNaturalist}\label{app:joint_inat}

\glsxtrshort{lora}-Ensemble keeps the backbone weights frozen and trains only the low-rank matrices. To assess the impact of relaxing this constraint, we evaluated the method in a setting where the full backbone is also trainable. Results on the iNat 2017 dataset, see Tab.~\ref{tab_app:_model_performance_INat2017}, show a substantial drop in both accuracy and calibration when the entire network is trained, although performance still surpasses that of a single model. 
%
%We hypothesize that enabling full backbone training reduces the LoRA effect caused by intruder dimensions, leading to less diversity among ensemble members and a convergence toward the spectral structure of the initial weights and LoRA-Ensemble resembles a single network. Similar behavior was observed for the Batch-Ensemble method, as discussed in Appendix~\ref{app_sec:batchEnsemble}.

We hypothesize that enabling backbone training washes out the low-rank adjustments introduced by LoRA. With a frozen backbone, each model’s unique low-rank matrices create intruder dimensions that yield diverse feature spaces. See Appendix~\ref{app:weight_space} for more details. Once the backbone is trainable, those adjustments merge into the dominant spectral modes, causing all ensemble members to collapse into the same parameter region and behave like a single network. Similar behavior was observed for the Batch-Ensemble method, as discussed in Appendix~\ref{app_sec:batchEnsemble}.

\begin{table}[h]
    \small
    \caption{Performance on the iNat 2017 dataset. 'full' indicates that the entire network, including the pre-trained backbone, is trainable. Ensembles consist of 4 members. Best score for each metric in \textbf{bold},  second-best \underline{underlined}.}
    \centering
    \begin{tabular}{lccccc}
    \toprule
    \textbf{Method} & \textbf{Accuracy ($\uparrow$)} & \textbf{F1 ($\uparrow$)} & \textbf{ECE ($\downarrow$)} & \textbf{NLL ($\downarrow$)} & \textbf{Brier ($\downarrow$)} \\
    \midrule
    Single Network & $42.6$ & $37.8$& $0.293$ & $1.054$ & $0.207$\\
    \midrule
    Single Net w/ LoRA & $\underline{47.7}$ & $\underline{43.1}$& $\underline{0.096}$ & $\underline{0.662}$ & $\underline{0.166}$\\
    
    Single Net w/ LoRA (full) & $42.8$ & $38.0$ & $0.271$ & $0.958$ & $0.201$ \\

    \midrule
    LoRA-Ensemble & $\textbf{49.3}$ & $\textbf{44.1} $ & $\textbf{0.045}$ & $\textbf{0.610}$ & $\textbf{0.160}$\\
    LoRA-Ensemble (full) &  $44.0$ & $39.4$ & $0.249$ & $0.886$  & $0.193$\\
    \bottomrule
    \end{tabular}
    \label{tab_app:_model_performance_INat2017}
\end{table}

\section{Placement of LoRA-Ensemble Modules and Selection of Trainable Weights}\label{app:placement_lora_ensemble}
Typically, LoRA is applied only to the weights in the multi-head attention module (i.e., the query, key, value, and output projections), as demonstrated in the original LoRA paper~\citep{Hu2021LoRA:Models}. We acknowledge that, due to the modular nature of transformer architectures, LoRA layers can also be inserted into the feedforward MLP blocks. While this alternative placement has been shown in prior work to improve performance on certain datasets, it may also lead to reduced robustness and lower overall performance~\citep{fomenko2024notelora}. Moreover, the projection matrices in the MLP blocks typically have significantly higher dimensionality, often four times larger than those in the attention layers. As a result, this placement introduces a substantial number of additional parameters, which can increase memory consumption. This effect is especially pronounced when ensemble members are executed in parallel using vectorized mapping rather than sequential execution.

Table~\ref{tab_app:ablation_study_architecture} presents the results on the HAM10000 dataset. Notably, adding LoRA-Ensemble layers to the MLP blocks leads to improved accuracy, but at the cost of poorer calibration performance. Additionally, when the embedding layers of the Vision Transformer (ViT) are also trained alongside the low-rank matrices for the purpose of patch feature extraction, we observe a marked drop in classification accuracy. This performance degradation can be attributed to the substantial number of additional parameters introduced in the early stages of the model, which are by design an order of magnitude larger than in the subsequent LoRA layers, potentially leading to suboptimal training dynamics.

Finally, we observe that assigning a separate classification head to each LoRA-Ensemble member yields further performance gains. However, we also note that this design choice can be omitted in favor of improved parameter efficiency, depending on the application constraints.

\begin{table}[h]
    \scriptsize
    \caption{Ablation Study. Investigates the placement of LoRA-Ensemble layers and additional trainable components on the HAM10000 dataset. Ensembles consist of 8 members. Best score for each metric in \textbf{bold},  second-best \underline{underlined}.}
    \centering
    \resizebox{1.0\linewidth}{!}{
    \begin{tabular}{lccccccc}
    \toprule
    \textbf{LoRA-Ensemble Config.} & \textbf{Extra Trainable Layers}& Trainable Params. & \textbf{Accuracy ($\uparrow$)} & \textbf{F1 ($\uparrow$)} & \textbf{ECE ($\downarrow$)} & \textbf{NLL ($\downarrow$)} & \textbf{Brier ($\downarrow$)} \\
    \midrule
    Multi-head attention & Cls. head & 2'364'679 &$87.5$ & $77.7$& $0.041$ & \underline{$0.365$} & $0.187$\\
    
    Multi-head attention  & Cls. head + tokenizer & 4'724'743 &$84.6$ & $73.8$& $\textbf{0.025}$ & $0.422$ & $0.217$\\
    Multi-head attention + MLP & Cls. head & 5'313'799 &$\textbf{90.1}$ & $\textbf{80.9}$& $0.077$ & $0.383$ & $\textbf{0.157}$\\
    Multi-head attention + MLP & Cls. head + tokenizer & 7'673'863 &$87.4$ &$77.1$& $0.083$ & $0.438$ & $0.192$\\
    Multi-head attention  & Full backbone & 90'114'055 &$85.2$ & $73.3$& $0.126$ & $1.000$ & $0.264$\\
    \midrule
    Multi-head attention  & Ensemble cls. head & 2'402'360 &\underline{$88.0$} & \underline{$78.0$}& \underline{$0.036$} & $\textbf{0.347}$ & \underline{$0.179$}\\
    \bottomrule
    \end{tabular}
    }
    \label{tab_app:ablation_study_architecture}
\end{table}

\section{LoRA-Ensemble for CNNs}\label{app:CNN}

We extend LoRA Ensemble to convolutional neural networks (CNN) by applying it to a ResNet-18 backbone with an ensemble of four members. We mainly follow the original Batch-Ensemble~\citep{Wen2020BatchEnsemble:Learning} implementation. For detailed experimental settings, see \citep{Turkoglu2022FiLM}. Table~\ref{tab_app:lora_cnn} reports the CIFAR-100 results. LoRA-Ensemble achieves the second-best performance among implicit ensembling methods, behind only FiLM-Ensemble, but it does not match its efficacy on transformer architectures. As discussed in the main text and in the Appendix~\ref{app_sec:imlicit_baseline}, this gap stems from the fundamentally different computational structures of transformers compared with MLPs and CNNs, which makes direct adaptation of techniques between these domains challenging.

\begin{table}[ht]
    \small
    \caption{Performance on the CIFAR-100 dataset for CNN architecture. Ensembles have 4 members and Resnet-18 is used as a backbone. For implicit ensemble methods, the best score for each metric in \textbf{bold},  second-best \underline{underlined}.}
    \centering
    \begin{tabular}{lcc}
    \toprule
    \textbf{Method} & \textbf{Accuracy ($\uparrow$)} & \textbf{ECE ($\downarrow$)}  \\
    \midrule
    Single Network & $78.0\pm0.4$ & $0.046\pm0.001$\\
    Deep Ensemble & $81.6\pm0.3$& $0.041\pm0.002$ \\
    \midrule

    MC-Dropout & $75.5\pm0.6$& $0.064\pm0.003$ \\
    MIMO & $48.0\pm2.6$& $0.083\pm0.017$ \\
    Masksemble & $72.5\pm0.5$& $0.075\pm0.004$ \\
    FiLM-Ensemble & $\textbf{79.4}\pm0.2$ & $\textbf{0.038}\pm0.000$ \\
    
    Batch-Ensemble & $77.7\pm0.1$& $0.052\pm0.002$ \\
    
    LoRA-Ensemble &   $\underline{78.4}\pm0.2$  & $\underline{0.048}\pm0.001$ \\
    \bottomrule
    \end{tabular}
    \label{tab_app:lora_cnn}
\end{table}

\section{LoRA-Ensemble Fine-Tuned on the Same Dataset as the Backbone Model}\label{app:finetuned_same_dataset}

While our study explicitly focused on transfer learning setups, we also explored how LoRA-Ensemble can be applied when the backbone is trained on the same dataset. To this end, we initialized the LoRA ensemble with weights from a single network trained for 65 epochs on HAM10000, and subsequently fine-tuned it for one epoch without learning rate warmup. Fig.~\ref{app_tab:same_dataset} presents the results of fine-tuning a LoRA-Ensemble with rank 2 for a single epoch. It is evident that even in this scenario, the LoRA-Ensemble improves both performance and calibration with minimal computational overhead. We also highlight that alternative methods, such as explicit ensembling, are not directly applicable in this context.

\begin{table}[ht]
    \caption{LoRA-Ensemble performance when it is fine-tuned on a pre-trained dataset. The HAM10000 dataset is used, and the ensemble consists of 8 members. The backbone is identical to that of the Single Network, which is fine-tuned for one epoch. Best score for each metric in \textbf{bold}.}
    \centering
    %\scriptsize % Reduce font size
    %\setlength{\tabcolsep}{10pt} % Reduce horizontal space between columns
    \resizebox{1.0\linewidth}{!}{
    \begin{tabular}{lccccc}
        \toprule
    \textbf{Method} & \textbf{Accuracy ($\uparrow$)} & \textbf{F1 ($\uparrow$)} & \textbf{ECE ($\downarrow$)} & \textbf{NLL ($\downarrow$)} & \textbf{Brier ($\downarrow$)} \\
    \midrule
    Single Network & $84.1\pm0.3$ & $71.4\pm0.7$ & $0.139\pm0.004$ & $1.138\pm0.040$ & $0.291\pm0.009$ \\
    Single Net w/ LoRA & $83.2\pm0.7$ & $70.7\pm1.3$& $0.085\pm0.004$ & $0.569\pm0.027$ & $0.256\pm0.011$\\
    \midrule
    LoRA-Ensemble (finetuned for 1 epoch) & $\textbf{84.8}$ & $\textbf{72.2}$& $\textbf{0.059}$ & $\textbf{0.514}$ & $\textbf{0.238}$\\
    \bottomrule
    \end{tabular}
    }
    \label{app_tab:same_dataset}
\end{table}

% None,0.5143164396286011,0.847728431224823,0.7223573934216043,0.7775190431226108,0.683791282304802,0.0592220202088356,0.03104272298514843,227.61495971679688,0.5143164396286011,0.23814478516578674

\section{Post-Hoc Temperature Scaling for Model Calibration}\label{app_sec:temp_scaling}

Temperature scaling is a simple yet effective post-hoc calibration method used to improve the confidence of probabilistic models \citep{Guo2017OnNetworks}. It rescales the logits of a trained model by a scalar parameter $T > 0$ (the temperature). Given logits $\mathbf{z}$, the calibrated probabilities $\hat{p}_i$ for class $i$ are computed as:
\begin{equation}
    \hat{p}_i = \frac{\exp(z_i / T)}{\sum_{j} \exp(z_j / T)}.
\end{equation}

Here, $T = 1$ corresponds to no scaling, and $T > 1$ reduces overconfidence by softening the logits. 

To assess the impact of temperature scaling on calibration, we conducted experiments on CIFAR-100 with varying temperature values, as shown in Tab.~\ref{tab:model_performance_temperature_scaling}. For each method, the model parameters were fixed, and the effect of different temperatures on calibration was evaluated. We observe that calibration can be improved across all methods, with the exception of the single network with LoRA, which does not require temperature scaling. 

As discussed in Section~\ref{subsec:CIFAR100_results}, \glsxtrshort{lora}-Ensemble is under-confident on CIFAR-100, as evidenced by the optimal temperature being less than 1 for this method.

\subsection*{When should temperature scaling be applied?}
\begin{itemize}
    \item \textbf{Use it for under-confidence.} Temperature scaling is a good default post-hoc correction when predictions are systematically under-confident (accuracy $>$ confidence across bins).
    \item \textbf{Benefit is dataset-specific.} The need for temperature scaling (and the magnitude of improvement) depends on the dataset and training regime; in some cases LoRA-Ensemble is already close to calibrated, while in others residual miscalibration remains.
    \item \textbf{Decision rule.} Evaluate calibration on a held-out validation set; apply temperature scaling only if it measurably improves calibration metrics (e.g., ECE/NLL) without harming accuracy.
\end{itemize}

\begin{table}[ht]
    \caption{Model performance on the CIFAR-100 dataset with different temperature. Best score for each metric and method in \textbf{bold},  second-best \underline{underlined}.}
    \centering
  %   \scriptsize % Reduce font size
   % \setlength{\tabcolsep}{10pt} % Reduce horizontal space between columns
    \begin{tabular}{lcccccc}
        \toprule
        \textbf{Method} & \textbf{Temp.} &\textbf{Accuracy ($\uparrow$)} & \textbf{F1 ($\uparrow$)} & \textbf{ECE ($\downarrow$)} & \textbf{NLL ($\downarrow$)} & \textbf{Brier ($\downarrow$)} \\
        \midrule
        %Single Network & 1.0 &$\textbf{76.8}$ & $\textbf{76.7}$ & $0.143$ & $1.195$ & $0.371$ \\
        %Single Network & 1.2 & & & $0.119$ & $1.052$ & $0.356$ \\
        Single Network & 1.4 & \textbf{76.8} & \textbf{76.7} & $0.091$ & $0.969$ & $0.344$ \\
        Single Network & 1.6 & & & $0.061$ & $\underline{0.928}$ & $\underline{0.334}$ \\
        Single Network & 1.8 & & & $\underline{0.034}$ & $\textbf{0.920}$ & $\textbf{0.329}$ \\
        Single Network & 2.0 & & & $\textbf{0.029}$ & $0.939$ & $\textbf{0.329}$ \\
        Single Network & 2.2 & & & $0.078$ & $0.982$ & $0.335$ \\
        %Single Network & 2.4 & & & $0.134$ & $1.045$ & $0.349$ \\
        %Single Network & 2.6 & & & $0.193$ & $1.125$ & $0.371$ \\
        %Single Network & 2.8 & & & $0.252$ & $1.216$ & $0.399$ \\
        %Single Network & 3.0 & & & $0.309$ & $1.316$ & $0.434$ \\
        \midrule 
        Single Net w/ LoRA & 0.4 &$\textbf{79.2}$ & $\textbf{79.1}$ & $0.130$ & $1.020$ & $0.332$ \\
        Single Net w/ LoRA & 0.6 & & & $0.088$ & $0.772$ & $0.308$ \\
        Single Net w/ LoRA & 0.8 & & & $\underline{0.042}$ & $\underline{0.688}$ & $\underline{0.294}$ \\
        Single Net w/ LoRA & 1.0 & & & $\textbf{0.013}$ & $\textbf{0.680}$ & $\textbf{0.290}$ \\
        Single Net w/ LoRA & 1.2 & & & $0.073$ & $0.722$ & $0.298$ \\
        \midrule
        MC Dropout & 0.4 &$\textbf{76.6}$ & $\textbf{76.6}$ & $0.203$ & $1.554$ & $0.372$ \\
        MC Dropout & 0.6 & &  & $0.174$ & $1.223$ & $0.361$ \\
        MC Dropout & 0.8 & & & $\underline{0.111}$ & $\textbf{1.114}$ & $\underline{0.344}$ \\
        MC Dropout & 1.0 & & & $\textbf{0.057}$ & $\underline{1.163}$ & $\textbf{0.342}$ \\
        MC Dropout & 1.2 & & & $0.175$ & $1.333$ & $0.393$ \\
        \midrule
        Explicit Ensemble & 1.0 & $\textbf{79.8}$ & $\textbf{79.9}$ & $0.100$ & $0.744$ & $0.285$ \\
        Explicit Ensemble & 1.2 & & & $0.072$ & $\underline{0.719}$ & $\underline{0.282}$ \\
        Explicit Ensemble & 1.4 & & & $\underline{0.041}$ & $\textbf{0.718}$ & $\textbf{0.281}$ \\ 
        Explicit Ensemble & 1.6 & & & $\textbf{0.019}$ & $0.737$ & $0.284$ \\
        Explicit Ensemble & 1.8 & & & $0.046$ & $0.777$ & $0.290$ \\
        \midrule
        LoRA-Ensemble & 0.4 & $\textbf{82.4}$ & $\textbf{82.4}$ & $0.103$ & $0.628$ & $\underline{0.252}$ \\
        LoRA-Ensemble & 0.6 & & & $0.063$ & $\underline{0.565}$ & $\textbf{0.247}$ \\
        LoRA-Ensemble & 0.8 & & & $\textbf{0.018}$ & $\textbf{0.557}$ & $\textbf{0.247}$ \\
        LoRA-Ensemble & 1.0 & & & $\underline{0.034}$ & $0.587$ & $0.253$ \\
        LoRA-Ensemble & 1.2 & & & $0.095$ & $0.650$ & $0.269$ \\
        \bottomrule
    \end{tabular}
    \label{tab:model_performance_temperature_scaling}
\end{table}

\section{Implementation of LoRA-Ensmeble} \label{app:implementation}
In practice, our \glsxtrshort{lora}-Ensemble is implemented by replacing the respective linear layers ($W_q$, $W_k$, $W_v$, and $W_o$) in the pre-trained model architecture with custom \glsxtrshort{lora} modules.

As a backbone for experiments with image datasets, we employ a \glsxtrfull{vit} model \citep{Dosovitskiy2020AnScale}. The chosen architecture is the \textit{base} variant with patch size $32\times32$ as defined in \citep{Dosovitskiy2020AnScale}. We load the weights from \texttt{torchvision}, which were trained on ImageNet-1k \citep{Deng2009ImageNet:Database}, using a variant of the training recipe from \citep{Touvron2020TrainingAttention}, for details refer to their documentation.

The forward pass through the backbone is parallelized by replicating the input along the batch dimension. In each \glsxtrshort{lora} module, the data is split into separate inputs per member and passed to the respective member with the help of a \emph{vectorized map}, which allows a parallelized forward pass even through the \glsxtrshort{lora} modules. The outputs are then again stacked along the batch dimension. In this way, one makes efficient use of the parallelization on \glsxtrshort{gpu}, while at the same time avoiding loading the pre-trained backbone into memory multiple times.
%
%\textcolor{green}{For language tasks, we adopt the BERT-base architecture in its uncased variant \citep{Devlin2019BERTPO}, pre-trained on a large English corpus comprising Wikipedia and BookCorpus data. The \glsxtrshort{lora} modules are integrated into the attention layers following the same design as in our vision experiments.}
%
As a backbone for audio experiments, we use the \glsxtrfull{ast} backbone~\citep{Gong2021AST:Transformer}. That architecture was inspired by \glsxtrshort{vit} (more specifically the data-efficient version of \glsxtrshort{vit} akin to \glsxtrshort{deit}~\citep{Touvron2020TrainingAttention}) but is designed specifically for audio spectrograms. Following \citep{Gong2021AST:Transformer}, we initialize the audio model weights by transferring and appropriately interpolating them from ImageNet pre-training. See Appendix~\ref{app_C:AST_implementation} and~\ref{app_C:AST_Validation} for details.
As the \glsxtrshort{ast} version of \glsxtrshort{lora}-Ensemble would run into memory limits, we introduce chunking. While the forward pass through the backbone is still parallelized, the \glsxtrshort{lora} modules are called sequentially.%
\footnote{For the Explicit Ensemble the vectorization could not be used on GPU, due to a technical issue with the \glsxtrshort{vit} implementation in PyTorch.}

Finally, the pre-trained model does not have the correct output dimension for our prediction tasks (i.e., it was trained for a different number of classes). Therefore we entirely discard its last layer and add a new one with the correct dimensions, which we train from scratch. Obviously, the weights of that last layer are different for every ensemble member. We parallelize it in the same way as the \glsxtrshort{lora} module described above.

%A PyTorch implementation of \glsxtrshort{lora}-Ensemble, as well as pre-trained weights to reproduce the experiments, will be publicly released on GitHub. We also submit our code in the supplementary file.

\section{Training Details of LoRA-Ensemble} \label{subsec:2trainingsettings}
The CIFAR-10/100, HAM10000, and iNaturalist 2017 dataset experiments are based on the ViT-Base-32 architecture \citep{Dosovitskiy2020AnScale}. This model has 12 layers and uses 768-dimensional patch embeddings, and the multi-head attention modules have 12 heads. All \glsxtrlong{vit} models for image classification are trained using the AdamW optimizer \citep{Loshchilov2017DecoupledRegularization}, except for iNat 2017, which is trained with SGD using a momentum of 0.9. The base learning rate is initially set to 0.0001 with a batch size of 32 for all experiments, except for iNat 2017, where a learning rate of 0.1 is used with a batch size of 128. Training employs a learning rate warm-up of 500 steps for all experiments, except for iNat 2017, which uses 2500 warm-up steps. During the warm-up phase, the learning rate increases linearly from 0 to the base value, after which it follows a cosine decay schedule for the remaining steps. 
For iNat 2017, an exponential learning rate decay with a factor of 0.94 is applied every 4 epochs. During the experiments, the gradients were calculated and then clipped not to exceed a maximum norm of 1. In the case of HAM10000, we used a weighted cross-entropy loss that considered the estimated effective number of samples, which was determined using a beta parameter of 0.9991 \citep{Cui2019Class-balancedSamples}. Uniform class weights were used for all other datasets. The maximum number of training epochs varies depending on the dataset. For CIFAR-10/100, the model is trained for 16 epochs (just over 25,000 steps), while for HAM10000 and iNat 2017, it is trained for 65 and 64 epochs, respectively. Overall, the hyperparameters used in this work were loosely based on \citep{Conrad2023Fine-tuningTransformers}. The models were trained using pre-trained weights from \texttt{torchvision 0.17.1} on an NVIDIA Tesla A100 graphics card. 
Moreover, the \glsxtrshort{lora} models are configured with a rank of 8 for CIFAR-10/100, 4 for HAM10000, and 64 for iNat 2017. For Monte Carlo Dropout the dropout rate was empirically set to be $0.2$. Refer to Appendix~\ref{app_E:MCDropout_HyperparameterTuning} for details.

The settings used for the ESC-50 dataset training are similar to those used in \citep{Gong2021AST:Transformer}. However, we used a batch size of 1 instead of 48 to enable training on a single GPU. The base learning rate is set to 0.00001 for the Explicit Ensemble as well as MC Dropout experiments and 0.00005 for \glsxtrshort{lora}-Ensemble. These learning rates are lower than the ones used in \citep{Gong2021AST:Transformer}, which is due to the smaller batch size. Refer to the Appendix~\ref{app_D:AST_HyperparameterTuning} for more details. The \glsxtrshort{lora} models were implemented with a rank of 16. The dropout rate for MC dropout was kept at $0.2$.

For language experiments on the SST-2 dataset \citep{socher2013recursive} we used the BERT base uncased model \citep{Devlin2019BERTPO}, loaded via the HuggingFace Transformers library \citep{wolfetal2020transformers}. Training utilizes the AdamW optimizer \citep{Loshchilov2017DecoupledRegularization} with $\beta_1 = 0.9$ and $\beta_2 = 0.999$, a linearly decaying learning rate over three epochs, and a batch size of 16. These settings were informed by prior work that used BERT on SST2 \citep{gchhablani2023bertbasecasedsst2}. We conduct a separate hyperparameter tuning for each method and select the learning rate from the candidate set $\{2\times10^{-6},\,7\times10^{-6},\,9\times10^{-6},\,2\times10^{-5},\,3\times10^{-5},\,5\times10^{-5},\,7\times10^{-5}\}$ that yields the highest accuracy. For MC Dropout, we used a dropout rate of 0.2. For LoRA-Ensemble, we set the rank to 64 and initialize the LoRA layers using Xavier uniform initialization \citep{Glorot2010UnderstandingNetworks} with a gain factor of 10.

As \citet{Fort2019DeepPerspective} have shown, varying initializations of the weights are most important to getting diverse ensemble members. For this reason, various initialization methods and corresponding parameters were tried, with a Xavier uniform initialization \citep{Glorot2010UnderstandingNetworks} with gain 10, giving the best combination of accuracy and calibration. For iNat 2017, a gain value of 1 is used. For more information, refer to Appendix~\ref{app_A:init_lora}. This setting is kept for models across all datasets, including the one with an \glsxtrshort{ast} backbone.

For the same reason, we investigated whether adding noise to the pre-trained parameters of an Explicit Ensemble increases its performance through a higher diversity of members. However, the results did not show any additional benefits beyond what the randomly initialized last layer already provided, hence we did not use that option. For more details, refer to Appendix~\ref{app_B:explicit_init}.

%Due to an issue with an element of the explicit ensemble for the Vision Transformer, an implementation in a similar fashion using a vectorized map was not possible. The issue could, however, not be pinpointed precisely. Thus, the implementation of the explicit ensemble calls the members sequentially in the forward pass. 

\section{Initialization of LoRA-Ensemble Parameters} \label{app_A:init_lora}
Randomness in initialization is a key driver of diversity among ensemble members \citep{Fort2019DeepPerspective}. Therefore, finding the right balance between diversity and overly disrupting parameters is crucial. \citet{Hu2021LoRA:Models} propose using a random Gaussian initialization for $A$ while setting $B$ to zero. This approach results in $\Delta W = BA$ being zero at the start of training. In our experiments, we adopt this pattern by always initializing $B$ to zero while varying the parameters and methods for initializing $A$.
Following the method outlined by \citet{Hu2021LoRA:Models}, our initial experiments concentrated on the Gaussian initialization of $A$, with a mean $\mu=0$ and varying standard deviations.
Additionally, we tested the Xavier uniform initialization \citep{Glorot2010UnderstandingNetworks} using different values for the gain.
All tests were conducted on the CIFAR-100 dataset and subsequently applied to other experiments.
We compared results in terms of accuracy and \glsxtrlong{ece}.

\begin{table}[h]
    %\caption{Accuracy and Expected Calibration Error for different types of initialization with varying distribution parameters.}
    \caption{Accuracy and \glsxtrlong{ece} for different initialization methods and varying distribution parameters for LoRA-Ensemble.}
    \centering
    \begin{tabular}{llll}
        \toprule
        \textbf{Init. Type} & \textbf{Std. / Gain} & \textbf{Accuracy ($\uparrow$)} & \textbf{ECE ($\downarrow$)} \\
        \midrule
        \multirow{7}{*}{Gaussian} & 0.02 & 81.2 & 0.041 \\
         & 0.05 & 81.4 & 0.037 \\
         & 0.1 & 81.7 & 0.035 \\
         & 0.2 & 82.1 & 0.034 \\
         & 0.5 & 82.6 & 0.036 \\
         & 1 & 82.5 & 0.039 \\
         & 2 & 81.7 & 0.046 \\
        \midrule 
        \multirow{7}{*}{Xavier Uniform} & 1 & 81.5 & 0.039 \\
         & 5 & 82.2 & 0.034 \\
         & 10 & 82.4 & 0.034 \\
         & 15 & 82.6 & 0.037 \\
         & 20 & 82.4 & 0.038 \\
         & 30 & 82.2 & 0.043 \\
        \bottomrule
    \end{tabular}
    \label{tab_app:LoRA_init_table}
\end{table}

\begin{figure}[ht]
    \centering
    \subfloat[Gaussian initialization with varying standard deviation.]{\label{subfig_app:lora_init_gaussian}\includegraphics[width=.9\textwidth]{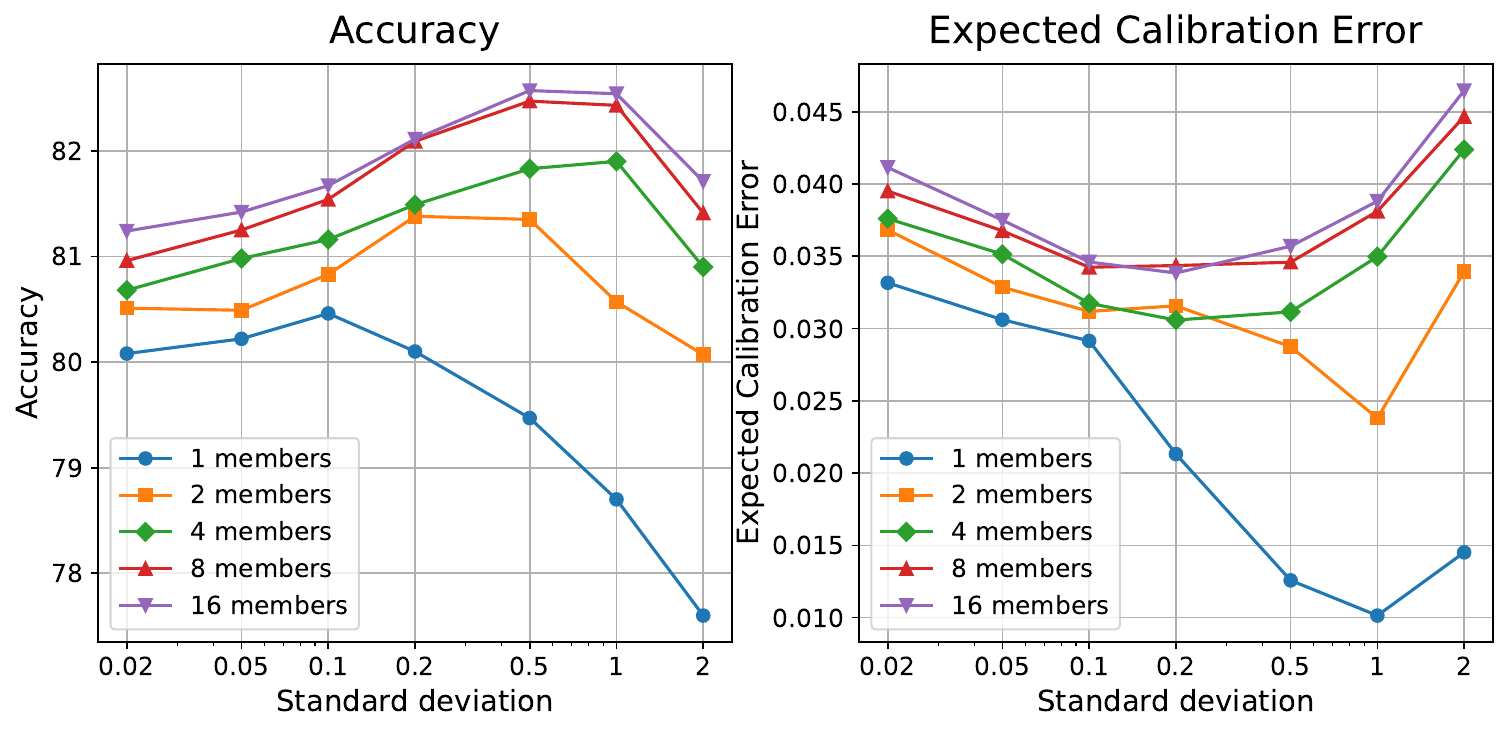}} \\
    \subfloat[Xavier uniform initialization with varying gain]{\label{subfig_app:lora_init_xavier}\includegraphics[width=.9\textwidth]{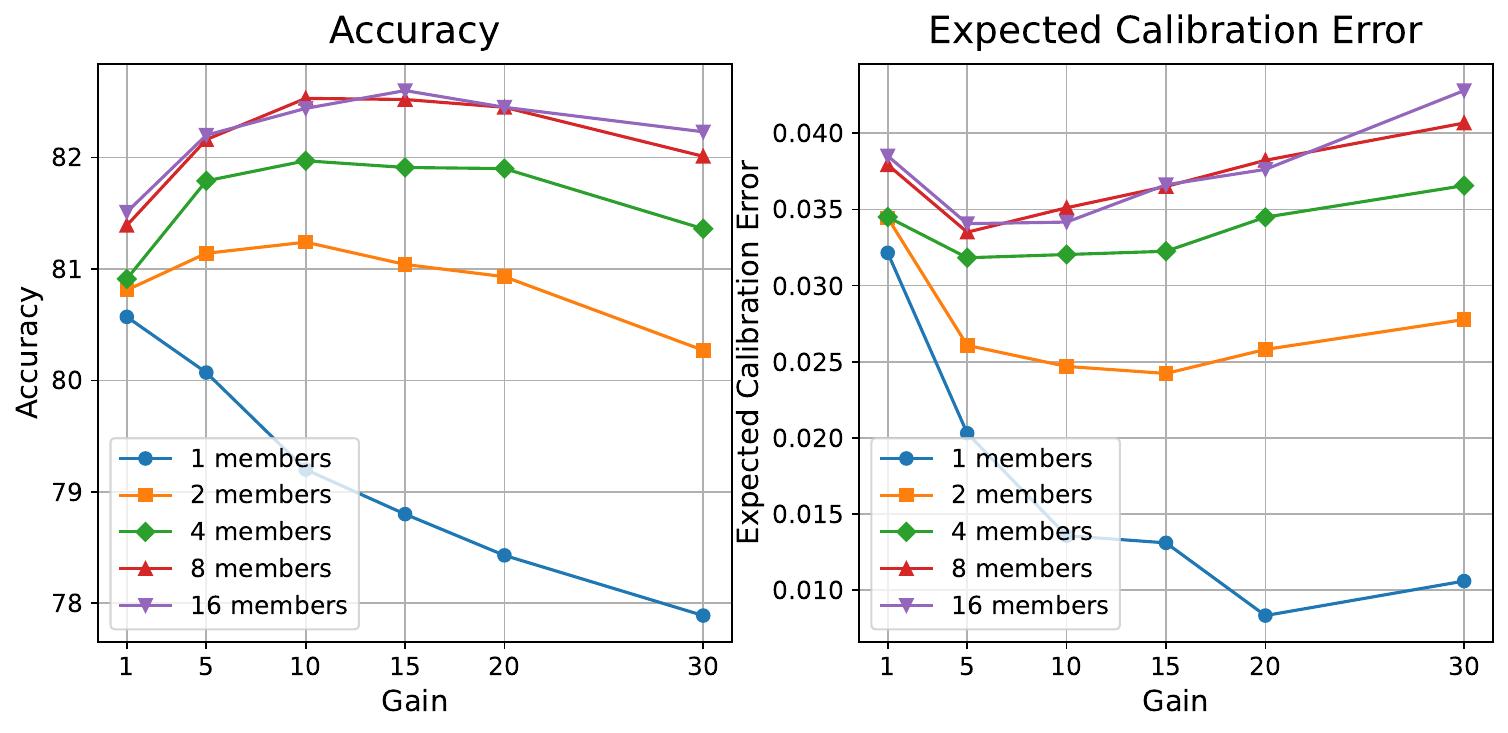}}   
    %\caption{Accuracy and \glsxtrlong{ece} for different initialization methods and varying distribution parameters for changing ensemble size for LoRA-Ensemble.}
    \caption{Accuracy and \glsxtrlong{ece} for different initialization methods and varying distribution parameters across different ensemble sizes for \glsxtrshort{lora}-Ensemble.}
    \label{fig_app:lora_init_xavier_gaussian}
\end{figure}

In Tab.~\ref{tab_app:LoRA_init_table}, the results are quantitatively presented. It is immediately evident that both techniques and all tested parameters perform similarly. While more specialized models may surpass our results in terms of accuracy, our primary focus is on calibration, with the goal of maintaining comparable predictive performance. Visual inspection of the results in Fig.~\ref{fig_app:lora_init_xavier_gaussian} confirms the high similarity among all results.
Choosing a small calibration error while maintaining high accuracy as a decision criterion, both Gaussian initialization with a standard deviation of $0.5$ and Xavier uniform initialization with a gain of $10$ or $15$ are viable candidates. Since a gain of $10$ combines high accuracy with the lowest \glsxtrlong{ece}, we select Xavier uniform initialization with a gain of $10$ for our experiments.

\section{AST Implementation} \label{app_C:AST_implementation}

A different backbone is used for the experiment on the audio dataset. Specifically, we use the \glsxtrfull{ast} following the implementation of \citet{Gong2021AST:Transformer}, with slight modifications to fit our general architecture. Appendix~\ref{app_C:AST_Validation} demonstrates the equivalence of our implementation. In their experiments, \citet{Gong2021AST:Transformer} used two different types of pre-trained weights: one pre-trained on a large image dataset and the other on an audio dataset. For our research, we transfer the weights of a vision transformer model known as \glsxtrshort{deit} \citep{Touvron2020TrainingAttention}, which has been pre-trained on the ImageNet dataset \citep{Deng2009ImageNet:Database}, to the original \glsxtrshort{ast} architecture by \citet{Gong2021AST:Transformer}. The model has 12 layers, uses 768-dimensional patch embeddings, and the multi-head attention modules have 12 heads. This task is considered more challenging than using models pre-trained on audio datasets.

\section{Validation of AST Implementation} \label{app_C:AST_Validation}

The \glsxtrfull{ast} model provided by \citet{Gong2021AST:Transformer} was copied without any changes. However, the training and evaluation pipeline was adapted to fit our architecture. Correspondingly, it was essential to validate the equivalence of our implementation by training a single \glsxtrshort{ast} on the ESC-50 dataset. The results of our model should closely match those provided in \citet{Gong2021AST:Transformer}.

They offer two sets of pre-trained weights: one where the weights of a \glsxtrlong{vit} pre-trained on ImageNet \citep{Deng2009ImageNet:Database} are transferred to \glsxtrshort{ast}, and another where the \glsxtrshort{ast} was pre-trained on AudioSet \citep{Gemmeke2017AudioEvents}. To verify our implementation, we ran it using the settings provided by \citet{Gong2021AST:Transformer} and compared the results, which are summarized in Tab.~\ref{tab_app:AST_comparison}.
The results for both pre-training modes fall within the uncertainty range provided by \citep{Gong2021AST:Transformer}. This suggests that our pipeline yields comparable outcomes, validating our implementation for continued use.

\begin{table}[h]
    \centering
 \caption{Comparison of the results obtained for the \glsxtrshort{ast} as given in \cite{Gong2021AST:Transformer} and those obtained by our implementation. AST-S refers to the \glsxtrshort{ast} pre-trained on ImageNet, and AST-P refers to the AudioSet pre-training. Both results fall within the uncertainty range provided by \cite{Gong2021AST:Transformer}.}
    \begin{tabular}{lcc}
        \toprule
        \textbf{Model} & \textbf{Accuracy \citep{Gong2021AST:Transformer}} & \textbf{Accuracy (our implementation)} \\
        \midrule
        AST-S & $88.7\pm0.7$ & $88.0$ \\
        AST-P & $95.6\pm0.4$ & $95.8$ \\
        \bottomrule        
    \end{tabular}
    \label{tab_app:AST_comparison}
\end{table}

\section{Hyper-parameter Tuning for AST Experiment} \label{app_D:AST_HyperparameterTuning}

The original training settings of the AST-S model in \citep{Gong2021AST:Transformer} utilize a batch size of 48. However, due to the memory constraint of single GPU training on an NVIDIA Tesla A100 with 80 GB memory, replicating a batch size of 48 as in the original publication was infeasible for training an Explicit AST-S Ensemble with 8 members. Consequently, we perform minimal hyper-parameter tuning by employing a batch size of 1 for both the explicit AST-S and the \glsxtrshort{lora} AST-S model, exploring various learning rates. Apart from batch size and learning rate adjustments, all other settings remain consistent with \citep{Gong2021AST:Transformer}.

The hyper-parameter tuning results for the explicit model using a batch size of 1, as shown in Tab.~\ref{tab_app:AST_ExplicitTuning}, demonstrate performance similar to the original implementation with a batch size of 48, allowing for a fair comparison with our method \citep{Gong2021AST:Transformer}. Additionally, Tab.~\ref{tab_app:AST_LoRATuning} showcases the outcomes of tuning the learning rate for our \glsxtrshort{lora} AST-S model.

\begin{table}[ht]
    \centering
    \caption{Single model 5-Fold cross-validation results of AST-S on ESC-50 sound dataset with different learning rates and batch size 1. The model settings selected based on accuracy for the experiments are \textbf{highlighted}.}
    \begin{tabular}[h]{lccc}
        \toprule
        \textbf{Model} & \textbf{Learning rate} & \textbf{Accuracy ($\uparrow$)} & \textbf{ECE ($\downarrow$)} \\
        \midrule
        \textbf{AST-S} & \textbf{0.00001}&$\textbf{88.2}$ & $\textbf{0.0553}$ \\
        AST-S & 0.00005&$81.7$ & $0.0933$ \\
        \bottomrule        
    \end{tabular}
    \label{tab_app:AST_ExplicitTuning}
\end{table}

\begin{table}[ht]
    \centering
    \caption{Single model 5-Fold cross-validation results for our \glsxtrshort{lora} AST-S implementation on ESC-50 sound dataset with different learning rates and batch size 1. The model settings selected based on accuracy for the experiments are \textbf{highlighted}.}
    \begin{tabular}{lccc}
        \toprule
        \textbf{Model} & \textbf{Learning rate} & \textbf{Accuracy ($\uparrow$)} & \textbf{ECE ($\downarrow$)} \\
        \midrule
        LoRA AST-S & 0.00001&$85.6$ & $0.0447$ \\
       \textbf{LoRA AST-S} & \textbf{0.00005}&$\textbf{87.9}$ & $\textbf{0.0487}$ \\
        LoRA AST-S & 0.0001&$84.7$ & $0.0501$ \\
        LoRA AST-S & 0.0005&$24.1$ & $0.0291$ \\
        LoRA AST-S & 0.001&$11.8$ & $0.0295$ \\
        \bottomrule        
    \end{tabular}
    \label{tab_app:AST_LoRATuning}
\end{table}

\section{Computational Cost for AST Models}\label{app_sec:resource_ast}
Similarly to the way we did for the \glsxtrlong{vit} models, we estimate the required resources for \glsxtrshort{ast} models. The resource needs are presented in Tab.~\ref{tab_app:ast_model_resources}.
\begin{table}
    \caption{Parameter counts and computation times for an Explicit Ensemble of 8 \glsxtrshort{ast} models and the corresponding \glsxtrshort{lora}-Ensemble. Training time is the average duration for one epoch on ESC-50, with batch size 1. Inference time is the average duration of a forward pass, with batch size 1.}
    \centering
    \begin{tabular}{lccc}
        \toprule
        \textbf{Method} & \textbf{Parameter overhead} & 
        \textbf{Training time [s]} & \textbf{Inference time [ms]}\\
        \midrule
        Explicit Ensemble & \ \ \ \ \ $8 \times 87\mathrm{M}$ & $517$ & $8 \times 7.3$ \\
        \glsxtrshort{lora}-Ensemble & $1.08 \times 87\mathrm{M}$ & $348$ & \ \ \ \ \ $73.9 $ \\
    \bottomrule
    \end{tabular}
    \label{tab_app:ast_model_resources}
\end{table}
The number of parameters is reported for an ensemble of 8 members, with the $A$ and $B$ matrices in models using \glsxtrshort{lora} having a rank of 16. Training and inference times were measured on a single NVIDIA Tesla A100-80GB \glsxtrshort{gpu}, with a batch size of 1. Training time is given as the average wall clock time per training epoch while training on ESC-50, with 8 ensemble members. Inference time is reported as the average time for a single forward pass of an ESC-50 sample with a batch size of 1. 

As mentioned in Appendix~\ref{app:implementation}, the Explicit Ensemble processes the members sequentially, while \glsxtrshort{lora}-Ensemble is parallelized. However, fully parallelizing the training of \glsxtrshort{ast} models causes memory issues, so chunking was introduced. Thus, in \glsxtrshort{lora}-Ensemble models, the pass through the backbone runs in parallel, while \glsxtrshort{lora} modules are called sequentially. This also explains the significantly higher inference time compared to the results in Sec.~\ref{subsec:computationl_cost}.
Additionally, the one-time delay incurred by PyTorch's \emph{vmap} function causes \glsxtrshort{lora}-Ensemble to be slightly slower at inference time.

\section{Baselines} 
\subsection{Hyperparameter Tuning for MC Dropout} \label{app_E:MCDropout_HyperparameterTuning}

\begin{figure}[ht]
    \centering
    {\label{subfig_app:dropout_rate}\includegraphics[width=.9\textwidth]{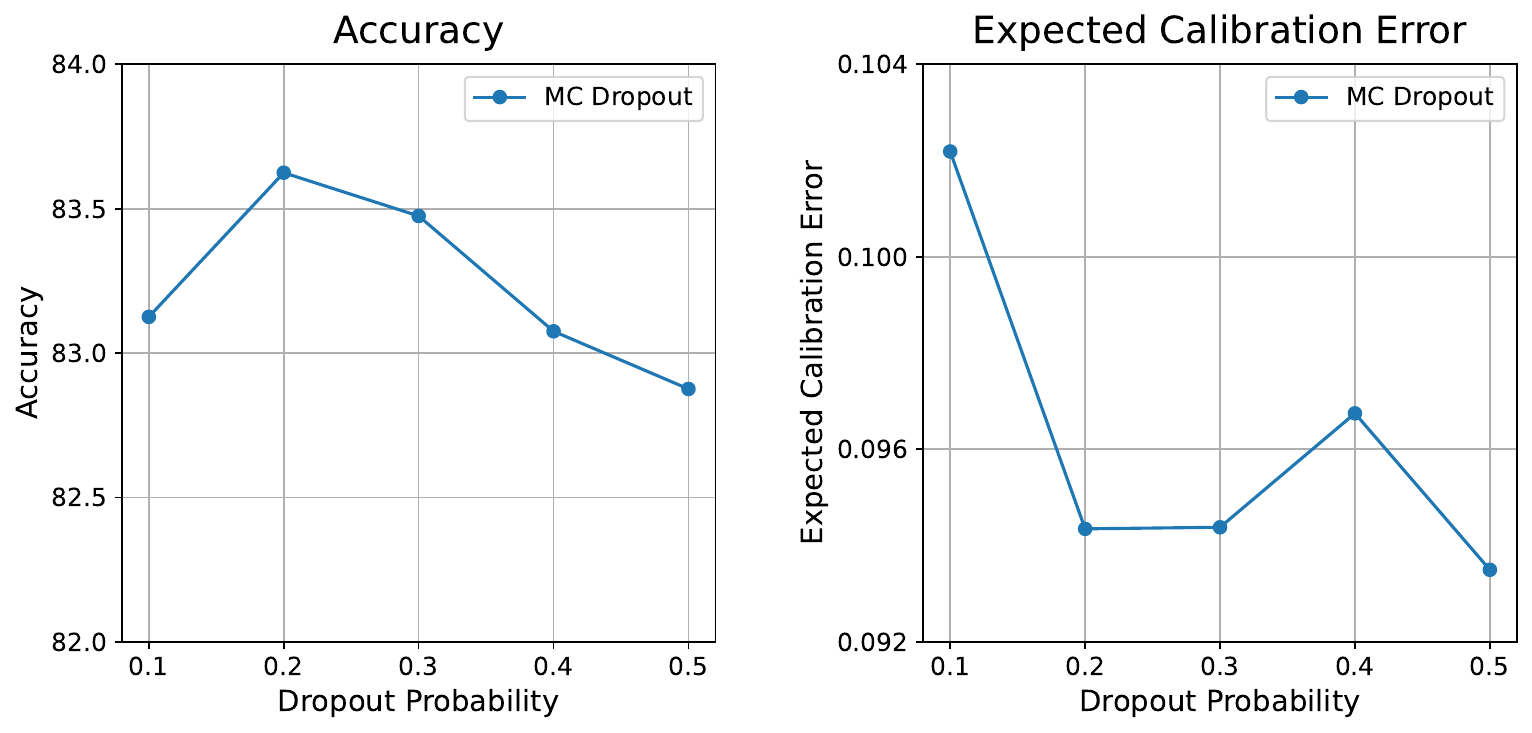}} \\
    \caption{Accuracy and \glsxtrlong{ece} for different dropout probabilities methods for MC Dropout on HAM10000 dataset.}
    \label{fig_app:dropout_rate}
\end{figure}

We conducted an analysis to determine the impact of dropout probability on the accuracy and calibration of the \glsxtrshort{vit} with Monte Carlo dropout. Fig.~\ref{fig_app:dropout_rate} displays the accuracy and \glsxtrshort{ece} scores for various dropout probabilities. The experiment is carried out on the HAM10000 dataset with 16 members. Our findings show that a dropout probability of $0.2$ offers a good balance between accuracy and calibration. 

\subsection{Initialization of Explicit Ensemble Parameters} \label{app_B:explicit_init}
A pre-trained \glsxtrlong{vit} model is the backbone for our computer vision experiments. Correspondingly, the parameters of all members in an Explicit Ensemble are initialized to the same values across members. Initialization is a primary driver of diversity in ensemble members \citep{Fort2019DeepPerspective}. Hence, it is crucial to study the effect of noise in the parameter initialization on the calibration of the resulting ensemble.
In the case of pre-trained model weights not having been trained on a dataset with the same number of classes, the last layer of all models is replaced completely. This means that regardless of the ensemble technique used, the weights of the last layer, which is responsible for classification, will vary across members. This variation in the weights of the classification layer is expected to contribute significantly to the diversity of the members.
Nonetheless, we studied the impact of adding noise to the parameters of an Explicit Ensemble. This was done using the following formula:
\begin{equation}
    W_{\mathrm{new}} = W + \alpha \cdot dW\,,
\end{equation}
where $dW \sim \mathcal{N}(0,\sigma_W)$. Here $\alpha$ is a scale factor to control the amount of noise and $\sigma_W$ is the standard deviation of the parameters within a weight matrix. This was applied to all weight matrices separately.

It is expected that the initial layers of a neural network will learn basic features, while the later layers will include dataset-specific properties. Therefore, it is assumed that adding noise to the later layers would increase diversity while maintaining pre-training. However, adding noise to the earlier layers might disrupt pre-training more significantly, especially with smaller datasets, as these parameters may not converge to meaningful values again. To address this, an experiment was set up where noise was added only to the last encoder layers of the model, increasing the number of affected encoder layers gradually. Additionally, several different noise scales $\alpha$ were tried, ranging from $1$ to $0.0001$.
In the presented experiment, the last classification layer is initialized using PyTorch's default method for linear layers. At the time of writing it is as follows:
\begin{align}
    W_{\mathrm{init}} &= \mathrm{Unif}\left(-\sqrt{5}\cdot\sqrt{\frac{3}{fan\_in}}, \sqrt{5}\cdot\sqrt{\frac{3}{fan\_in}}\right) \\
    B_{\mathrm{init}} &= \mathrm{Unif}\left(-\sqrt{\frac{1}{fan\_in}},\sqrt{\frac{1}{fan\_in}}\right).
\end{align}
Here $W$ specifies the weight matrix and $B$ is the bias.
Experiments are conducted on the CIFAR-100 dataset.

\subsubsection{Results}
\begin{figure}[ht]
    \centering
    \subfloat[Single model]{\label{subfig_app:noise_study_1_members}\includegraphics[width=.9\textwidth]{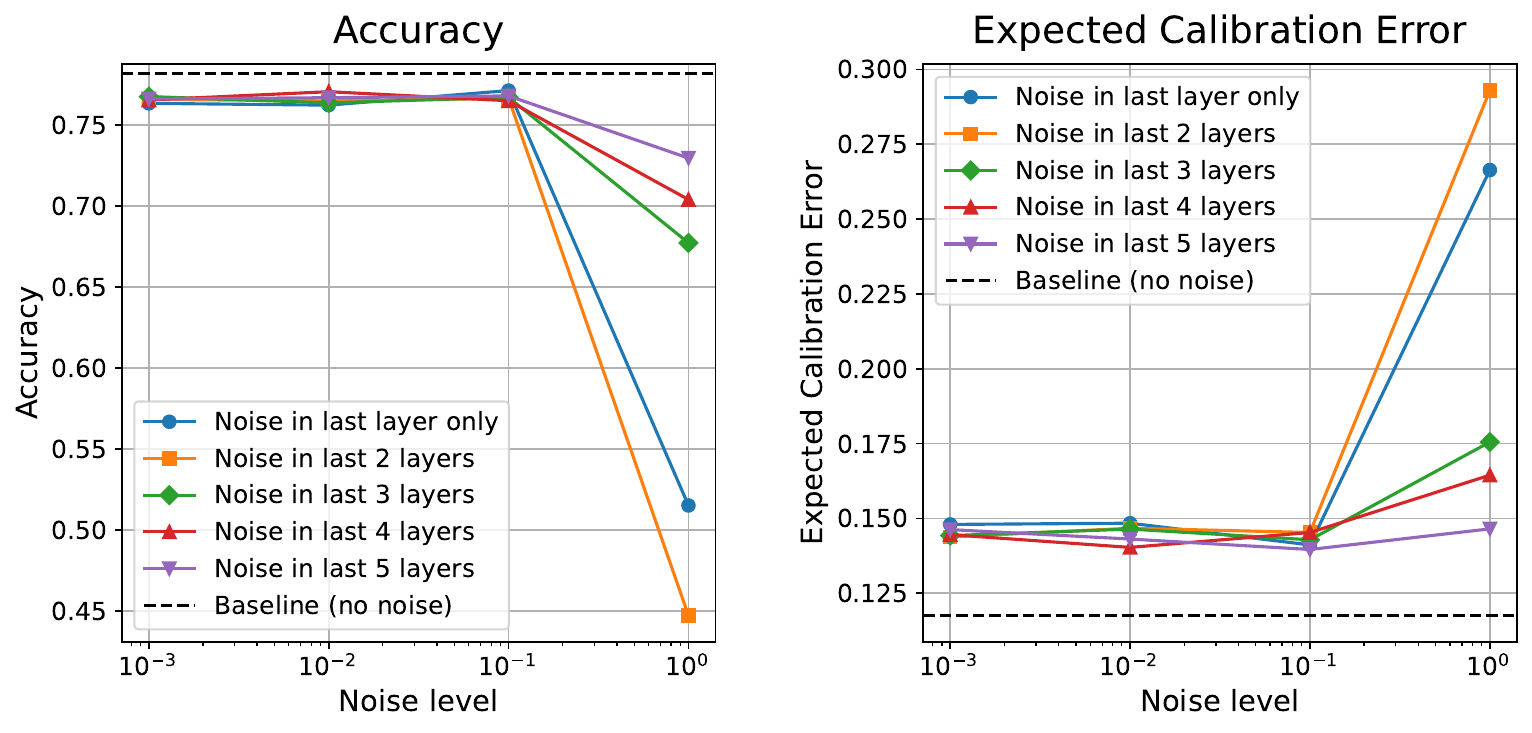}} \\
    \subfloat[16 ensemble members]{\label{subfig_app:noise_study_16_members}\includegraphics[width=.9\textwidth]{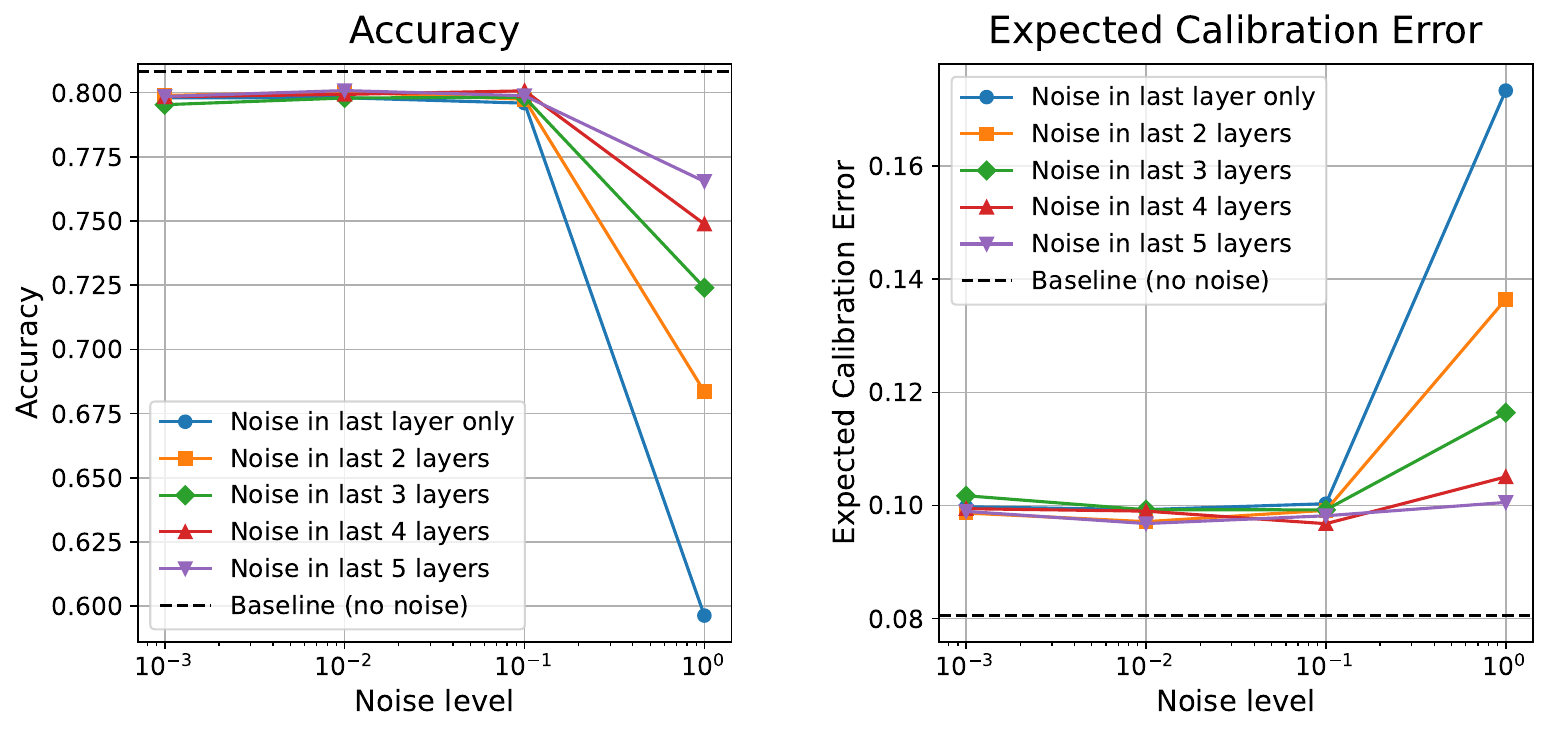}}   
    \caption{Accuracy and \glsxtrlong{ece} for different noise levels across varying numbers of layers for the Explicit Ensemble. The baseline with no noise is indicated by a dashed black line.}
    \label{fig_app:noise_study}
\end{figure}

The most important metrics for this section are accuracy and \glsxtrlong{ece}. The results for adding noise to the last layer up to the last five layers are summarized in Fig.~\ref{fig_app:noise_study}. Fig.~\ref{subfig_app:noise_study_1_members} depicts the results for a single model, while Fig.~\ref{subfig_app:noise_study_16_members} shows the results for an ensemble of 16 members.

It is evident that none of the experiments surpass the baseline of not using any additional noise beyond the random initialization of the last classification layer. After the last five layers, the results become uninteresting, as they do not vary significantly from those shown in the plots. Therefore, the presentation is truncated at five layers.
Based on the presented results, no additional noise is injected into the Explicit Ensemble, and only the last layer initialization is varied.

\subsection{Batch-Ensemble}\label{app_sec:batchEnsemble}

\subsubsection{Implementation}

Probably the closest method to LoRA-Ensemble is Batch-Ensemble introduced in \citet{Wen2020BatchEnsemble:Learning}.
Batch-Ensemble was originally developed for MLPs, but we extend it to self-attention networks as an implicit ensemble baseline. The methodology draws inspiration from our development of LoRA-Ensemble, as the two implementations share many similarities. The primary difference lies in the parametrization of ensemble members.
For each projection matrix (query, key, value, and output), we define one shared full-rank trainable matrix initialized with the pre-trained weights of the base network, along with two additional trainable vectors, \( r \) and \( s \), which are specific to each ensemble member. The projection matrix for ensemble member \( i \) is defined as:
\begin{equation} \label{eq:Batchensemble_member}
W_i = W_{\text{shared}} \circ r_i s_i^{T},
\end{equation}
where \( W_{\text{shared}} \) is the shared trainable matrix, and \( \circ \) denotes element-wise multiplication.
Within each transformer block, a unique forward pass is computed for each ensemble member \( i \):
\begin{equation}
h_i = W_i x,
\end{equation}
resulting in \( N \) different predictions \( T_{\theta_i}(X) \) for a given input \( X \). The final ensemble prediction is obtained by averaging the individual predictions:
\begin{equation}
\mathbb{E}[Y | X] = \frac{1}{N} \sum_{i=1}^{N} T_{\theta_i}(X).
\end{equation}
The forward pass for the Batch-Ensemble layer with shared weights is implemented as shown in Listing~\ref{listing:batch_ensemble}.

\begin{lstlisting}[language=Python, caption=Pytorch forward pass for Batch-Ensemble layer, label=listing:batch_ensemble]
def forward(self, x):
    """
    Forward pass for the Batch-Ensemble layer
    """
    # Step 1: Compute the ensemble member-specific weights
    r = self.r.weight  # Shape: [1, dim]
    s = self.s.weight  # Shape: [out_dim, 1]
    W_rs = s @ r       # Shape: [out_dim, dim]

    # Step 2: Combine with the shared weight
    W_combined = self.shared_w * W_rs  # Element-wise multiplication

    # Step 3: Compute the output for a specific ensemble member
    out = x @ W_combined.T  # x must have shape [batch_size, dim]

    return out
\end{lstlisting}

The \( r \) and \( s \) vectors are initialized from a Gaussian distribution centered around 1, specifically \( r, s \sim \mathcal{N}(1, \sigma^2) \), where \( \sigma^2 \) controls the variance. We empirically set \( \sigma^2 = 0.02 \). This initialization ensures that at the beginning of the training, the combined projection matrix for each ensemble member remains close to the pre-trained weights of the shared matrix, preventing disruption of learned pre-trained weights.
The implementation and training details followed the LoRA-Ensmeble approach; for details, refer to \ref{app:implementation} and \ref{subsec:2trainingsettings}.

\subsubsection{Why LoRA-Ensemble Outperforms Batch-Ensemble} \label{why_batchensemble_inferior}

Both LoRA-Ensemble and Batch-Ensemble leverage shared weights with member-specific low-rank modifications to enable efficient ensembling. The key difference lies in their parameterization: LoRA-Ensemble uses additive low-rank updates, while Batch-Ensemble applies element-wise multiplicative scaling. Despite the conceptual similarity between the two methods, Batch-Ensemble performs significantly worse than LoRA-Ensemble in both accuracy and calibration, as demonstrated in Tab.~\ref{tab:model_performance_cifar100} and Tab.~.\ref{tab:model_performance_HAM10000} This performance gap persists even when applied to non–self-attention architectures such as convolutional neural networks, which were the original target application of Batch-Ensemble, as shown in Tab.~\ref{tab_app:lora_cnn}.

To clarify this difference, we examine the gradients of the member-specific parameters. For LoRA-Ensemble, the layer output is:
\[
h_i = W_{\text{shared}} \cdot x + B_i A_i x,
\]
with gradients:
\begin{align}
    \frac{\partial \mathcal{L}}{\partial B_i} &= \delta \cdot A_i \cdot x, \\
    \frac{\partial \mathcal{L}}{\partial A_i} &= \delta \cdot B_i \cdot x,
\end{align}
where \( \delta = \frac{\partial \mathcal{L}}{\partial h_i} \).

For Batch-Ensemble, the output is:
\[
h_i = (W_{\text{shared}} \odot r_i s_i^{T})x,
\]
with gradients:
\begin{align}
    \frac{\partial \mathcal{L}}{\partial s_i} &= \delta \cdot (W_{\text{shared}} \odot r_i)\cdot x, \\
    \frac{\partial \mathcal{L}}{\partial r_i} &= \delta \cdot (W_{\text{shared}} \odot s_i^T)\cdot x.
\end{align}

In Batch-Ensemble, the gradient updates for \( r_i \) and \( s_i \) are directly scaled by the shared weights \( W_{\text{shared}} \), which can constrain the learning dynamics and reduce the independence of ensemble members. This scaling introduces sensitivity to the magnitude and sparsity of \( W_{\text{shared}} \), potentially limiting the diversity of the ensemble.

We define \textbf{Batch-Ensemble++} by modifying the original Batch-Ensemble algorithm, replacing the point-wise multiplication operation with an addition operation as follows:

\begin{equation} \label{eq:Batchensemble_member2}
W_i = W_{\text{shared}} + r_i s_i^{T}.
\end{equation}

In this case, the r and s vectors are also initialized from a Gaussian distribution but centered around 0.

We compare the performance of Batch-Ensemble++, the original Batch-Ensemble, and LoRA-Ensemble in Tab.~\ref{tab:batch++}. Batch-Ensemble++ significantly outperforms the original Batch-Ensemble in both accuracy and uncertainty calibration. However, its performance does not reach that of LoRA-Ensemble.
%
% We attribute this performance gap to following key differences between the methods:  
%     (i) By design, Batch-Ensemble restricts ensemble-specific parameters to rank-1 matrices, which inherently limits the expressive power of individual ensemble members.
%     (ii) The shared pre-trained matrix $W_{\text{shared}}$ is not kept frozen in Batch-Ensemble, which might disrupt the learned pre-trained weights and/or shared weights limit the learning diverse enough set of ensemble parameters. These issues might arise from the coupled learning dynamics between $W_{\text{shared}}$ and the ensemble-specific parameters $r$ and $s$.
%     (iii) Differences in parameter initialization may also contribute.

We attribute this performance gap to the following key differences between the methods:
\begin{itemize}
    \item \textbf{Limited Expressiveness:} Batch-Ensemble restricts its ensemble-specific parameters to rank-1 matrices, inherently limiting the expressive power of individual ensemble members.
    \item \textbf{Coupled Learning Dynamics:} In Batch-Ensemble, the shared pre-trained matrix \(W_{\text{shared}}\) is not kept frozen. This design choice can disrupt the learned pre-trained weights and may restrict the ability of the ensemble-specific parameters \(r\) and \(s\) to learn a sufficiently diverse set of features. A similar effect was observed in LoRA-Ensemble when the backbone was also updated during training; see the Appendix~\ref{app:joint_inat} for details.
    \item \textbf{Initialization Variations:} Differences in parameter initialization may also contribute to the performance gap.
\end{itemize}

%
%These limitations explain why Batch-Ensemble++ shows improvements over the original Batch-Ensemble but does not match the performance of LoRA-Ensemble.

\begin{table}[t]
    \caption{Model performance on the CIFAR-10 dataset for the compared methods. Ensembles have 4 members. Best score for each metric in \textbf{bold},  second-best \underline{underlined}.}
    \centering
    \begin{tabular}{lccccc}
        \toprule
        \textbf{Method} & \textbf{Accuracy ($\uparrow$)} & \textbf{F1 ($\uparrow$)} & \textbf{ECE ($\downarrow$)} & \textbf{NLL ($\downarrow$)} & \textbf{Brier ($\downarrow$)} \\
        \midrule
        
        Single Network & $\underline{92.8}$ & $\underline{92.8}$ & $0.051$ & $0.333$ & $\underline{0.120}$ \\
         \midrule
         Batch-Ensemble & $88.5$ & $88.5$& $0.046$ & $0.345$ & $0.171$\\
        
        Batch-Ensemble++ & $91.7$ & $91.7$& $\underline{0.033}$ & $\underline{0.260}$ & $0.128$\\
        
        \midrule
        LoRA-Ensemble & $\textbf{95.6}$ & $\textbf{95.6}$ & $\textbf{0.003}$ & $\textbf{0.133}$ & $\textbf{0.067}$\\
        \bottomrule
    \end{tabular}
    \label{tab:batch++}
\end{table}

\subsection{Snapshot Ensemble Implementation Details}\label{app_sec:snapshot_implementation}
Snapshot Ensemble \citep{Huang2017SnapshotFree}, in its pure form, consists of training a single model with cycling learning and taking snapshots every few epochs. This can make it hard, however, for the model to converge to anything meaningful within the low number of epochs available for training per snapshot. Therefore, Snapshot Ensemble was modified slightly, by first letting training run for a number of epochs, without any cycling of the learning rate. After this burn-in period the learning rate is at 0 and a first snapshot is taken. The remaining number of epochs is split evenly. If the remaining number of epochs is not divisible by the desired number of ensemble members, the burn-in period is extended until it is. For the HAM10000 dataset training is left at 65 epochs, with 20 burn-in epochs. For CIFAR-10 and CIFAR-100 using only 16 epochs would only leave 1 epoch per cycle for bigger models. Therefore, training is extended to 30 epochs with a burn-in period of 15 epochs. In our experiments (Tables~\ref{tab:model_performance_cifar100} and~\ref{tab:model_performance_HAM10000}), snapshot ensembles outperformed a single network in both calibration and accuracy, but underperformed Explicit Ensemble and LoRA-Ensemble, consistent with prior work~\citep{LEE202522, Thuy2025}.

\subsection{Last Layer Ensemble}\label{app_sec:lle} 

Last Layer Ensemble is a simple baseline where multiple independent classification heads are trained on a shared frozen backbone. Diversity arises purely from different random initializations of the heads, making it a minimal-cost approach to implicit ensembling.

Given a frozen pre-trained backbone $\phi$ with feature dimension $d=768$ (ViT-B/32) and $M$ classification heads $\{h_i\}_{i=1}^{M}$, the prediction for ensemble member $i$ is $f_i(\mathbf{x}) = h_i(\phi(\mathbf{x}))$. The ensemble prediction is obtained by averaging softmax outputs across members. Each head is initialized using a Gaussian distribution $W_i \sim \mathcal{N}(0, \sigma_{\text{init}}^2 \cdot \mathbf{I})$ with $\sigma_{\text{init}} = 0.01$ and bias initialized to zero. To ensure reproducibility while maintaining diversity, head $i$ uses random seed $42 + i$.

The training objective is the average cross-entropy loss across all members: $\mathcal{L} = \frac{1}{M} \sum_{i=1}^{M} \mathcal{L}_{\text{CE}}(h_i(\phi(\mathbf{x})), y)$. Since the backbone is frozen, gradients only flow through the classification heads. We use 16 ensemble members trained with AdamW optimizer (learning rate $10^{-4}$, weight decay $0.01$) for 16 epochs on CIFAR-100 with batch size 32, 500 warmup steps, cosine learning rate schedule, and gradient clipping at 1.0.

\begin{lstlisting}[language=Python, caption={Last-Layer Ensemble}, label={lst:lastlayer}]
class LastLayerEnsemble(nn.Module):
    def __init__(self, num_classes, num_members=16, init_std=0.01):
        super().__init__()
        self.backbone = vit_b_32(weights='IMAGENET1K_V1')
        self.backbone.heads = nn.Identity()
        self.backbone.eval()  # Frozen
        
        # Create M heads with different seeds
        self.heads = nn.ModuleList([
            nn.Linear(768, num_classes) for _ in range(num_members) ])
        for i, head in enumerate(self.heads):
            torch.manual_seed(42 + i * 1000)
            nn.init.normal_(head.weight, std=init_std)
            nn.init.zeros_(head.bias)
    
    def forward(self, x):
        features = self.backbone(x)  # [B, 768]
        return [head(features) for head in self.heads]
\end{lstlisting}

The main limitation of LLE is that all ensemble members share identical features, fundamentally constraining diversity to the final linear transformation. This prevents the method from expressing uncertainty about feature relevance and probably leads to heads converging to similar solutions during training.

% ============================================================================
% APPENDIX Y: EPISTEMIC NEURAL NETWORKS (EPINET)
% ============================================================================

\subsection{Epistemic Neural Networks (EpiNet)}\label{appendix:epinet}

Epistemic Neural Networks (EpiNets)~\citep{osband2022epistemic} estimate uncertainty by augmenting a base network with a small auxiliary network (the epinet) conditioned on a random epistemic index $\mathbf{z} \sim \mathcal{N}(\mathbf{0}, \mathbf{I}_{D_z})$. The prediction is:
\begin{equation}
    f_\theta(\mathbf{x}, \mathbf{z}) = \mu_\zeta(\mathbf{x}) + \sigma_\eta(\text{sg}[\phi_\zeta(\mathbf{x})], \mathbf{z}),
\end{equation}
where $\mu_\zeta(\mathbf{x})$ is the base network output, $\phi_\zeta(\mathbf{x}) \in \mathbb{R}^{768}$ are features from the penultimate layer, $\text{sg}[\cdot]$ denotes stop-gradient (preventing gradients from flowing back to the base network through the epinet path), and $\sigma_\eta$ is the epinet. Following the original paper, we decompose the epinet into learnable and prior components: $\sigma_\eta = \sigma^L_\eta + \alpha \cdot \sigma^P$, where $\sigma^P$ is a frozen randomly-initialized network providing initial diversity and $\alpha=1.0$.

The epinet architecture is an MLP that takes the concatenation of features $\phi \in \mathbb{R}^{768}$ and epistemic index $\mathbf{z} \in \mathbb{R}^{D_z}$ as input. The MLP outputs a matrix $\mathbf{M} \in \mathbb{R}^{D_z \times C}$, which is then contracted with the epistemic index to produce class logits: $\sigma_\eta(\phi, \mathbf{z})_c = \sum_{k=1}^{D_z} M_{k,c} \cdot z_k = \mathbf{M}^\top \mathbf{z}$. This design ensures the epinet output is linear in $\mathbf{z}$, enabling smooth interpolation across the epistemic index space. The MLP uses two hidden layers of 256 units each with ReLU activations.

We use $D_z = 10$ and $M=16$ ensemble members, each with its own learnable epinet (weights initialized near zero with std $0.01$) and frozen prior network (initialized with seed $42 + i \times 1000$). The base network is a trainable ViT-B/32. Training minimizes: $\mathcal{L} = \frac{1}{M} \sum_{i=1}^{M} \mathcal{L}_{\text{CE}}(f_\theta(\mathbf{x}, \mathbf{z}_i), y)$ with freshly sampled $\mathbf{z}_i$ per member. We use the same training setup as Last Layer Ensemble: AdamW (learning rate $10^{-4}$, weight decay $0.01$), batch size 32, 16 epochs, 500 warmup steps, cosine decay, and gradient clipping at 1.0.

\begin{lstlisting}[language=Python, caption={PyTorch implementation of EpiNet.}, label={lst:epinet}]
class Epinet(nn.Module):
    """Epinet: MLP outputting matrix M, contracted with z to get logits."""
    def __init__(self, feature_dim, num_classes, epistemic_dim, hidden=256):
        super().__init__()
        self.epistemic_dim = epistemic_dim
        self.num_classes = num_classes
        self.mlp = nn.Sequential(
            nn.Linear(feature_dim + epistemic_dim, hidden),
            nn.ReLU(),
            nn.Linear(hidden, hidden),
            nn.ReLU(),
            nn.Linear(hidden, epistemic_dim * num_classes))
    
    def forward(self, features, z):
        x = torch.cat([features, z], dim=-1)              # [B, 768+D_z]
        M = self.mlp(x).view(-1, self.epistemic_dim, self.num_classes)#[B, D_z, C]
        return torch.bmm(M.transpose(1, 2), z.unsqueeze(-1)).squeeze(-1)  # [B, C]

class EpiNet(nn.Module):
    def __init__(self, num_classes=100, num_members=16, epistemic_dim=10,
                 prior_scale=1.0, prior_seed_start=42):
        super().__init__()
        self.backbone = vit_b_32(weights=ViT_B_32_Weights.IMAGENET1K_V1)
        self.backbone.heads = nn.Linear(768, num_classes)  # Trainable base
        
        self.learnable_epinets = nn.ModuleList([
            Epinet(768, num_classes, epistemic_dim) 
            for _ in range(num_members)])
        self.prior_networks = nn.ModuleList([
            self._frozen_epinet(768, num_classes, epistemic_dim,
                               seed=prior_seed_start + i * 1000)
            for i in range(num_members)])
        self.prior_scale = prior_scale
    
    def forward(self, x, z_samples=None):
        features, base_output = self.extract_features(x)
        features_sg = features.detach()  # Stop gradient
        if z_samples is None:
            z_samples = torch.randn(len(self.learnable_epinets), 
                                   x.size(0), self.epistemic_dim, device=x.device)
        outputs = []
        for i, (learnable, prior) in enumerate(
                zip(self.learnable_epinets, self.prior_networks)):
            epinet_out = learnable(features_sg, z_samples[i]) \
                       + self.prior_scale * prior(features_sg, z_samples[i])
            outputs.append(base_output + epinet_out)
        return outputs
\end{lstlisting}

\subsection{FiLM-Ensemble and Other Implicit Baselines: Practical Challenges}\label{app_sec:imlicit_baseline}

Many implicit ensemble methods, such as those proposed in \citep{Wen2020BatchEnsemble:Learning, Turkoglu2022FiLM, Durasov2020MasksemblesEstimation, Havasi2020TrainingPrediction}, are architecture-specific and predominantly designed for MLPs or CNNs. As a result, adapting these techniques to transformer architectures presents significant challenges, since transformers' computation structure is quite different than MLPs and CNNs.

In particular, we attempted to implement FiLM-Ensemble~\citep{Turkoglu2022FiLM} on a self-attention network, given the promising results reported by its authors. However, the authors themselves noted that applying FiLM-Ensemble to transformers is not straightforward, mainly because transformers rely on LayerNorm, whereas FiLM-Ensemble was developed with BatchNorm in mind. Our experiments confirmed that directly using BatchNorm in transformers led to notable performance degradation. We explored several approaches to adapt LayerNorm, but the most effective results were achieved by fixing all affine parameters for each ensemble member. This allowed for slight initial variations to introduce randomness and diversity, while keeping the variation among members minimal. The results, summarized in Tab.~\ref{tab:film_results}, show that increasing the ensemble size slightly improved accuracy, though the Expected Calibration Error (ECE) fluctuated without consistent improvement. In fact, when using larger ensemble sizes, such as 8 or 16, both accuracy and calibration worsened across all settings we tested.

\begin{table}[h]
    \caption{Performance of FiLM-Ensemble for Vision Transformer (ViT) on CIFAR-10. Increasing the ensemble size slightly improves accuracy, but ECE fluctuates without showing consistent improvement.
}
    \centering
    \begin{tabular}{ccc}
        \toprule
        \textbf{\# ensemble members} & \textbf{Accuracy ($\uparrow$)} & \textbf{ECE ($\downarrow$)}\\
        \midrule
        1 & 90.54 & 0.0286 \\
        2 & 91.18 & \textbf{0.0269} \\
        4 & \textbf{91.23} & 0.0289 \\
        \bottomrule
    \end{tabular}
    \label{tab:film_results}
\end{table}

\subsection{Bayesian LoRA} \label{app:bayes_details}

Bayes-LoRA~\citep{yang2024bayesian} introduces a Bayesian approach on the LoRA adapter parameters by fitting a Gaussian posterior around the maximum a posteriori (MAP) estimate of the fine-tuned model. In practice, this means we first obtain a standard LoRA fine-tuned network and then apply a Laplace approximation over its adapter weights. To make this tractable at scale, Bayes-LoRA relies on a Kronecker-factored approximation of the Hessian, which allows efficient estimation of the posterior covariance. The result is a Bayesian model that can capture uncertainty while remaining computationally efficient compared to traditional Bayesian neural networks.

We evaluate Bayes-LoRA on the SST-2 sentiment classification task using a BERT base uncased~\citep{socher2013recursive, Devlin2019BERTPO} backbone. The method is applied in a post-hoc fashion after fine-tuning. 
The original implementation of~\citep{yang2024bayesian} used\footnote{\url{https://github.com/MaximeRobeyns/bayesian_lora}}.
For MAP training, we follow the standard LoRA setup with a learning rate of $5 \times 10^{-5}$, training for 3 epochs with batch size 16. The LoRA rank is set to 64, identical to the LoRA-Ensemble. The prior variance is chosen as $10^{-3}$, since larger values tend to degrade performance. To balance computational efficiency and uncertainty estimation, the number of posterior samples (i.e., ensemble members) is fixed at 512. For the Kronecker-factored approximation, we use $n_{\text{kfac}} = 10$.  

In terms of results, Bayes-LoRA falls short of LoRA-Ensemble and Explicit Ensemble methods in predictive performance, measured by accuracy and F1. This observation is in line with previous findings in the literature, such as~\citep{daxberger2021laplace}. We attribute this limitation to the reliance on a local Gaussian approximation around a single MAP solution, in contrast to the diversity gained through sufficiently independent ensemble members. However, the main strength of Bayes-LoRA lies in its ability to capture predictive uncertainty effectively, reaching a level comparable to both LoRA-Ensemble and Explicit Ensemble.  Detailed results can be found in Tab.~\ref{tab_app:model_performance_Bert} in Appendix~\ref{app:sst2}.

From an efficiency perspective, Bayes-LoRA requires significantly more computation at inference: evaluating a single test example takes roughly 250 ms (512 posterior samples are used), compared to 22.7 ms (Tab.~\ref{tab:model_resources}) for LoRA-Ensemble (16 members are used). This overhead makes Bayes-LoRA impractical for real-time applications but potentially valuable in settings where predictive uncertainty is crucial and strict latency constraints are less relevant.

\subsection{SNGP on Vision Transformers}\label{app:sngp}

Spectral-normalized Neural Gaussian Process (SNGP)~\citep{liu2020simple} is an efficient single-model uncertainty estimation method that achieves competitive calibration with deep ensembles while requiring only a single forward pass at inference. However, SNGP was originally designed and validated exclusively on residual convolutional architectures (e.g., Wide-ResNet)~\citep{liu2022simple}, and its theoretical guarantees do not extend to attention mechanisms. %Specifically, \citet{kim2021lipschitz} proves that dot-product self-attention has an unbounded Lipschitz constant, violating the fundamental assumption underlying SNGP's distance-awareness. They propose an alternative L2 self-attention that satisfies the Lipschitz condition; we also include this variant as a baseline in our HAM10000 experiments (see Tab.~\ref{tab:model_performance_HAM10000}).

\subsubsection{Method Overview}

SNGP combines two architectural modifications to standard neural networks:

\paragraph{Spectral Normalization.} Hidden layers are constrained to satisfy the bi-Lipschitz property by applying spectral normalization~\citep{miyato2018spectral}, which bounds the spectral norm of weight matrices:
\begin{equation}
    W_{\text{SN}} = \frac{W}{\sigma(W)} \cdot c,
\end{equation}
where $\sigma(W)$ denotes the largest singular value and $c < 1$ is the spectral norm bound (typically 0.95). This ensures that the network preserves distances in the input space, which is critical for meaningful uncertainty estimation, inputs far from the training distribution should map to representations that are correspondingly distant.

\paragraph{Gaussian Process Output Layer.} The classification head is replaced with a Gaussian Process layer approximated via Random Fourier Features (RFF)~\citep{rahimi2007random}. For an RBF kernel, the feature map is:
\begin{equation}
    \phi(x) = \sqrt{\frac{2}{D}} \cos(Wx + b),
\end{equation}
where $W \in \mathbb{R}^{d \times D}$ with entries sampled from $\mathcal{N}(0, 1/\ell^2)$ for length scale $\ell$, and $b \sim \text{Uniform}(0, 2\pi)$. The GP posterior is computed via Laplace approximation, maintaining a precision matrix $\Lambda$ that is updated during training:
\begin{equation}
    \Lambda \leftarrow \Lambda + \phi(x)\phi(x)^\top.
\end{equation}
At inference, predictive uncertainty is obtained through mean-field approximation with factor $\pi/8$ or Monte Carlo sampling from the approximate posterior.

\paragraph{Why SNGP Fails on Transformers.} The method's theoretical guarantees rely on the bi-Lipschitz property of residual connections in CNNs. However, \citet{kim2021lipschitz} prove that dot-product self-attention has an unbounded Lipschitz constant, violating the fundamental assumption underlying SNGP's distance-awareness. They propose an alternative L2 self-attention that satisfies the Lipschitz condition; we include this variant as a baseline in our HAM10000 experiments (Table~\ref{tab:model_performance_HAM10000}).

\subsubsection{Training Details}

We implement SNGP for Vision Transformer (ViT-B/32) pretrained on ImageNet-1K using PyTorch. Table~\ref{tab:sngp_hyperparams} summarizes the hyperparameters.

\begin{table}[h]
\centering
\caption{SNGP-ViT hyperparameters for CIFAR-100.}
\label{tab:sngp_hyperparams}
\small
\begin{tabular}{ll}
\toprule
\textbf{Hyperparameter} & \textbf{Value} \\
\midrule
\multicolumn{2}{l}{\textit{Backbone}} \\
Architecture & ViT-B/32 \\
Pretrained weights & ImageNet-1K \\
Hidden dimension & 768 \\
Backbone trainable & Yes \\
\midrule
\multicolumn{2}{l}{\textit{Spectral Normalization}} \\
Spectral norm bound ($c$) & 0.95 \\
Power iterations & 1 \\
Layers normalized & Attention + MLP (also tested MLP only) \\
\midrule
\multicolumn{2}{l}{\textit{GP Output Layer}} \\
Number of RFF ($D$) & 1024 \\
RFF init std ($1/\ell$) & 0.05 \\
Kernel scale & 1.0 \\
Input normalization & LayerNorm \\
Covariance momentum & $-1$ (exact) \\
Ridge penalty & $10^{-3}$ \\
Mean-field factor & $\pi/8$ \\
\midrule
\multicolumn{2}{l}{\textit{Training}} \\
Optimizer & AdamW \\
Learning rate & $10^{-4}$ \\
Weight decay & 0.01 \\
Batch size & 32 \\
Epochs & 30 \\
Warmup steps & 500 \\
LR schedule & Cosine decay \\
Gradient clipping & 1.0 \\
\bottomrule
\end{tabular}
\end{table}

We applied spectral normalization to both attention and MLP layers within transformer blocks, replacing the classification head with an RFF-GP layer following the original implementation guidelines~\citep{liu2022simple}. We also experimented with applying spectral normalization to MLP layers only, which yielded similar results.

\subsubsection{Results and Discussion}

Despite careful implementation, SNGP-ViT exhibited severely degraded performance on CIFAR-100: training accuracy plateaued around 42\% while test accuracy reached only 32.2\% with an NLL of 2.7. This is substantially worse than all other baselines, including a deterministic ViT which achieves over 76\% test accuracy (Table~\ref{tab:model_performance_cifar100}).

Our observations align with documented challenges in the literature. \citet{vazhentsev2022uncertainty} note that the spectral normalization is theoretically proven to ensure the bi-Lipschitz constraint on the transformation defined by standard residual connection networks. However, the self-attention blocks in Transformers have a more complicated architecture than ResNet layers, which means the theoretical guarantees do not hold in general. Their experiments with SNGP on ELECTRA~\citep{clark2024electra} demonstrate highly unstable performance; on CoNLL-2003, SNGP achieves an RCC-AUC of 56.43 compared to 6.08 for the baseline, nearly an order of magnitude worse.

Similarly, \citet{ulmer2022exploring} report that SNGP significantly underperforms on sequence tasks: on the Danish NER task~\citep{plank2020dan}, SNGP achieves only 22\% accuracy while LSTM achieves 93\%. They observe that even with pre-trained BERT models as feature extractors, training is quite unstable. The NLP Uncertainty Zoo library likewise warns that SNGP transformer models exhibit significant training instability.\footnote{\url{https://dennisulmer.eu/nlp-uncertainty-zoo/nlp_uncertainty_zoo.models.sngp_transformer.html}}

%Given these fundamental architectural incompatibilities and consistently poor empirical results across both our experiments and the literature, we exclude SNGP from our main Vision Transformer comparisons.

%\input{Appendices/ZB_SWAG}

\section{Definitions of Evaluation Metrics}\label{app:metrics}
We primarily evaluate our models on accuracy and  Expected Calibration Error \cite[ECE,][]{Guo2017OnNetworks}. In addition to accuracy and \glsxtrlong{ece}, we have calculated several other scores that have been used in the context of probabilistic deep learning. In the following section, we present the formulations used in our implementations.

%\subsection{Accuracy}
%The accuracy is implemented instance-wise as follows:
%\begin{equation}
%    \mathrm{Acc} = \frac{1}{N} \sum_{i=1}^N \frac{\abs{\hat{y}_i \cap y_i}}{\abs{\hat{y}_i \cup y_i}}
%\end{equation}
%Here $y_i$ denotes the true label of the sample $i$, $\hat{y}_i$ is the predicted label of the sample $i$, and $N$ means the total number of samples.

\subsection{Accuracy}
Accuracy is defined as the fraction of correctly classified samples:
\begin{equation}
    \mathrm{Acc} = \frac{1}{N}\sum_{i=1}^{N} \mathbf{1}(\hat{y}_i = y_i).
\end{equation}
Here $y_i$ denotes the true label of sample $i$, $\hat{y}_i$ is the predicted label of sample $i$, $N$ is the total number of samples, and $\mathbf{1}(\cdot)$ is the indicator function.

\subsection{Expected Calibration Error}
The \glsxtrlong{ece} is a widely used metric for measuring the calibration of neural networks. We use the definition given in \citep{Guo2017OnNetworks}. \glsxtrshort{ece} is defined as the expected difference between accuracy and confidence across several bins. In our experiments, we use $M = 10$ equally spaced bins over the confidence interval $[0,1]$. We first need to define accuracy and confidence per bin $B_m$ as follows: 
\begin{align}
    \mathrm{Acc}(B_m) &= \frac{1}{\abs{B_m}}\sum_{i\in B_m} \mathbf{1}(\hat{y}_i = y_i), \\
    \mathrm{Conf}(B_m) &= \frac{1}{\abs{B_m}} \sum_{i\in B_m} \hat{p}_i.
\end{align}
Again, $y_i$ and $\hat{y}_i$ denote the true and predicted labels of sample $i$ respectively, and $\hat{p}_i$ is the predicted confidence of sample $i$.
With this the \glsxtrlong{ece} is given as:
\begin{equation}
    \mathrm{ECE} = \sum_{m=1}^M \frac{\abs{B_m}}{n}\abs{\mathrm{Acc}(B_m) - \mathrm{Conf}(B_m)}
\end{equation}

\subsection{Macro F1-score}

\begin{equation}
    F1=\frac{1}{C}\sum_{j=1}^{C}\frac{2p_jr_j}{p_j+r_j},
\end{equation}

 where $r_j$ represents the Recall of class $j$, defined as $r_j=\frac{TP}{TP + FN}$, and $p_j$ represents the Precision of class $j$, defined as $p_j=\frac{TP}{TP + FP}$, and $C$ refers to the number of classes, Here, $TP$, $FP$, and $FN$ denote True Positives, False Positives, and False Negatives respectively.
 
\subsection{Negative Log-Likelihood (NLL)}

\begin{equation}
    NLL = -\frac{1}{N}\sum_{i=1}^{N}\sum_{j=1}^{C} \left(y_{i,j}\log\hat{p}_{i,j}\right)=-\frac{1}{N}\sum_{i=1}^{N}\log\hat{p}_{i},
\end{equation}

where $N$ denotes the number of datapoints, $C$ the number of classes, $y_{i,j}$ is 1 if the true label of point $i$ is $j$ and 0 otherwise and $\hat{p}_{i,j}$ is the predicted probability of sample $i$ belonging to class $j$.

\subsection{Brier score}
For Brier score we take the definition by \citep{Brier1950Verification}, which is as follows:

\begin{equation}
    BS = \frac{1}{N}\sum_{i=1}^{N}\sum_{j=1}^C(\hat{p}_{i,j} - y_{i,j})^2,
\end{equation}

where $N$ denotes the number of datapoints, $C$ the number of classes, $y_{i,j}$ is 1 if the true label of point $i$ is $j$ and zero otherwise and $\hat{p}_{i,j}$ is the predicted probability of sample $i$ belonging to class $j$.

\subsection{Area Under the Receiver Operating Characteristic Curve (AUROC)} The AUROC score evaluates the performance of a binary classifier by measuring its ability to distinguish between positive and negative classes, as introduced by \citep{hanley1982meaning}. In our out-of-distribution (OOD) detection experiments, the positive class corresponds to an in-distribution sample, while the negative class corresponds to an out-of-distribution sample.

The AUROC is computed as the area under the ROC curve, which plots the true positive rate (TPR) against the false positive rate (FPR) across various decision thresholds. The TPR and FPR are defined as follows:

\begin{align} \mathrm{TPR} &= \frac{\text{TP}}{\text{TP} + \text{FN}}, 
\end{align}

\begin{align} \mathrm{FPR} &= \frac{\text{FP}}{\text{FP} + \text{TN}}, 
\end{align}

where TP, FP, FN, and TN represent the true positives, false positives, false negatives, and true negatives, respectively.

The AUROC score is given by the following integral:

\begin{equation} \mathrm{AUROC} = \int_{0}^{1} \mathrm{TPR}(\mathrm{FPR}) , d\mathrm{FPR}. \end{equation}

A higher AUROC score indicates better classification performance, with a score of 1 representing a perfect classifier, and a score of 0.5 indicating performance equivalent to random chance.

\subsection{Area Under the Precision-Recall Curve (AUPRC)}

The Area Under the Precision-Recall Curve (AUPRC) assesses the performance of a binary classifier by measuring its ability to accurately identify positive instances, as described by \citep{davis2006relationship}. In our out-of-distribution (OOD) detection experiments, the positive class corresponds to in-distribution samples, while the negative class corresponds to out-of-distribution samples.

The AUPRC is calculated as the area under the Precision-Recall (PR) curve, which plots precision against recall at various decision thresholds. Precision and recall are defined as follows:

\begin{equation}
\mathrm{Precision} = \frac{\text{TP}}{\text{TP} + \text{FP}},
\end{equation}

\begin{equation}
\mathrm{Recall} = \frac{\text{TP}}{\text{TP} + \text{FN}},
\end{equation}
where TP, FP, and FN represent true positives, false positives, and false negatives, respectively.

The AUPRC score is the integral of precision with respect to recall, expressed as:

\begin{equation}
\mathrm{AUPRC} = \int_{0}^{1} \mathrm{Precision}(\mathrm{Recall}) \, d\mathrm{Recall}.
\end{equation}

A higher AUPRC score indicates better classifier performance in recognizing positive instances, with a score near 1 representing a good classifier, characterized by both high recall and high precision. This metric is especially valuable for evaluating classifiers on imbalanced datasets.

\subsection{False Positive Rate at 95\% True Positive Rate (FPR@95\%TPR)}
We use the false positive rate at 95\% true positive rate (FPR@95\%TPR) as an evaluation metric. This metric measures the proportion of negative samples that are incorrectly classified as positives when the true positive rate is fixed to 95\%. Lower values of FPR@95\%TPR indicate better performance,  corresponding to fewer false positives at the same, high true positive rate.

Formally, let $\tau$ be a decision threshold such that
\begin{equation}
\mathrm{TPR}(\tau) = \frac{\mathrm{TP}(\tau)}{\mathrm{TP}(\tau) + \mathrm{FN}(\tau)} = 0.95.
\end{equation}
The reported score is then
\begin{equation}
\mathrm{FPR@95\%TPR} = \mathrm{FPR}(\tau)
= \frac{\mathrm{FP}(\tau)}{\mathrm{FP}(\tau) + \mathrm{TN}(\tau)}.
\end{equation}

%\input{Appendices/GenerativeAI_statement}

%\clearpage
%\bibliography{neurips_2025}

%\input{Appendices/L_results_rebutal}

%\input{Appendices/0A_training_details}

%\input{Appendices/0AA_LoRA_Rank_Ablation}

%\input{Appendices/A_LoRA_Initialization_Study}

%\input{Appendices/B_Explicit_Noise_Study}

%\input{Appendices/C_Validation_AST}

%\input{Appendices/D_Hyperparameter_Tuning_AST}

%\input{Appendices/E_MCDropout_Rate}

%\input{Appendices/F_Supplementary_Evaluation_Metrics}

%\input{Appendices/G_Supplementary_Figures}

%\input{Appendices/J_Eval_Metrics}

\end{document}